%% file: main.tex
\documentclass[11pt]{article}

\usepackage[english]{babel}

\usepackage[letterpaper,margin=1in]{geometry}

\usepackage{amsmath, mathtools, amssymb}
\usepackage{graphicx}
\usepackage{microtype}
\usepackage[scr=boondoxo]{mathalpha}
\usepackage{tabularx}
\usepackage{enumitem}
\setlist{topsep=3pt,itemsep=2pt}
\usepackage{subcaption}
\usepackage{booktabs}
\usepackage{amsthm}
\usepackage{bbm}
\usepackage{algorithm}
\usepackage{algpseudocode}
\newenvironment{ruledalg}[1]{\par\noindent\hrule height.8pt\kern2pt
  \captionof{algorithm}{#1}\kern2pt\hrule\kern2pt\raggedright}{\par\kern2pt\hrule}
\usepackage[dvipsnames]{xcolor}
\usepackage{tikz}
\usetikzlibrary{arrows.meta,calc,fit,matrix,backgrounds}
\usepackage{csquotes}
\usepackage[
    backend=biber,
    style=alphabetic,
    natbib=true,
    maxbibnames=99,
    maxalphanames=3,
    minalphanames=3
]{biblatex}
\renewrobustcmd*{\citet}{\cite}
\AtBeginBibliography{\emergencystretch=1.5em}
\usepackage[colorlinks=true,allcolors=blue]{hyperref}
\usepackage[capitalize,noabbrev]{cleveref}
\theoremstyle{plain}
\newtheorem{theorem}{Theorem}[section]
\newtheorem{proposition}[theorem]{Proposition}
\newtheorem{lemma}[theorem]{Lemma}
\newtheorem{fact}[theorem]{Fact}
\crefname{fact}{fact}{facts}
\Crefname{fact}{Fact}{Facts}
\newtheorem{corollary}[theorem]{Corollary}
\newtheorem{claim}[theorem]{Claim}
\theoremstyle{definition}
\newtheorem{definition}[theorem]{Definition}

\theoremstyle{remark}

\theoremstyle{plain}

\makeatletter
\newcounter{algorithmicH}
\let\oldalgorithmic\algorithmic
\renewcommand{\algorithmic}{\stepcounter{algorithmicH}\oldalgorithmic}
\def\theHALG@line{ALG@line.\thealgorithmicH.\arabic{ALG@line}}
\makeatother

\algrenewcommand\algorithmicrequire{\textbf{Input:}}

\def\br{\mathbf{u}}

\def\bbE{{\mathbb E}}

\def\bbN{{\mathbb N}}

\def\bbR{{\mathbb R}}

\newcommand{\set}[1]{\left\{#1\right\}}
\newcommand{\zo}{\set{0,1}}

\newcommand{\eps}{\varepsilon}

\newcommand{\ind}{\mathbbm{1}}                 
\newcommand{\Lap}{\operatorname{Lap}}          
\newcommand{\Bin}{\operatorname{Bin}}          
\newcommand{\Gauss}{\mathcal{N}}               
\newcommand{\Renyi}[1]{\boldsymbol{R}_{#1}}    
\newcommand{\rord}{\vartheta}                  
\newcommand{\acc}{\alpha}                      
\newcommand{\fprob}{\beta}                     

\newcommand{\dom}{\mathcal{X}}      
\newcommand{\nul}{\bot}             
\newcommand{\ds}{s}                 
\newcommand{\fds}{{\hat{\ds}}}        
\newcommand{\cand}{x}               
\newcommand{\dist}{P}               
\newcommand{\fail}{\mathsf{fail}}   
\newcommand{\nslots}{n}             
\newcommand{\bsz}{b}        
\newcommand{\nblk}{m}               
\newcommand{\SV}{\mathrm{SV}}       
\newcommand{\filt}{\mathcal{F}}     
\newcommand{\pred}{F}               
\newcommand{\qry}{q}                
\newcommand{\qfam}{\mathbf{q}}      
\newcommand{\tbit}{u}               
\newcommand{\cnt}{c}                
\newcommand{\nfail}{{N_{\mathrm f}}} 
\newcommand{\athr}{t}               
\newcommand{\svcap}{\kappa}          
\newcommand{\ncap}{{\tilde\kappa}}   
\newcommand{\bcap}{{\bar\kappa}}     
\newcommand{\svscale}{\sigma}            
\newcommand{\qnoise}{\gamma}        
\newcommand{\cnoise}{\eta}          
\newcommand{\qscale}{{\sigma'}}     
\newcommand{\cscale}{{\sigma}}           
\newcommand{\qmargin}{{\bar{\qnoise}}}   
\newcommand{\cmargin}{{\bar{\cnoise}}}   
\newcommand{\fpar}{\bsz,\athr,\svcap,\cscale,\qscale} 
\newcommand{\ufrac}{{p_0}}            
\newcommand{\bfail}{{\beta_0}}        
\newcommand{\PFR}{\mathrm{PFR}}     
\newcommand{\PFRd}{\PFR^{\eps,\delta}_{\bsz}} 
\newcommand{\mech}{M}               
\newcommand{\fscore}{\varphi}       
\newcommand{\thr}{\tau}             
\newcommand{\Lconst}{L}             
\newcommand{\nint}{\nu}             
\newcommand{\ppred}{\psi}              
\newcommand{\cCap}{12}                                
\newcommand{\cOffset}{\the\numexpr2*\cCap\relax}      
\newcommand{\cSlack}{\cCap}                           
\newcommand{\cBound}{\the\numexpr3*\cCap\relax}       
\newcommand{\cTest}{108}                              
\newcommand{\cBlk}{400}                               
\newcommand{\cRet}{1800}                              
\newcommand{\Lbase}{16}                               
\newcommand{\Ldef}{\ln(\Lbase/\delta)}                
\newcommand{\cFprob}{\the\numexpr\Lbase/2\relax}      
\newcommand{\fprobMin}{\delta/\cFprob}                
\newcommand{\cFprobQ}{\the\numexpr\cFprob/4\relax}    
\newcommand{\athrOf}[1]{7#1/8}                        
\newcommand{\retMarginOf}[1]{#1/16}                   
\newcommand{\athrRetOf}[1]{15#1/16}                   
\newcommand{\athrVal}{\athrOf{\nblk}}
\newcommand{\athrValNB}{7\nslots/(8\bsz)}             
\newcommand{\retMargin}{\retMarginOf{\nblk}}
\newcommand{\athrRet}{\athrRetOf{\nblk}}
\newcommand{\qscaleVal}{\cTest\Lconst/\eps^{3/2}}     
\newcommand{\cscaleVal}{\cCap/\eps}                   
\newcommand{\capOffset}{\cOffset\Lconst/\eps}         
\newcommand{\capSlack}{\cSlack\Lconst/\eps}           
\newcommand{\mPriv}{\cBlk\Lconst^2/\eps^{3/2}}        
\newcommand{\nPriv}{\cBlk\bsz\Lconst^2/\eps^{3/2}}    
\newcommand{\mRet}[1]{\cRet\Lconst\ln(2\nslots/#1)/\eps^{3/2}} 
\newcommand{\mRetQ}{\cRet\Lconst\ln(8\nslots/\fprob)/\eps^{3/2}} 
\newcommand{\eFilt}{\eps/2}                           
\newcommand{\eRel}{\eps/2}                            
\newcommand{\dRel}{\delta/4}                          
\newcommand{\dCap}{\delta/32}                         
\newcommand{\dFilt}{17\delta/32}                      
\newcommand{\dFriend}{\delta/32}                      
\newcommand{\cCapJDP}{8}                              
\newcommand{\cBoundJDP}{\the\numexpr3*\cCapJDP\relax} 
\newcommand{\eCapJDP}{\eps/\cCapJDP}                  
\newcommand{\eSVJDP}{\the\numexpr\cCapJDP-1\relax\eps/\cCapJDP} 
\newcommand{\svscaleVal}{7\sqrt{\svcap\ln(2/\delta)}/\eps} 
\newcommand{\dimn}{d}               
\newcommand{\rank}{k}               
\newcommand{\ridge}{\lambda}        
\newcommand{\stat}{g}               
\newcommand{\psens}{\Delta}         
\newcommand{\smom}{Q}               
\newcommand{\tracescore}{{\fscore_{\mathrm{tr}}}}
\newcommand{\tracestat}{{\stat_{\mathrm{tr}}}} 
\newcommand{\gaussprof}{\delta_{\Gauss_{\sigma}}} 
\newcommand{\lev}{\mathcal{L}}      
\newcommand{\regmom}{\widehat\Sigma} 
\newcommand{\nsk}{r}                
\newcommand{\slev}{\omega}          
\newcommand{\rdpbd}{\rho}           
\newcommand{\loss}{\mathrm{loss}}
\newcommand{\ncat}{d}               
\newcommand{\pcount}{\lambda}       
\newcommand{\KL}{D_{\mathrm{KL}}}   
\newcommand{\unif}{u}               
\newcommand{\sfreq}{{\hat p}}       
\newcommand{\klstat}{{\stat_{\mathrm{KL}}}} 
\newcommand{\fKL}{{\fscore_{\mathrm{KL}}}}  
\newcommand{\hypclass}{\mathcal{H}} 
\newcommand{\nhyp}{N}              
\newcommand{\weight}{w}            
\newcommand{\totweight}{W}         
\newcommand{\wrisk}{R}             
\newcommand{\sensW}{{\fscore_{\mathrm{w}}}} 
\newcommand{\cThr}{4}                                 
\newcommand{\cBsz}{4}                                 
\newcommand{\cBszW}{2}                                
\newcommand{\Kconst}{\sqrt{\nsk\Lconst}+\Lconst}      
\newcommand{\sketchPriv}{2(\Kconst)/(\thr\eps)}       
\newcommand{\empmom}{\bar\Sigma}    
\newcommand{\tQ}{{J_Q}}             

\newcommand{\Tr}{\text{Tr}}

\newcommand{\Tableofcontents}{
	\thispagestyle{empty}
	\hypersetup{pageanchor=false}
	\pagenumbering{gobble}
	\clearpage
	\setcounter{tocdepth}{2}
	\tableofcontents
	\thispagestyle{empty}
	\clearpage
	\pagenumbering{arabic}
	\hypersetup{pageanchor=true}
}

\newcommand{\wtO}{\widetilde O}
\newcommand{\wtOmega}{\widetilde\Omega}
\newcommand{\projk}{{P_{\rank}}}
\newcommand{\effdim}{{D_{\ridge}}}
\newcommand{\bdry}{a}
\newcommand{\hist}{H}
\newcommand{\nobs}{T}
\newcommand{\avgH}{{\bar H}}
\newcommand{\verts}{V}
\newcommand{\lapl}{\mathbf L}
\newcommand{\reff}{{R_{\mathrm{eff}}}}
\newcommand{\lmax}{{\ell_{\max}}} 
\newcommand{\cov}{z}
\newcommand{\trt}{a}
\newcommand{\outc}{y}
\newcommand{\prop}{\pi}

\title{Propose-Filter-Release: Scaling Noise to Typical Sensitivity via Private Filtering}

\ifdefined\IsAnonymous
\author{Anonymized for Submission}
\else
\author{
	Moshe Shenfeld\thanks{Equal contribution.}\textsuperscript{\hphantom{$\ast$},}\thanks{School of Computer Science and Engineering, The Hebrew
	University of Jerusalem.\newline E-mail: \texttt{moshenfeld@gmail.com}.}
	\and
	Eliad Tsfadia\footnotemark[1]\textsuperscript{\hphantom{$\ast$},}\thanks{Department of Computer Science and Artificial Intelligence, Bar-Ilan University.\newline E-mail: \texttt{eliad.tsfadia@biu.ac.il}.}
	\and
	Katrina Ligett\thanks{School of Computer Science and Engineering and Federmann Center for the Study of Rationality, The Hebrew
	University of Jerusalem.\newline E-mail: \texttt{katrina.ligett@mail.huji.ac.il}.}
}
\fi
\date{}

\begin{document}
\maketitle

\begin{abstract}
Differential privacy requires that algorithms be stable to the change of any record in their input, even to an extreme outlier.
Private algorithms thus commonly introduce noise scaled to the worst-case possible influence of any single record.
This work studies how to build differentially private algorithms that can effectively tailor their noise to the typical case rather than the worst case.

We introduce the general \emph{propose-filter-release} (PFR) framework, which, by privately filtering its input, turns a mechanism that protects the addition or removal of a \emph{typical} record into one that is differentially private in the standard sense, on \emph{every} input.
We show how to instantiate PFR with a novel filtering step to obtain a general meta-algorithm.
Given only a concave friendliness score (which measures how typical a record is with respect to a dataset) and a release mechanism that is friendly-DP (that is, private conditioned on the changed record being typical of the remaining data), it guarantees differential privacy on sufficiently large datasets.
The FriendlyCore \cite{FriendlyCore22} and Propose-Test-Release \citep{dwork2009differential} algorithms can be viewed as instantiations of our general approach.

We illustrate the generality of PFR by instantiating it for scalar statistics, principal subspace estimation with a dimension-independent sample size, the KL divergence of user histograms, the effective resistances in a resistor network, and treatment-rule selection.
\end{abstract}

\Tableofcontents

\input{sections/sec_introduction}
\input{sections/sec_preliminaries}
\input{sections/sec_filtering}
\input{sections/sec_pfr}

\input{sections/sec_applications}
\input{sections/sec_discussion}
\label{page:body-end}
\clearpage
\appendix
\input{sections/apd_preliminaries}
\input{sections/apd_filtering}
\input{sections/apd_pfr}
\input{sections/apd_applications}

\subsection*{AI Methodology Statement}
In this work, the authors used OpenAI's Codex and Anthropic's Claude to assist with the literature review and with drafting and revising text throughout the main text and appendices, including proofs.
The main research ideas were developed solely by the authors.
All parts of the paper, including AI-assisted material, were checked and edited by the authors, who take full responsibility for the paper's content, correctness, originality, and references.

\subsection*{Acknowledgments}
Shenfeld and Ligett's work was supported in part by ERC grant 101125913, Simons Foundation Collaboration 733792, Israel Science Foundation (ISF) grant 2861/20, and a gift from Apple.
Views and opinions expressed are however those of the authors only and do not necessarily reflect those of the European Union or the European Research Council Executive Agency.
Neither the European Union nor the granting authority can be held responsible for them.
Shenfeld's work was additionally supported by the Apple Scholars in AI/ML PhD Fellowship.
\clearpage
\printbibliography
\end{document}

%% file: sections/sec_introduction.tex
\section{Introduction}\label{sec:intro}

Statistical analysis uses individual data samples to estimate population-level patterns.
When the samples are individuals' personal data, one might try to protect their privacy by enforcing that the statistical results not reveal much about any single contributor.
\emph{Differential privacy} (DP)~\cite{dwork2006calibrating} makes this requirement precise: a randomized algorithm is differentially private if changing the record of any single individual barely changes the distribution of its output.\footnote{For convenience, we model a change of one individual's data as replacing their record with a null placeholder, typically referred to as \emph{zero-out} adjacency \citep{kairouz2021practical}. This notion is essentially the same as removing it, and equivalent, up to constant factors in the privacy parameters, to replacing it with another record (\Cref{claim:zero_out_replacement}).}
This must hold for every neighboring pair of datasets that differ in one individual, including unlikely datasets and those on which the analysis is not expected to be accurate.
A common approach in the differential privacy literature calibrates the noise added by the algorithm to the worst-case change between any neighboring pair, which usually is induced by swapping in a record that is atypical of the rest of its dataset.
On typical data, where every record is fairly typical, such noise can be disproportionate to the influence of any of the records.
This work develops a general private technique that allows noise to be calibrated to a bound on the influence of typical records, which we informally call the \emph{typical sensitivity}.
The approach is allowed to halt or return an uninformative answer on unlikely, atypical datasets with many outliers, but privacy must hold even for the worst case.

\paragraph{An illustrative example.}
Our running example is principal subspace estimation.
Consider independent samples from a centered Gaussian distribution in $\bbR^\dimn$ that lies close to the subspace spanned by the $\rank$ leading eigenvectors of its covariance, in the sense that the other $\dimn-\rank$ directions together demonstrate little variance.
The samples thus concentrate near this subspace (see \Cref{fig:friendliness} for an illustration), and for suitable spectra, one can estimate this subspace accurately from a number of samples that grows only with its dimension $\rank$, rather than the ambient dimension $\dimn$ (\Cref{sec:sketching}).
Here, the error of an estimate is the distance in operator norm between its projector and the projector onto this subspace.
One such spectrum, which we call the \emph{large-gap spectrum}, has $\rank$ eigenvalues in $[1/2,1]$ and $\dimn-\rank$ eigenvalues of at most $\dimn^{-2}$.
One might hope that privacy would not change this picture, so that a private estimate would also need a number of samples that grows only with $\rank$.
The rest of the introduction surveys prior work, illustrates where various existing techniques fall short of this hope, and discusses how our technique avoids an explicit dependence on $\dimn$ in the running example.

\paragraph{Global and local sensitivity.}
The most common way to achieve DP adds noise calibrated to the target statistic's \emph{global sensitivity}, the largest change in it induced by any two neighboring datasets~\cite{dwork2006calibrating}.
This worst-case calibration is private on every dataset, but it ignores the structure that makes typical data easy.
For instance, a single sufficiently large record orthogonal to the empirical subspace of the other records can drastically change the empirical subspace, regardless of the sample size.
An instance-dependent alternative is the \emph{local sensitivity}~\cite{nissim2007smooth}, the largest change in the statistic between a given dataset and one of its neighbors, which can in some cases be much smaller on typical data.
Per-instance differential privacy (pDP)~\cite{wang2019perinstance} goes one step further in this direction: it bounds the privacy loss of a mechanism at a single pair of neighboring datasets, rather than over all the neighbors of a given dataset.
Simply calibrating the noise to an instance-dependent sensitivity measure, however, does not generally protect privacy, since the measure itself, and hence the noise scale, can reveal information about the underlying data.

\paragraph{Propose-test-release.}
A clever workaround might be to rely on a bound on the local sensitivity only after privately testing that the data satisfy it.
The \emph{propose-test-release} (PTR) framework \citep{dwork2009differential} introduced this approach, which combines two kinds of privacy guarantees.
Its inner mechanism adds noise scaled to a proposed bound on the local sensitivity, and it is private only when run on a dataset where the bound holds both on it and on its neighbors.
The complete algorithm runs in two stages: it first tests privately whether the bound holds, and then either runs the inner mechanism or reports failure.
Checking the bound on the dataset itself would not be private, since a single record can flip the answer.
Instead, the test measures how far, in terms of the number of records changed, the dataset is from one that violates the bound.
This distance has low global sensitivity, and thus can easily be privatized.
PTR thus already converts a mechanism that is private only under some condition to one that is fully private.
Generalized PTR~\cite{redberg2023gptr} privately tests a data-dependent bound on the privacy loss of a mechanism instead, which extends PTR beyond noise addition.

A potential drawback of both forms of PTR is that their test concerns the dataset as a whole: a dataset fails if even one of its records or the addition of a single atypical record causes a violation, even if all its other records are typical.
In the running example, \emph{removing} one of the samples barely changes their empirical subspace (\Cref{fig:friendliness}), but, as noted above, \emph{adding} an arbitrary record can change it drastically.
The local sensitivity of the empirical subspace is therefore large near every dataset, and PTR applied directly to any nontrivial bound on it fails on every dataset, except with small probability (\Cref{prop:ptr_fails}).
A similar issue arises when applying PTR to other statistics that are not robust to outliers, such as the mean of unbounded vectors.

\paragraph{Private filtering.}
To overcome this limitation of PTR, one might hope instead to privately filter out the records that are not typical of the rest, and then run a mechanism that only guarantees privacy on the datasets that such a filter can produce.
\emph{FriendlyCore}~\cite{FriendlyCore22} does so through pairwise comparisons: a predicate specifies which pairs of records are \emph{friends}, its filtering step keeps the records that have many friends, and its release mechanism needs to be private only on datasets in which every two records have a common friend.
As with the distance in PTR, the number of friends of each record changes by at most one when another record changes, and can thus be easily privatized.
A distance-based friendliness notion, for example, would retain a concentrated cluster of records and exclude a few distant outliers.
The distance between two of the Gaussian samples in our running example, however, does not reflect their influence on the subspace.
Two records far apart along the subspace would not be friends under the element-to-element distance notion, although both may be typical; however, a record that lies off the subspace but near the origin may be a friend of many despite being an outlier (see \Cref{fig:friendliness}a for an illustration).
This illustrates a potential limitation of filtering based on \emph{element--element} friendliness.

To capture structure that single records do not reveal, FriendlyCore can also be applied in the spirit of \emph{sample-and-aggregate}~\cite{nissim2007smooth}, comparing groups of records instead of single records: the records are split into blocks, a model is estimated non-privately on each block, and the models are aggregated privately.
For subspace estimation, \citet[Algorithms~4.1--4.3]{tsfadia2024subspace} applies FriendlyCore to the subspaces estimated on the blocks, an approach that we call \emph{block--block} FriendlyCore.
\citet{singhal2021subspaces} instead aggregate these subspaces through a private histogram of projected reference points.
To avoid a dependence on $\dimn$ in our running example, the published guarantees of both methods require the ratio between the $(\rank+1)^{th}$ and $\rank^{th}$ eigenvalues to decrease polynomially with $\dimn$.
The large-gap spectrum satisfies this condition.
A harder example is the \emph{geometric spectrum}, which has $\rank-\lceil\log_2(\dimn)\rceil$ unit eigenvalues followed by eigenvalues that halve at every step, so that its $\rank^{th}$ eigenvalue is only twice the next.
On this spectrum, these guarantees depend polynomially on $\dimn$ or do not apply.

\begin{figure}[t]
\centering
\begin{tikzpicture}[
    font=\footnotesize,
    stage/.style={draw, rounded corners=2pt, align=center, inner sep=3pt, minimum height=10mm, minimum width=20mm},
    data/.style={align=center, inner sep=2pt},
    flow/.style={-{Stealth[length=2mm]}, semithick},
    link/.style={-{Stealth[length=1.6mm]}, densely dotted, semithick},
    yes/.style={text=green!50!black},
    no/.style={text=red!70!black},
    blk/.style={draw, rounded corners=1.5pt, fill=gray!10, minimum width=9mm, text height=1.9ex, text depth=0.45ex},
    set/.style={draw, rounded corners=1.5pt, fill=gray!10},
    pick/.style={blue!70!black},
]
\node[data] (in) at (0,1.45) {$\ds$};
\node[stage, fill=blue!6] (filt) at (1.55,1.45) {JDP filtering\\step $\filt$};
\node[data] (fds) at (3.2,1.45) {$\fds$};
\node[stage, fill=orange!10] (rel) at (4.85,1.45) {friendly-DP\\mechanism $\mech$};
\node[data] (out) at (6.75,1.45) {$\mech(\fds)$};
\draw[flow] (in) -- (filt);
\draw[flow] (filt) -- (fds);
\draw[flow] (fds) -- (rel);
\draw[flow] (rel) -- (out);
\node[data, font=\scriptsize, text=black!75, anchor=south] (expl) at (3.2,2.0) {$\ds$ with some\\records set to $\nul$};
\draw[gray, thin] (expl) -- (fds);
\node[data, anchor=north] (cond) at (3.2,0.35) {$\pred(\fds^{i\to\nul},\fds_i)=1$, for each $\fds_i\ne\nul$};
\draw[link] ([xshift=0.85cm]filt.south) -- node[left, font=\scriptsize] {ensures w.h.p.} ([xshift=0.85cm]filt.south |- cond.north);
\draw[link] ([xshift=-0.8cm]rel.south |- cond.north) -- node[right, font=\scriptsize] {private under} ([xshift=-0.8cm]rel.south);
\node[draw, densely dashed, rounded corners=4pt, fit=(filt)(rel)(expl)(cond), inner xsep=5pt, inner ysep=3pt] (box) {};
\node[data, anchor=south] at (box.north) {DP algorithm $\mech\circ\filt$};
\node[font=\footnotesize\bfseries] at (box.north |- 0,3.5) {(a) PFR};
\begin{scope}[xshift=8.0cm]
\matrix (T) [matrix of math nodes, anchor=north west, row sep=1.5pt, column sep=0pt,
    nodes={inner sep=1pt, minimum width=6.5mm, text height=1.9ex, text depth=0.45ex, anchor=center}]
    at (0,3.1) {
    &[3pt] \ds_1 & \ds_2 & \ds_3 & \ds_4 & \ds_5 & \cdots & |[pick]|\ds_\nslots \\[3pt]
    |[blk]|\ds^{(1)} &[3pt] |[yes]|1 & |[yes]|1 & |[yes]|1 & |[no]|0 & |[yes]|1 & \cdots & |[yes]|1 \\
    |[blk, draw=blue!70!black, text=blue!70!black]|\ds^{(2)} &[3pt] |[yes]|1 & |[no]|0 & |[yes]|1 & |[no]|0 & |[yes]|1 & \cdots & |[yes]|1 \\
    \vdots &[3pt] \vdots & \vdots & \vdots & \vdots & \vdots & \ddots & \vdots \\
    |[blk]|\ds^{(\nblk)} &[3pt] |[yes]|1 & |[yes]|1 & |[no]|0 & |[no]|0 & |[yes]|1 & \cdots & |[yes]|1 \\[3pt]
    &[3pt] \ds_1 & \ds_2 & \ds_3 & |[no]|\nul & \ds_5 & \cdots & \ds_\nslots \\
};
\begin{pgfonlayer}{background}
\node[set, draw=none, fit=(T-1-2)(T-1-8), inner sep=0pt] {};
\node[set, draw=none, fit=(T-6-2)(T-6-8), inner sep=0pt] {};
\fill[red!8] (T-2-5.north west) rectangle (T-6-5.south east);
\end{pgfonlayer}
\node[set, fill=none, fit=(T-1-2)(T-1-8), inner sep=0pt] (S) {};
\node[set, fill=none, fit=(T-6-2)(T-6-8), inner sep=0pt] (Shat) {};
\node[anchor=east, inner sep=1pt, text height=1.9ex, text depth=0.45ex] at ($(S.west |- T-1-2)+(-1.5pt,0)$) {$\ds={}$};
\node[anchor=east, inner sep=1pt, text height=1.9ex, text depth=0.45ex] at ($(Shat.west |- T-6-2)+(-1.5pt,0)$) {$\fds={}$};
\draw[gray!60] (T-2-2.north west) rectangle (T-5-8.south east);
\node[font=\footnotesize\bfseries] at (T.north |- 0,3.5) {(b) Element--block filtering};
\node[draw=blue!70!black, circle, inner sep=0pt, minimum size=4.6mm] (sel) at (T-3-8) {};
\node[data, pick, anchor=west] (callout) at ($(T-3-8.east)+(0.45,0.5)$) {$\pred(\ds^{(2)},\ds_\nslots)$};
\draw[pick, -{Stealth[length=1.6mm]}] (callout.south west) to[out=-120, in=30] (sel);
\end{scope}
\end{tikzpicture}
\caption{
(a) The general Propose-Filter-Release (PFR) scheme (\Cref{lem:friendly_comp}): a jointly differentially private step filters individual records so that, for each fixed slot $i$, the filtered dataset $\fds$ satisfies with high probability the condition under which the release mechanism $\mech$ is private.
The complete mechanism, whose only output is the output of $\mech$, is then differentially private on every input.
(b) Element--block filtering is one possible instantiation of the filtering step in the general PFR scheme.
Entry $(j,i)$ of the table is the friendliness predicate $\pred(\ds^{(j)},\ds_i)$ of record $\ds_i$ against reference block $\ds^{(j)}$.
The filter sets $\fds_i=\nul$ if the friend count $\cnt_i$ (the number of ones in column $i$) plus noise is at most the acceptance threshold $\athr$.}\label{fig:pfr_framework}
\end{figure}

\subsection{Our contribution}\label{sec:contribution}
Our contribution toward the goal stated above is twofold.
First, we introduce \emph{propose-filter-release} (PFR), a general technique that extends the structure of FriendlyCore to other filtering steps and other conditions for privacy (\Cref{fig:pfr_framework}a).
Second, we develop \emph{element--block filtering}, a new filtering step for PFR that helps capture meaningful structure in data (\Cref{fig:pfr_framework}b).

\paragraph{Propose-filter-release.}
PFR generalizes PTR: rather than running a test that either retains the entire dataset or halts (\Cref{def:PTR}), PFR applies a record-\emph{filtering step} that separately keeps or nulls each individual record, followed by a \emph{release mechanism} that runs on the filtered data.
The release mechanism need only be private with respect to neighboring datasets whose differing record satisfies a sort of typicality condition with respect to the remaining data.
We call a release mechanism that is private under this condition \emph{friendly-DP}; it is pDP at every pair that satisfies the condition.
The filtering step must be jointly differentially private (JDP)~\cite{kearns2014jdp}: on neighboring datasets it must induce filtered output distributions that would be indistinguishable if one hid the entry on which the neighbors differ.
The filtering step must also ensure, for every input dataset and each fixed slot, that the probability of retaining the record in that slot while it is not friendly with the remaining filtered data is small.
Our \emph{PFR theorem} (\Cref{lem:friendly_comp}) shows that these requirements make the complete mechanism differentially private on every input.
Since the filtering step nulls individual records, PFR need not fail when the dataset is close to a single violation, unlike PTR.
FriendlyCore can be viewed as one instantiation of PFR.\footnote{Formally, the FriendlyCore paper suggests two filters. Their ``zCDP filter'' is a direct instantiation of PFR, but the connection to their ``BasicFilter'' is somewhat more subtle.}

\begin{figure}[t]
\centering
\includegraphics[width=\textwidth]{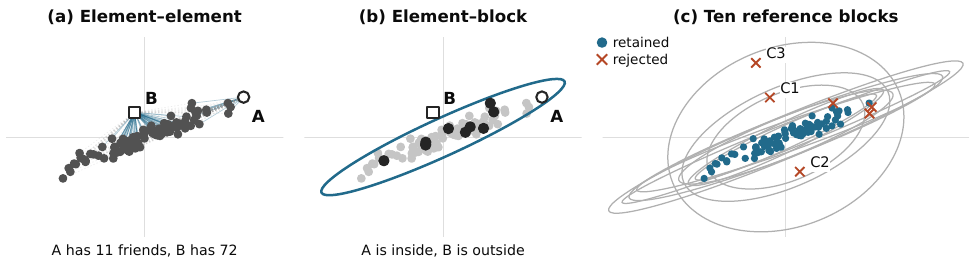}
\caption{An illustration of the friendliness tests on Gaussian data in $\bbR^2$, in which the probes A and B are not records of the data.
(a) Using an element--element distance-based friendliness notion, A, which lies along the data, has few friends; B, which lies off the data near the origin, has many.
(b) Using an element--block friendliness notion, A is friendly with the block consisting of the black points; B is not.
(c) Element--block tests against several reference blocks separate the outliers C1, C2, and C3 while retaining the vast majority of the typical data.}\label{fig:friendliness}
\end{figure}

\paragraph{Element--block filtering.}
Comparing records with blocks can capture structure that other tests can miss (\Cref{fig:friendliness}a,b).
Element--block filtering evaluates each record against several \emph{reference blocks}, random disjoint subsets of the data, using a \emph{friendliness score} (\Cref{fig:friendliness}c).
It nulls the records that fail too many tests and passes the rest of the records to the release mechanism.
If all but sufficiently few records have friend counts exceeding the acceptance threshold by a suitable noise margin, then with high probability the filter does not abort and retains all such records (\Cref{thm:out_reg_acc}).
The filtering step aborts when too many records are nulled, so PFR with element--block filtering tolerates order $\log(1/\delta)/\eps$ outliers, not a constant fraction of contamination.
In order for element--block friendliness to imply that a record is also typical of the entire dataset, we need a further technical condition, that the friendliness score be \emph{concave}: the score of a record against a union of datasets is at least the size-weighted average of its scores against them (\Cref{sec:technique}).

With element--block filtering, PFR becomes a meta-algorithm, instantiated by a concave friendliness score and a release mechanism that is friendly-DP with respect to it.
Applied to this filtering step, the PFR theorem shows that the resulting algorithm is private given sufficiently many reference blocks.

\begin{theorem}[PFR with element--block filtering, informal version of \Cref{thm:gen_frnd_core_priv}]\label{thm:intro:gen_frnd_core_priv}
There exists a universal constant $C$ such that the following holds.
Given $\nslots,\bsz\in\bbN$; $\eps\in(0,1]$; $\delta\in(0,1)$; a friendliness threshold $\thr>0$, a concave friendliness score $\fscore$, and a release mechanism $\mech$ that is $(\eRel,\dRel)$-friendly-DP with respect to the friendliness predicate $\pred_\thr(\fscore;\ds,\cand)\coloneqq\ind\{\fscore(\ds,\cand)\ge\thr\}$, if the number $\nblk=\nslots/\bsz$ of reference blocks of size $\bsz$ is at least $C\ln^2(\Lbase/\delta)/\eps^{3/2}$, then PFR with element--block filtering is $(\eps,\delta)$-DP on datasets of $\nslots$ records.
\end{theorem}

Privacy holds also on inputs with many outliers, on which the process may abort.
The required number of blocks depends only on the privacy parameters, so filtering uses a number of records linear in the block size, which adds to the sample requirement of the release mechanism rather than multiplying it.
Unlike in block--block FriendlyCore, the blocks serve only to test records and not to estimate the answer, so their size need not depend on a target accuracy (\Cref{fig:friendliness}b,c).

\paragraph{Subspace estimation.}
In our running example, the meta-algorithm needs only a release mechanism and a concave score.
The release mechanism returns $\nsk$ Gaussian \emph{sketches}, random Gaussian combinations of the records, whose principal subspace is the estimate.
Nulling a record changes the distribution of the sketches by an amount controlled by its \emph{leverage} with respect to the remaining data, a squared Mahalanobis norm that is small along the directions in which the data vary and large across them.
The friendliness score, the reciprocal of the leverage, is concave, and the test asks whether a record lies in an ellipsoid determined by a reference block (\Cref{fig:friendliness}b).
Neither the privacy of the sketches nor that of the filter depends on $\dimn$ (\Cref{thm:filt_sketch_DP}).
With a small ridge added to the leverage, a typical Gaussian record passes the tests once the blocks are somewhat larger than the number of directions whose variance exceeds the ridge, which on both spectra of the running example is of order $\rank$ rather than $\dimn$.
On the geometric spectrum, $\nslots=\wtO(\rank/\eps^{3/2}+\rank/\acc^2+\rank^{3/2}/(\acc\eps))$ samples suffice, with high probability, for every unit vector of the estimate to lie within distance $\acc$ of the target subspace (\Cref{thm:subspace_informal}).
The same bound holds on the large-gap spectrum when its $\rank^{th}$ eigenvalue is public, since it sets the ridge.
The bound has no explicit factor of $\dimn$, and its hidden logarithmic factors do not involve $\dimn$ either.
It assumes no bound on the norms of the records, but the geometric spectrum requires $\rank>\lceil\log_2(\dimn)\rceil$.
\Cref{sec:sketching} states these results formally, and \Cref{sec:discussion_related} compares their dependence on $\dimn$ with that of other methods.\footnote{For ease of exposition, we present utility guarantees only for the geometric spectrum, but they can be extended to other distributions where most of the ``energy'' is contained in the top-$\rank$ subspace. To handle such cases, our algorithm requires an additional input bounding the data scale. Fortunately, this scale parameter can be privately estimated from the data via another instantiation of the PFR algorithm (\Cref{sec:trace}).}

\paragraph{Further applications.}
In \Cref{sec:applications} we illustrate the power of the PFR scheme paired with element--block filtering, giving friendly-DP release mechanisms for a variety of applications: (1) releasing a scalar with Gaussian noise scaled to the typical scale of the data despite potential outliers; (2) releasing the KL divergence between user histograms and a uniform reference with Laplace noise; (3) estimating the effective resistances of all pairs of vertices of a resistor network with Gaussian vertex potentials; and (4) releasing a treatment rule selected with the exponential mechanism.

\subsection{Technical Overview}\label{sec:technique}

The proof of \Cref{thm:intro:gen_frnd_core_priv} verifies, in two components, the two requirements that the PFR theorem places on element--block filtering: its joint differential privacy, for an arbitrary friendliness predicate (\Cref{thm:out_reg_jdp}), and, through a concavity argument, that it establishes the condition of the release mechanism (\Cref{sec:PFR}).
The goal is that a number of records linear in the block size $\bsz$ suffices, up to factors that depend only on $\eps$ and $\delta$.

\paragraph{First component: a private filtering step.}
The filtering step compares the \emph{friend count} of each record, the number of reference blocks against which its score is at least the stricter threshold $2\thr$, with an acceptance threshold $\athr$ (\Cref{fig:pfr_framework}b); for the leverage score of subspace estimation, the friend count is the number of block ellipsoids that contain the record (\Cref{fig:friendliness}c).
Changing one record changes one reference block, so the count of every other record changes by at most one, but the test of the changed slot now asks about a different record, so its answer can change arbitrarily.
Protecting all $\nslots$ comparisons by direct composition gives noise whose margin grows as $\sqrt{\nslots}$, so for counts in $[0,\nblk]$ it needs $\nslots=\Omega(\left(\bsz\ln(1/\delta)/\eps\right)^2)$ records, quadratic in the block size (\Cref{sec:filter_privacy}).

The filtering step therefore runs the tests through the sparse vector technique, whose privacy analysis pays only for a few \emph{paid} answers, here the rejections, and it aborts once the rejections reach a cap.
It uses the variant with a fixed public threshold, which follows from the target-charging technique of \citet{cohen2023target}.
With a fixed threshold and independent Laplace noise of scale $\qscale$ on each count, the tests share no random state, so the proof can set aside the exceptional test of the changed record and treat every other test as a query of sensitivity one.

The exceptional test can still add one rejection, so that only one of the executions on $\ds$ and on $\ds^{i\to\nul}$ reaches the cap, where $\ds^{i\to\nul}$ denotes $\ds$ with its $i^{th}$ record replaced by $\nul$.
Laplace noise added to the cap hides this extra rejection.
Together, the two noises make the filtering step jointly differentially private (\Cref{def:jdp,thm:out_reg_jdp}).
We set the cap to order $\ln(1/\delta)/\eps$ rejections, enough to absorb its own noise when no record is expected to fail.
Advanced composition of order $\ln(1/\delta)/\eps$ rejecting tests, each $(1/\qscale,0)$-DP, costs order $\ln(1/\delta)/(\sqrt{\eps}\,\qscale)$, which is at most $\eps$ for $\qscale$ of order $\ln(1/\delta)/\eps^{3/2}$.
The noise of one test exceeds order $\qscale\ln(1/\delta)=\ln^2(1/\delta)/\eps^{3/2}$ only with probability of order $\delta$, and this margin is small compared with the range $[0,\nblk]$ of a friend count once $\nblk=\Omega(\ln^2(1/\delta)/\eps^{3/2})$, that is, once $\nslots=\Omega(\bsz\ln^2(1/\delta)/\eps^{3/2})$.
On the geometric spectrum, with blocks of $\wtO(\rank)$ records, filtering thus costs $\wtO(\rank/\eps^{3/2})$ records.

\paragraph{Second component: concavity.}
The filtering step guarantees only that a retained record is friendly with many separate blocks, whereas the release mechanism needs it to be friendly with the remaining filtered data as a whole.
If the changed record is retained, then, up to the noise of its own test, it passed the test against more than $\athr=\athrVal$ of the blocks, and since the run did not abort, at most order $\ln(1/\delta)/\eps$ records were rejected.
Setting aside the blocks of the rejected records and the block of the changed record leaves at least $\nblk/2$ \emph{intact} blocks, which contain no rejected record and against which the record has score at least $2\thr$.
Concavity, with a nonnegative score against the rest of the filtered data, then gives the record a score of at least $(1/2)\cdot2\thr=\thr$ against the remaining filtered data, which is why the filtering step tests at $2\thr$.
The size-weighted average uses all $\nslots$ slots, since rejected records are nulled rather than deleted.
With the leverage score, a record in the $1/(2\thr)$-ellipsoids of half of the blocks, all of them intact, thus lies in the $1/\thr$-ellipsoid of the pooled retained data.
The PFR theorem then combines the two components: joint differential privacy compares the filtered outputs on the two neighboring inputs, and the friendly DP of the release mechanism pays for the changed record when it is retained.

This argument controls a bad event for one fixed changed slot, which involves only the noise of its own test and of the cap, rather than certifying all the retained records at once, so the privacy bound on $\nblk$ has no $\log(\nslots)$ factor.
A guarantee that every record of typical data is retained must control all the tests, and it pays a factor logarithmic in $\nslots$ (\Cref{cor:pfr_retention}).

\subsection{Further Related Work}\label{sec:related}
The discussion above already relates PFR to previous work on local sensitivity, pDP, PTR, FriendlyCore and sample-and-aggregate.
\Cref{sec:ptr_comparison} compares PFR with FriendlyCore and PTR, and compares the numbers of records that FriendlyCore and element--block filtering need.
Stability-based estimators~\cite{BrownHS23} and the subsample-test-aggregate reduction~\cite{rubinstein2026down} also exploit stability under the deletion of records, through a stabilized estimator or through certified releases on accepted subsamples, which are then aggregated, whereas PFR applies one friendly-DP release mechanism to all the retained records.
General frameworks for black-box statistics estimate a real-valued statistic from its values on subsets of the data, under a Lipschitz condition on the statistic~\cite{linder2025blackbox} or for i.i.d.\ data~\cite{steinke2026blackbox,brown2026monotone}, whereas the output of a friendly-DP release mechanism need not be a scalar.
\Cref{sec:discussion_related} discusses further related work, among them smooth sensitivity, restricted sensitivity and private Gaussian sketching, and compares PFR in more detail with the black-box frameworks, with stable estimators and with subsample-test-aggregate.

%% file: sections/sec_preliminaries.tex
\section{Preliminaries}\label{sec:prelims}

\begin{table}[t]
    \centering\small
    \begin{tabular}{@{}lp{0.82\textwidth}@{}}
        \toprule
        Symbol & Meaning \\
        \midrule
        $\nslots$ & sample size of the full dataset, null records included \\
        $\bsz$ & block size: filter uses reference blocks of size $\bsz$ \\
        $\nblk$ & number of reference blocks, $\nblk\coloneqq\nslots/\bsz$\\
        $\athr$ & acceptance threshold: filter retains a record if its friend count plus Laplace noise exceeds $\athr$ \\
        $\svcap$ & cap: filter aborts once the number of rejected records reaches $\svcap$ plus Laplace noise \\
        $\cscale$, $\qscale$ & Laplace noise scales: $\cscale$ for the cap, and $\qscale$ for the friend counts, which are at most $\nblk$ \\
        $\fscore$ & friendliness score $\fscore(\ds,\cand)\ge0$ of record $\cand$ against dataset $\ds$, higher for smaller influence \\
        $\thr$ & friendliness threshold, $\thr>0$: $\cand$ is $\thr$-friendly with $\ds$ if $\fscore(\ds,\cand)\ge\thr$, and filter tests at $2\thr$ \\
        \bottomrule
    \end{tabular}
    \caption{Recurring notation.
    Friendliness scores are defined in \Cref{def:friendliness}.}\label{tab:notation}
\end{table}

We denote by $\ind\{E\}$ the indicator of an event $E$, and by $\Gauss(\mu,\Sigma)$ and $\Lap(\sigma)$ the Gaussian and centered Laplace distributions respectively.
Asymptotic notation $O$, $\Omega$ and $\Theta$ suppresses universal constant factors, and $\wtO$ and $\wtOmega$ also suppress polylogarithmic factors.
\Cref{app:prelims} collects the remaining notational conventions, along with further tools, such as R\'enyi divergence and concentration bounds.

We augment the data domain with a \emph{null} element $\nul$ that serves as a placeholder, and denote the augmented domain by $\dom$.
A dataset $\ds=(\ds_1,\ldots,\ds_\nslots)\in\dom^\nslots$ is a sequence of $\nslots$ records, each occupying a \emph{slot}, and $\dom^*$ is the set of datasets of all sizes, including the empty dataset $\emptyset$.
If an individual's record is withheld, one can think of this as replacing their record with the placeholder instead of deleting it.
The size $|\ds|=\nslots$ includes null records, and $(\ds,\ds')$ denotes the concatenation of two datasets.
We denote by $\ds^{I\to\nul}$ the dataset obtained from $\ds$ by \emph{nulling} the records indexed by $I\subseteq[\nslots]\coloneqq\{1,\ldots,\nslots\}$, that is, by replacing each with $\nul$, and write $\ds^{i\to\nul}$ when $I=\{i\}$.
Under \emph{zero-out} adjacency~\cite[Definition~1.1]{kairouz2021practical}, two datasets are \emph{neighbors}, denoted $\ds\simeq\ds'$, if they are $\ds$ and $\ds^{i\to\nul}$ for some $i\in[\nslots]$.
Neighbors therefore have the same number of slots, which is public.
This adjacency is a common convention in private learning, with guarantees essentially equivalent to those of the add/remove notion~\cite[Section~2.1.1]{ponomareva2023dpfy}.

We state differential privacy through the indistinguishability of two random variables.

\begin{definition}[Differential privacy \Cite{dwork2006calibrating}]\label[definition]{def:dp}
    Given $\eps\ge0$ and $\delta\in[0,1]$, we say that two random variables $\mathbf{X}$ and $\mathbf{Y}$ over a common domain are \emph{$(\eps,\delta)$-indistinguishable}, and denote it by $\mathbf{X}\approx_{\eps,\delta}\mathbf{Y}$, if $\Pr[\mathbf{X}\in O]\le e^{\eps}\Pr[\mathbf{Y}\in O]+\delta$ and $\Pr[\mathbf{Y}\in O]\le e^{\eps}\Pr[\mathbf{X}\in O]+\delta$ for every event $O$.
    A randomized mechanism $\mech$ will be called \emph{$(\eps,\delta)$-differentially private} (\emph{DP}) if $\mech(\ds)\approx_{\eps,\delta}\mech(\ds')$ for every pair of neighboring datasets $\ds\simeq\ds'$.
\end{definition}

For simplicity, we state all our privacy guarantees in the $(\eps,\delta)$ form of \Cref{def:dp}.
Many of the intermediate steps can also be analyzed under other notions of privacy, such as R\'enyi DP~\cite{mironov2017renyi} or zero-concentrated DP~\cite{bun2016concentrated} (\Cref{def:renyi}).
Our guarantees, however, hold only for $\delta>0$, so they do not extend to pure DP, that is, to $\delta=0$.
Since replacing a record takes two zero-out steps through a null slot, every $(\eps,\delta)$-DP mechanism with $\eps\le1$ is also private with respect to replacement, with parameters larger by constant factors (\Cref{claim:zero_out_replacement}).

We rely on the properties of group privacy~\cite{dwork2014algorithmic}, post-processing~\cite{dwork2014algorithmic}, and various composition theorems of differential privacy described in \Cref{app:prelims}.

The \emph{sensitivity} of a real-valued function $\qry$ on datasets is the supremum of $\vert \qry(\ds)-\qry(\ds')\vert$ over all pairs of neighboring datasets $\ds\simeq\ds'$.
The Gaussian mechanism $\Gauss_\sigma(\stat)$, which adds $\Gauss(0,\sigma^2)$ to a real-valued function $\stat$ of sensitivity at most $1$, is $(\eps,\gaussprof(\eps))$-DP for every $\eps>0$, where $\gaussprof(\eps)\coloneqq\Phi(1/(2\sigma)-\eps\sigma)-e^{\eps}\Phi(-1/(2\sigma)-\eps\sigma)$ is its privacy profile~\cite[Theorem~8]{balle2018improving} and $\Phi$ is the standard Gaussian CDF (\Cref{fact:gauss_profile}).
The Laplace mechanism~\cite{dwork2006calibrating}, which adds $\Lap(\sigma)$ to such a function, is $(1/\sigma,0)$-DP, and its noise exceeds $a\ge0$ in absolute value with probability exactly $e^{-a/\sigma}$ (\Cref{fact:laplace}).

%% file: sections/sec_filtering.tex
\section{Private Filtering}\label{sec:outlier_rej}

This section constructs the filtering step of PFR, its first internal stage: a private filter that nulls the records that are friendly with too few blocks of a random partition of the dataset, and aborts if there are too many such records.
PFR never publicly releases the output of the filter, so the privacy that the PFR theorem (\Cref{lem:friendly_comp}) asks of it is only a joint form of differential privacy (\Cref{def:jdp}).
Omitted proofs can be found in \Cref{app:outlier-rejection}.

\subsection{The Filter}\label{sec:algorithm}

\begin{figure}[t]
\begin{minipage}[t]{0.525\textwidth}
\vspace{0pt}
\begin{ruledalg}{Private filtering $\filt_{\fpar}(\pred;\ds)$}\label{alg:priv_filt}
    \begin{algorithmic}
        \Require Dataset $\ds = (\ds_1,\ldots,\ds_\nslots) \in \dom^{\nslots}$, friendliness predicate $\pred$, block size $\bsz \in \bbN$, acceptance threshold $\athr\ge0$, failure cap $\svcap\ge0$, noise scales $\cscale, \qscale > 0$
        \State $\nblk \gets \nslots/\bsz$
        \State $\ncap \gets \lceil \svcap + \cnoise \rceil$ where $\cnoise \sim \Lap(\cscale)$
        \State \algorithmicif\ $\ncap \le 0$ \algorithmicthen\ \Return $\nul^{\nslots}$
        \State $\ds^{(1)}, \ldots, \ds^{(\nblk)} \gets \text{partition}_{\bsz}(\ds)$
        \State $\qry_{i}(\cdot) \coloneqq \sum_{j \in [\nblk]} \pred\left(\cdot^{(j)}, \ds_{i}\right)$ for $i \in [\nslots]$
        \State $\br \gets \SV_{\athr,\ncap,\qscale}(\qry; \ds)$
        \State \algorithmicif\ $\Vert \br \Vert = \ncap$ \algorithmicthen\ \Return $\nul^{\nslots}$
        \State \Return $\fds \gets \ds^{R \to \nul}$ where $R \coloneqq \left\{i : \tbit_{i} = 1\right\}$
    \end{algorithmic}
\end{ruledalg}
\end{minipage}\hfill
\begin{minipage}[t]{0.455\textwidth}
\vspace{0pt}
\begin{ruledalg}{Fixed-threshold sparse vector $\SV_{\athr,\svcap,\svscale}(\qry; \ds)$}\label{alg:sparse_vec}
	\begin{algorithmic}
		\Require Dataset $\ds$, queries $\qry_{1}, \ldots, \qry_{\nslots}$, acceptance threshold $\athr \in \bbR$, failure cap $\svcap \in \bbN$, noise scale $\svscale > 0$
		\State $\br \gets ()$
		\For{$i \in [\nslots]$}
		\State $\qnoise_{i} \sim \Lap(\svscale)$
		\State $\tbit_{i} \gets \ind\left\{\qry_{i}(\ds) + \qnoise_{i} \le \athr\right\}$
		\Comment{failure}
		\State $\br \gets (\br, \tbit_{i})$
		\State \algorithmicif\ $\Vert \br \Vert = \svcap$ \algorithmicthen\ \Return $\br$
		\EndFor
		\State \Return $\br$
	\end{algorithmic}
\end{ruledalg}
\end{minipage}
\end{figure}
The filter tests each record through an input friendliness predicate, which specifies when a record is typical of a dataset.

\begin{definition}[Friendliness predicate]\label[definition]{def:set_friendliness}
	A \emph{friendliness predicate} is a function $\pred : \dom^{*} \times \dom \rightarrow \{0,1\}$.
	We say that a record $\cand$ is \emph{friendly} with a dataset $\ds$ if $\pred(\ds,\cand)=1$.
\end{definition}

Our element--block filter $\filt_{\fpar}(\pred;\cdot)$ (\Cref{alg:priv_filt}) splits the $\nslots$ slots into $\nblk=\nslots/\bsz$ \emph{reference blocks} of $\bsz$ slots each, and judges each record by its \emph{friend count} $\cnt_i$, the number of blocks that it is friendly with.
Counting blocks keeps the counts stable: changing one record changes only the block that contains it, so the friend count of every other record moves by at most one.
The blocks have equal sizes, so we assume throughout that $\bsz$ divides $\nslots$.\footnote{For other sizes, dropping the last $\nslots-\bsz\lfloor\nslots/\bsz\rfloor<\bsz$ records is a public preprocessing step that preserves privacy, and our guarantees then apply with the remaining $\bsz\lfloor\nslots/\bsz\rfloor$ records as the input, so that their targets and retention guarantees concern these records only.}
The operation $\text{partition}_{\bsz}(\ds)$ draws a uniformly random partition of the slot indices $[\nslots]$ into $\nblk$ subsets of size $\bsz$, and outputs the blocks $\ds^{(1)},\ldots,\ds^{(\nblk)}$, where $\ds^{(j)}$ holds the records in the $j^{th}$ subset in their original index order.
The partition is drawn independently of the data, so the runs on two neighboring inputs, which have the same number of slots, can use a common partition.

The record $\ds_i$ fails its test if $\cnt_i$ plus fresh Laplace noise of scale $\qscale$ is at most the acceptance threshold $\athr$.
The filter runs these tests through fixed-threshold sparse vector (\Cref{alg:sparse_vec}, analyzed in \Cref{prop:sparse_vec_priv}), whose privacy cost grows with the number of failed tests rather than with the number of tests.
The filter therefore caps the failures: once their number reaches a cap, it aborts and returns the all-null dataset $\nul^\nslots$ instead of paying for every failure on an input with many outliers.
The test of the changed record is the exception to the stability of the counts: on two neighboring inputs it asks about two different records, so its answer may flip.
With a fixed cap, this one answer could decide whether the run aborts.
Laplace noise of scale $\cscale$ added to the cap $\svcap$ masks this shift of one failure.
If the run completes, the filter nulls the records that failed rather than deleting them, so every retained record keeps its slot.
The values of the predicate at $\nul$ are unrestricted, since they do not affect privacy; on inputs with null slots, they decide only whether these slots count toward the cap.
The parameters of the filter are omitted from the notation when they are clear from the context.
\Cref{tab:notation} lists them together with the other recurring symbols.

The noisy cap $\ncap$ and the transcript $\br$ of the tests, including its length, are internal.
The privacy theorems (\Cref{thm:out_reg_jdp,thm:gen_frnd_core_priv}) set the cap to $\svcap=2\cscale\Ldef$, which does not grow with the sample size, because we analyze data in which no record is expected to fail.
This value is twice a high-probability bound on the noise of the cap, so a run in which no test fails aborts only with probability $O(\delta^2)$, and half of the cap leaves room for a few failing records (\Cref{thm:out_reg_acc}).
A run that does not stop early evaluates the predicate $\nslots\cdot\nblk$ times: the sparse vector technique reduces the privacy cost of these comparisons, and not their number.

\subsection{Fixed-Threshold Sparse Vector}\label{sec:sparse_vector}

The privacy analysis of the filter (\Cref{thm:out_reg_jdp}) builds on the privacy of its tests, the run of sparse vector on the friend counts.
The sparse vector technique~\cite{dwork2009complexity,roth2010interactive,hardt2010multiplicative,vadhan2017complexity} answers threshold queries with a privacy cost governed by the number of positive responses, rather than the total number of queries.
We use a fixed public threshold, with independent Laplace noise added only to the query values, which is a direct application of the target-charging technique of \citet[Algorithm 1]{cohen2023target} to private tests.
The tests then share no random state, so given the preceding transcript, each answer depends only on its own query and its own noise.
A query that changes arbitrarily, as the query of the changed slot does in the filter, therefore affects only its own bit, unlike in the variant of \citet[Algorithm~2]{dwork2014algorithmic}, which redraws its threshold noise after each positive response.
\Cref{alg:sparse_vec} releases one bit per query: a noisy query value of at most $\athr$ is a \emph{failure}, encoded by $1$, and the algorithm stops at the $\svcap^{th}$ failure or after all the queries have been answered.
For a binary transcript $\br=(\tbit_1,\tbit_2,\ldots)$, we denote by $\vert\br\vert$ its length and by $\Vert\br\Vert\coloneqq\sum_{j\in[\vert\br\vert]} \tbit_j$ the number of failures in it.
The queries in the next proposition are \emph{adaptive}: the choice of each query $\qry_i$ may depend on the preceding transcript $(\tbit_1,\ldots,\tbit_{i-1})$, but not otherwise on the dataset.
In the filter, the queries also depend on the input, which the proof of \Cref{thm:out_reg_jdp} handles separately.

\def\SVProposition{
	Given $\nslots,\svcap\in\bbN$; $\delta \in (0,1)$; $\eps\in(0,4\ln(2/\delta)]$; a threshold $\athr\in\bbR$ and adaptive queries $\qry_{1},\ldots,\qry_{\nslots}$ of sensitivity at most $1$, if $\svcap \ge 12\ln(2/\delta)$, then $\SV_{\athr,\svcap,\svscale}$ (\Cref{alg:sparse_vec}) is $(\eps, \delta)$-DP, where $\svscale \coloneqq \svscaleVal$.
}
\begin{proposition}[Privacy of fixed-threshold sparse vector]\label[proposition]{prop:sparse_vec_priv}
	\SVProposition
\end{proposition}

\begin{proof}[Proof outline]
	Conditional on the preceding transcript, the next bit is a threshold post-processing of a sensitivity-one Laplace mechanism.
	By a lemma of \citet{cohen2023target} (\Cref{lem:cl_boolean_not_prior}), its two possible distributions are mixtures, with a common weight, of the constant $0$ and of a pair of $(1/\svscale,0)$-indistinguishable distributions that output a failure with probability at least $1/(e^{1/\svscale}+1)\ge4/9$.
	We call a draw from the second component a \emph{private call}.
	A multiplicative Chernoff bound (\Cref{prop:multiplicative_chernoff}) shows that the run reaches its $\svcap^{th}$ failure within $W\le5\svcap$ private calls, except with probability $e^{-\svcap/4}$, which is at most $\delta/2$ since $\svcap\ge12\ln(2/\delta)$.
	Advanced composition (\Cref{fact:advanced_composition}) over these calls, with $\delta_0=\delta/2$, bounds the privacy loss by $W/(2\svscale^2)+\sqrt{2W\ln(2/\delta)}/\svscale$, which the stated $\svscale$ and $\eps\le4\ln(2/\delta)$ bring below $(10/49+\sqrt{10}/7)\eps<\eps$ (\Cref{prop:sparse_vec_priv_restated}).
\end{proof}

\subsection{Privacy and Retention}\label{sec:filter_privacy}

\paragraph{Privacy.}
Since a filter rejects a record by nulling it, the retained records keep their slots, and the privacy notion that we require compares the outputs on neighboring inputs only outside the changed slot.

\begin{definition}[Filter and joint differential privacy]\label[definition]{def:jdp}
	A randomized function $\filt : \dom^{*} \rightarrow \dom^{*}$ will be called a \emph{filter} if, given a dataset, it returns the same dataset up to replacing a subset of its records with $\nul$; that is, for every $\ds\in\dom^*$ we have $\filt(\ds)=\ds^{I \to \nul}$ for some random $I \subseteq [\vert \ds \vert]$.
	A filter will be called \emph{$(\eps, \delta)$-jointly differentially private} (\emph{JDP}) if for any $\nslots\in\bbN$, $\ds \in \dom^{\nslots}$ and $i \in [\nslots]$ we have $\filt(\ds)^{i \to \nul} \approx_{\eps, \delta} \filt(\ds^{i \to \nul})$.
\end{definition}

This is the notion of \citet{kearns2014jdp}, specialized to zero-out adjacency.
It compares the two filtered datasets only after slot $i$ is nulled in both, so the filter may reveal whether it retained $\ds_{i}$; a mechanism applied to its output pays for that record separately, either through ordinary privacy (\Cref{cor:jdp_comp}) or, in PFR, through friendly DP (\Cref{lem:friendly_comp}).
The noise scales that make the filter jointly differentially private do not depend on the number of records.

\def\OutlierJDPTheorem{
    Given $\nslots,\bsz\in\bbN$; $\eps\in(0,1]$; $\delta \in (0,1)$; $\athr\ge0$; and a friendliness predicate $\pred$, the filter $\filt_{\fpar}(\pred;\cdot)$ (\Cref{alg:priv_filt}) on datasets of $\nslots$ records is $(\eps, \delta)$-JDP, where $\svcap\coloneqq2\cscale\Ldef$, $\cscale \coloneqq \cCapJDP/\eps$ and $\qscale \coloneqq (\cCapJDP/\eps)\sqrt{\lceil \cBoundJDP\Ldef/\eps\rceil \ln(4/\delta)}$.
}
\begin{theorem}[Privacy of the filter]\label{thm:out_reg_jdp}
    \OutlierJDPTheorem
\end{theorem}

\begin{proof}[Proof outline]
    Fix $\ds$ and $i$, condition on the common partition, and compare the outputs on $\ds$ and $\ds^{i\to\nul}$, with slot $i$ nulled in both.
    The comparison pays separately for the two effects of the change described in \Cref{sec:algorithm}, first through the test noise and then through the cap noise.
    With the queries fixed, \Cref{prop:sparse_vec_priv} applies with the deterministic cap $\lceil\cBoundJDP\Ldef/\eps\rceil$, which bounds $\ncap$ except with probability $\delta/2$, and gives an $(\eSVJDP,\delta)$ indistinguishability.
    With the input fixed, only the test of slot $i$ changes, and the noise of the cap gives an $(\eCapJDP,0)$ indistinguishability (\Cref{thm:out_reg_jdp_restated}).
\end{proof}

A jointly differentially private filter composes with any differentially private mechanism (\Cref{cor:jdp_comp_restated}).

\def\JDPCompositionCorollary{
    Given $\eps_1,\eps_2\ge0$; $\delta_1,\delta_2\in[0,1]$; a mechanism $\mech$ that is $(\eps_{1}, \delta_{1})$-DP and a filter $\filt$ that is $(\eps_{2}, \delta_{2})$-JDP, the composition $\mech \circ \filt$ is $(\eps_{1} + \eps_{2}, \delta_{1} + \delta_{2})$-DP.
}
\begin{corollary}[Composition with a private mechanism]\label[corollary]{cor:jdp_comp}
    \JDPCompositionCorollary
\end{corollary}

Protecting all $\nslots$ comparisons directly, as the zCDP filter of FriendlyCore~\cite{FriendlyCore22} does for element--element friendliness, gives each test a margin of order $\sqrt\nslots\,\ln(1/\delta)/\eps$.
This margin is small compared with the range $[0,\nblk]$ of the counts once $\nslots=\Omega(\left(\bsz\ln(1/\delta)/\eps\right)^2)$.
Calibrating the test noise to the cap instead gives the margin $O(\ln^2(1/\delta)/\eps^{3/2})$, which is small once $\nslots=\Omega(\bsz\ln^2(1/\delta)/\eps^{3/2})$, the order of \Cref{thm:gen_frnd_core_priv}.

\paragraph{Retention.}\label{sec:filter_utility}

The next theorem describes when the filter retains the data instead of aborting, and what every retained record satisfies.

\def\OutlierAccuracyTheorem{
    Given $\nslots,\bsz\in\bbN$; $\fprob \in (0,1)$; $\athr,\svcap\ge0$; $\cscale,\qscale>0$; a dataset $\ds \in \dom^{\nslots}$ and a friendliness predicate $\pred$, with probability at least $1 - \fprob$ over the noises of the filter $\filt_{\fpar}(\pred;\ds)$ (\Cref{alg:priv_filt}), conditionally on any realization of its partition, the following hold simultaneously, where $\cnt_{i}$ is the number of blocks of the realized partition that $\ds_{i}$ is friendly with, $\qmargin \coloneqq \qscale \ln(2\nslots/\fprob)$ and $\cmargin \coloneqq \cscale \ln(2/\fprob)$.
    \begin{enumerate}
        \item \emph{Retention:} If $\left\vert \{i : \cnt_{i} \le \athr + \qmargin\} \right\vert < \svcap - \cmargin$, then the run does not abort and retains every $\ds_{i}$ with $\cnt_{i} > \athr + \qmargin$.
        \item \emph{Quality of retained records:} Every retained record $\ds_i$ satisfies $\cnt_{i} > \athr - \qmargin$.
    \end{enumerate}
}
\begin{theorem}[Retention guarantee]\label{thm:out_reg_acc}
    \OutlierAccuracyTheorem
\end{theorem}

\begin{proof}
    Pre-sample all $\nslots$ test noises, including those that an execution that reaches its cap would not use.
    By \Cref{fact:laplace} and a union bound, the event $E$ that $|\qnoise_j|\le\qmargin$ for every $j$ and $|\cnoise|\le\cmargin$ has probability at least $1-\fprob$, since $\Pr[|\qnoise_j|>\qmargin]=\fprob/(2\nslots)$ and $\Pr[|\cnoise|>\cmargin]=\fprob/2$, and since the noises are independent of the partition, this bound also holds conditionally on the partition.
    On $E$, a record with $\cnt_i>\athr+\qmargin$ passes its noisy test, so under the retention hypothesis fewer than $\svcap-\cmargin\le\svcap+\cnoise$ tests fail, the run does not abort, and it retains all such records.
    Conversely, a retained record passed its noisy test, so $\cnt_i+\qnoise_i>\athr$, which gives $\cnt_i>\athr-\qmargin$ on $E$.
\end{proof}

If, with high probability, almost all of the population is friendly with a random reference block, then an i.i.d.\ sample passes through the filter without aborting and with few rejections; \Cref{cor:out_reg_dist_acc} gives the precise condition.

%% file: sections/sec_pfr.tex
\section{Propose-Filter-Release}\label{sec:PFR}

This section defines PFR and states its general guarantee: a release mechanism that is private only on suitably \enquote{friendly} neighboring datasets, preceded by a filtering step that is a jointly private filter and establishes friendliness for the changed record, is differentially private in the standard sense.
It then instantiates PFR and its guarantee with element--block filtering, the filter of \Cref{sec:outlier_rej}, which guarantees only that a retained record is friendly with many individual reference blocks.
We therefore derive the friendliness predicate from a concave score, which combines these blockwise guarantees into friendliness with the remaining filtered data, as the release mechanism requires.
The proofs that are not given here can be found in \Cref{app:generalized-friendly-core}.

\subsection{Friendly DP and the PFR Theorem}\label{sec:pfr_theorem}

The release mechanism of PFR needs to be private only at the neighboring pairs that a friendliness predicate (\Cref{def:set_friendliness}) admits.

\begin{definition}[Friendly DP]\label[definition]{def:friendly_dp}
    Given $\eps\ge0$; $\delta\in[0,1]$; and a friendliness predicate $\pred$, a mechanism $\mech$ will be called \emph{$(\eps, \delta)$-friendly-DP with respect to $\pred$} if $\mech(\ds^{i \to \nul}) \approx_{\eps, \delta} \mech(\ds)$ for every $\nslots\in\bbN$, $\ds \in \dom^{\nslots}$ and $i \in [\nslots]$ such that $\ds_i\ne\nul$ and $\pred\left(\ds^{i \to \nul}, \ds_{i}\right) = 1$.
\end{definition}

Friendly DP with respect to $\pred$ is thus per-instance differential privacy (pDP)~\cite{wang2019perinstance} at every admitted pair, with the same $\eps$ and $\delta$ for all of them, where the instance consists of the remaining data and the changed record.
\citet{wang2019perinstance} adds the record to the dataset, whereas zero-out adjacency places it in a null slot.

PFR precedes the release mechanism with a filtering step, which makes the changed record friendly with the remaining filtered data, for each slot separately.

\begin{definition}[Propose-filter-release]\label[definition]{def:PFR}
    Given a friendliness predicate $\pred$, a filter $\filt$ (\Cref{def:jdp}) that retains each record, except with small probability for that record, only if the record is friendly with the remaining filtered data under $\pred$, and a mechanism $\mech$ that is friendly-DP with respect to $\pred$, \emph{PFR} with filtering step $\filt$ and release mechanism $\mech$ is the mechanism $\mech\circ\filt$.
\end{definition}

What PFR proposes is the friendliness predicate, which links the filtering step to the release mechanism.
The PFR theorem quantifies the two requirements of \Cref{def:PFR} and adds a third, that the filtering step is jointly differentially private.

\def\FriendlyCompositionLemma{
    Given $\nslots\in\bbN$; $\eps_1,\eps_2\ge0$; $\delta_1,\delta_2,\delta_3\in[0,1]$; a friendliness predicate $\pred$ and PFR $\mech\circ\filt$ with respect to $\pred$ (\Cref{def:PFR}), if $\filt$ is $(\eps_1,\delta_1)$-JDP and $\mech$ is $(\eps_2,\delta_2)$-friendly-DP with respect to $\pred$, both on datasets of $\nslots$ records, and for every $\ds\in\dom^\nslots$ and $i\in[\nslots]$ with $\ds_i\ne\nul$ the probability that $\filt(\ds)$ retains $\ds_i$ while $\pred(\filt(\ds)^{i\to\nul},\ds_i)=0$ is at most $\delta_3$, then $\mech\circ\filt$ is $(\eps_1+\eps_2,\,\delta_1+\delta_2+e^{\eps_1+\eps_2}\delta_3)$-DP on datasets of $\nslots$ records.
}
\begin{theorem}[PFR theorem]\label{lem:friendly_comp}
    \FriendlyCompositionLemma
\end{theorem}

\begin{proof}[Proof outline]
    Fix neighboring datasets $\ds$ and $\ds'\coloneqq\ds^{i\to\nul}$, where $\cand\coloneqq\ds_i\ne\nul$.
    Let $A\coloneqq\filt(\ds)^{i\to\nul}$ and $A'\coloneqq\filt(\ds')$.
    Thus $A$ records the retained data outside slot $i$, and joint privacy gives $A\approx_{\eps_1,\delta_1}A'$.

    For the analysis, define an auxiliary release $\widetilde Y$ by additionally nulling slot $i$ in $\filt(\ds)$ whenever $\pred(A,\cand)=0$, before applying $\mech$.
    Conditioning on $A=y$ means fixing the retained data outside slot $i$; whether the filter retained $\cand$ may still be random.
    If $\pred(y,\cand)=0$, the auxiliary release $\widetilde Y$ has law $\mech(y)$.
    If $\pred(y,\cand)=1$, its law is a mixture of $\mech(y)$ and $\mech(y^{i\to\cand})$, where $y^{i\to\cand}$ restores $\cand$ to slot $i$.
    Friendly DP makes the two mixture components indistinguishable, so in either case
    \[
        (\widetilde Y\mid A=y)\approx_{\eps_2,\delta_2}\mech(y).
    \]

    The release on $\ds'$ is $Y'\coloneqq\mech(A')$, whose law given $A'=y$ is $\mech(y)$.
    Two-stage composition (\Cref{prop:two_stage}), with $A,A'$ as the first stage and $\widetilde Y, Y'$ as the second, gives
    \[
        \widetilde Y\approx_{\eps_1+\eps_2,\,\delta_1+\delta_2}Y'.
    \]

    Under shared filter and release randomness, the actual release $Y\coloneqq\mech(\filt(\ds))$ agrees with $\widetilde Y$ unless the filter retains an unfriendly $\cand$, an event of probability at most $\delta_3$.
    Replacing $\widetilde Y$ by $Y$ therefore adds at most $\delta_3$ to the privacy inequality from $Y$ to $Y'$, and $e^{\eps_1+\eps_2}\delta_3$ to the reverse inequality, proving the claim.
\end{proof}

When $\mech$ is private on all neighboring pairs, it is friendly-DP with respect to the predicate that admits every non-null record, the last hypothesis holds with $\delta_3=0$, and the theorem reduces to \Cref{cor:jdp_comp}.
Any filter that satisfies the hypotheses of the theorem for some predicate can serve as the filtering step of PFR, and the zCDP filter of FriendlyCore and the test of PTR are examples (\Cref{sec:ptr_comparison}).

\subsection{PFR with Element--Block Filtering}\label{sec:friendliness}

In the instantiation of this work, the predicate of PFR is induced by a threshold on a friendliness score, which quantifies how typical a record is of a dataset.

\begin{definition}[Score-based friendliness]\label[definition]{def:friendliness}
    A \emph{friendliness score} is a function $\fscore : \dom^* \times \dom \rightarrow [0,\infty]$ that, at every non-null record $\cand\in\dom$, is invariant under permutations of its first argument and satisfies $\fscore(\emptyset,\cand)=0$.
    Given a friendliness threshold $\thr > 0$, the friendliness predicate \emph{induced} by $\fscore$ is $\pred_{\thr}(\fscore;\ds,\cand)\coloneqq\ind\{\fscore(\ds,\cand)\ge\thr\}$, and we say that $\cand$ is \emph{$\thr$-friendly} with $\ds$ if $\pred_{\thr}(\fscore;\ds,\cand)=1$.
    The score $\fscore$ will be called \emph{concave} if $\fscore((\ds, \ds'), \cand)\ge(|\ds|\,\fscore(\ds,\cand) + |\ds'|\,\fscore(\ds',\cand))/(|\ds| + |\ds'|)$ for any nonempty datasets $\ds$ and $\ds'$ and non-null record $\cand \in \dom$.
\end{definition}

A higher score means a smaller influence of the record on a release.
Since $\thr>0$, no non-null record is friendly with the empty dataset; null records of a nonempty reference dataset are included in its size.\footnote{When we define a score only for nonempty datasets and non-null records, its value on the empty dataset is $0$, and its values at $\nul$ remain unrestricted.}
For a predicate induced by a friendliness score, the condition of friendly DP concerns only the changed record: the remaining data may contain other outliers, and the release mechanism must be private on such pairs as well.

For a real-valued function $\stat$, the influence of a record at a neighboring pair can be measured by the \emph{per-instance sensitivity}~\cite{wang2019perinstance} $|\stat(\ds)-\stat(\ds^{i\to\nul})|$.
If $\psens(\ds^{i\to\nul},\ds_i)$ bounds this sensitivity at every neighboring pair and the score is $\fscore(\ds,\cand)\coloneqq1/(|\ds|\,\psens(\ds,\cand))$, then every admitted pair has per-instance sensitivity at most $1/(\thr\nslots)$.\footnote{Both members of a pair have $\nslots$ slots, so a score normalized by $|\ds|$ is normalized by the same $\nslots$ on both sides.}
Friendly DP then asks for pDP only at pairs of bounded per-instance sensitivity, which additive noise calibrated to $1/(\thr\nslots)$ provides (\Cref{thm:PFR_DP}).

Element--element friendliness gives a score with equality in the concavity condition: given a predicate $\ppred$ on pairs of records, the score $\fscore_\ppred(\ds,\cand)\coloneqq\frac{1}{|\ds|}\sum_{j\in[|\ds|]}\ppred(\cand,\ds_j)$ averages $\ppred(\cand,\cdot)$ over the records of $\ds$.
Its test $\pred_{1}(\fscore_\ppred;(y),\cand)=\ppred(\cand,y)$ against a singleton block recovers the test of FriendlyCore~\cite{FriendlyCore22}.

Concavity is the property that combines friendliness with separate blocks.
If $\nint$ of the $\nblk$ equal-size blocks of a dataset satisfy $\fscore(\ds^{(j)},\cand)\ge\thr'$, then permutation invariance, repeated concavity, and $\fscore\ge0$ on the remaining blocks give the lower bound $(\nint/\nblk)\thr'$ on the score against the whole dataset, which is the aggregation step in the proof of \Cref{thm:gen_frnd_core_priv}.
When $\fscore(\ds,\cand)$ depends on $\ds$ only through its empirical distribution, the concavity condition says that $\fscore$ is concave along mixtures of empirical distributions, which is the reason for the name.

Given a friendliness threshold $\thr>0$, a friendliness score $\fscore$ and a release mechanism $\mech$, \emph{PFR with element--block filtering} is the composition $\PFR_{\fpar}(\mech,\fscore,\thr)\coloneqq\mech\circ\filt_{\fpar}\left(\pred_{2\thr}(\fscore)\right)$ (\Cref{alg:priv_filt}).
Under the conditions of \Cref{thm:gen_frnd_core_priv}, $\PFR_{\fpar}(\mech,\fscore,\thr)$ is PFR with respect to $\pred_{\thr}(\fscore)$, the predicate of the release mechanism, although the filter tests $\pred_{2\thr}(\fscore)$.

The factor two lets the analysis set aside half of the blocks: the block of a retained record, those that the record is not $2\thr$-friendly with, which are few up to the noise of its test, and those of the few records that the filter can reject without aborting.
The other blocks are \emph{intact}, with none of their records rejected, and the aggregation step above gives the record a score of at least $(1/2)\cdot2\thr=\thr$ against the remaining filtered data.
Our main result makes this argument quantitative to bound $\delta_3$ in the PFR theorem (\Cref{lem:friendly_comp}): for a suitable choice of the parameters of the filter, PFR with element--block filtering is differentially private on sufficiently large datasets whenever $\fscore$ is concave and $\mech$ is friendly-DP with respect to $\pred_{\thr}(\fscore)$.

\def\GeneralizedFriendlyCoreTheorem{
    Given $\nslots,\bsz\in\bbN$ with $\nslots/\bsz\in\bbN$; $\eps \in (0,1]$; $\delta \in (0,1)$; a friendliness threshold $\thr > 0$, a concave friendliness score $\fscore$, and a release mechanism $\mech$ that is $(\eRel, \dRel)$-friendly-DP with respect to $\pred_{\thr}(\fscore)$ on datasets of $\nslots$ records, if $\nslots\ge\nPriv$, then $\PFR_{\fpar}(\mech,\fscore,\thr)=\mech\circ\filt_{\fpar}\left(\pred_{2\thr}(\fscore)\right)$ is $(\eps,\delta)$-DP on datasets of $\nslots$ records, where $\Lconst\coloneqq\Ldef$, $\athr\coloneqq\athrValNB$, $\svcap\coloneqq2\cscale\Lconst$, $\cscale\coloneqq\cscaleVal$ and $\qscale\coloneqq\qscaleVal$.
}
\begin{theorem}[PFR with element--block filtering]\label{thm:gen_frnd_core_priv}
    \GeneralizedFriendlyCoreTheorem
\end{theorem}

We denote by $\PFRd$ the mechanism $\PFR_{\fpar}$ with the parameters of \Cref{thm:gen_frnd_core_priv}, which are the ones that we use in the applications.
With these parameters, the filter aborts once it has rejected about $\svcap=\capOffset$ records, and $\PFRd$ then releases $\mech(\nul^\nslots)$ without reporting the abort, so its accuracy guarantees count the runs that abort as inaccurate.
The condition of the theorem involves only public quantities, and the number of blocks that it requires does not grow with $\nslots$, since the argument controls only the noise of the changed slot and of the cap.
The constant $\cBlk$ was not optimized.
The order $\ln^2(1/\delta)/\eps^{3/2}$ is that of the margin of the tests (\Cref{thm:out_reg_jdp,thm:out_reg_acc}), whose Laplace noise is calibrated to a cap of order $\ln(1/\delta)/\eps$.
Whether a cap that does not grow with $1/\eps$ can lower it to order $\ln^2(1/\delta)/\eps$ is open (\Cref{sec:limitations}).
The subset-extension wrapper of \citet{linder2025blackbox} deletes at most order $\ln(1/\delta)/\eps$ records, but it evaluates a different oracle, the statistic itself, on $\nslots^{O(\ln(1/\delta)/\eps)}$ datasets.

\begin{proof}[Proof outline]
    We apply \Cref{lem:friendly_comp} with $\eps_1=\eFilt$, $\eps_2=\eRel$, $\delta_1=\dFilt$, $\delta_2=\dRel$ and $\delta_3=\dFriend$, for which the total additive error is $(25+e^\eps)\delta/32<\delta$.
    With the stated noise scales, \Cref{prop:sparse_vec_priv} gives a $(5\eps/12,\delta/2)$ guarantee for fixed queries and cap $\lceil\cBound\Lconst/\eps\rceil$.
    The two comparisons of the proof of \Cref{thm:out_reg_jdp} then make the filter $(\eFilt,\dFilt)$-JDP, after adding the cap-overflow probability $\dCap$ and the exceptional-query cost $1/\cscale=\eps/\cCap$.
    It remains to bound $\delta_3$, which concerns the changed record only, by making the concavity argument above quantitative.
    Let $E_i$ be the event that $\qnoise_i\le\qscale\Lconst_{32}$ and $\cnoise\le\cscale\Lconst_{32}$, where $\Lconst_{32}\coloneqq\ln(32/\delta)$.
    By the one-sided Laplace tails, $E_i$ has probability at least $1-\dFriend$.
    On $E_i$, a retained $\cand\coloneqq\ds_i$ has $\cnt_i>\athr-\qscale\Lconst_{32}$, and fewer than $\svcap+\cscale\Lconst_{32}$ records were rejected.
    By the lower bound on $\nslots$, setting aside their blocks and the block of $\cand$ still leaves $\nint\ge\nblk/2$ intact blocks that $\cand$ is $2\thr$-friendly with.
    The concavity bound $\fscore(\fds^{i\to\nul},\cand)\ge(\nint/\nblk)\,2\thr\ge\thr$ uses the full number $\nslots$ of slots in the denominator, since rejected records are nulled and not deleted (\Cref{thm:gen_frnd_core_priv_restated}).
\end{proof}

\paragraph{Retention.}
Retention follows from \Cref{thm:out_reg_acc} with the parameters of \Cref{thm:gen_frnd_core_priv}, under a larger lower bound on $\nblk$ that controls all $\nslots$ test noises at once (\Cref{cor:pfr_retention}).
If $\nblk\ge\mRet{\fprob}$ for some $\fprob\ge\fprobMin$, then with probability at least $1-\fprob$, whenever fewer than $\capSlack$ records are friendly with at most $\athrRet$ of the blocks under $\pred_{2\thr}(\fscore)$, the filter does not abort and retains every other record.
The applications use this mainly for data in which every record is friendly with more than $\athrRet$ of the blocks, a condition that each application establishes for its own data and block size.
The slack of fewer than $\capSlack$ failing records absorbs the rare failing records of an i.i.d.\ sample (\Cref{cor:out_reg_dist_acc}) and a few arbitrary outliers.

\subsection{Relationship to FriendlyCore and Propose-Test-Release}\label{sec:ptr_comparison}

FriendlyCore, PTR and element--block filtering can all be viewed as instantiations of PFR, which differ in their filtering steps.

\paragraph{FriendlyCore.}
FriendlyCore~\cite{FriendlyCore22} compares elements with elements: its filtering step makes the retained records pairwise compatible, in that every two of them have a common friend, and its release mechanism is private on neighboring datasets whose records are pairwise compatible.
This condition is a friendliness predicate of the remaining data and the changed record.\footnote{Formally, \Cref{lem:friendly_comp} covers the zCDP filter of FriendlyCore~\cite[Algorithm~4.5]{FriendlyCore22}, which is analyzed in zero-concentrated differential privacy and which we translate to zero-out adjacency: it is jointly private and its core is pairwise compatible except with probability $\delta$~\cite[Lemma~4.6]{FriendlyCore22}, as \Cref{app:fc_gaussian} outlines. Its BasicFilter~\cite[Algorithm~4.3]{FriendlyCore22} makes the core pairwise compatible with probability one, but it is not jointly private by itself, and the analysis of FriendlyCore~\cite[Theorem~4.11]{FriendlyCore22} pays for the difference between the two cores through the group privacy of the release mechanism.}
\Cref{lem:friendly_comp} has the form of the composition lemma of FriendlyCore~\cite[Lemma~4.9]{FriendlyCore22}, which combines private decisions on the shared records with a release that is private on every neighboring pair of possible cores.
What the PFR theorem adds is the bound $\delta_3$ for each changed record: FriendlyCore instead conditions on events on which the union of the two cores is friendly~\cite[Lemma~4.6]{FriendlyCore22}, so that the release is private on every such pair.
Element--block filtering establishes only that the changed record is friendly with the remaining filtered data, which is what allows friendliness to be judged against blocks.
The price is that this guarantee requires a lower bound on the number of blocks, whereas the privacy guarantee of FriendlyCore holds for every $\nslots$.
The two instantiations place different conditions on the release mechanism, so a release mechanism designed for one need not be private under the other.
The purpose of element--block filtering is the case of larger blocks, in which a record is judged against a model supported by a block.
The number of records that then suffices for privacy is the maximum of the filtering requirement $\nPriv$, where $\Lconst=\Ldef$, and the requirement of the release mechanism, which is their sum up to a factor of two, not their product.

\paragraph{Propose-test-release.}
PTR~\cite{dwork2009differential} can be viewed as the instantiation of PFR whose filtering step retains either every record or none: it privately tests whether the input is sufficiently far from every dataset that violates a proposed bound, and otherwise rejects every record and returns the failure output $\fail$.
\Cref{def:PTR} states PTR in our notation, and \Cref{app:ptr_pfr} verifies the hypotheses of \Cref{lem:friendly_comp} for its test.
Its release mechanism needs the same kind of condition as that of element--block filtering, friendliness of the changed record with the remaining data.
Since its filtering step cannot reject only the offending records, the proposed bound is useful only if the whole input satisfies it, whereas element--block filtering needs only that few records violate it.
This comparison concerns testing a given statistic directly, since PTR can itself be applied to a robust or stabilized statistic.
For the trace and leverage scores (\Cref{thm:empirical_second_moment_PFR,def:leverage}), on an unbounded domain some value is always unfriendly with the remaining data, so every input is within one adjacency step of a violation, and PTR applied to the unfiltered statistic returns $\fail$ on every input, except with small probability (\Cref{prop:ptr_fails}).
PTR thus seeks to certify the local sensitivity of the original statistic, whereas element--block filtering exploits its typical sensitivity.
For noise addition, friendly DP follows from a bound on down sensitivity~\cite{chen2013recursive,linder2025blackbox}, the change of a statistic when records that are present are removed, and \Cref{app:down_sensitivity} explains why a test of PTR cannot simply count such removals.

%% file: sections/sec_applications.tex
\section{Applications}\label{sec:applications}

The applications in this section follow the same recipe.
Each defines a friendliness score $\fscore$ and a release mechanism $\mech$, where $\fscore(\ds,\cand)$ is large when the record $\cand$ has little influence on the target estimated from $\ds$.
The release mechanism must be friendly-DP with respect to $\pred_\thr(\fscore)$, that is, private whenever the changed record is $\thr$-friendly with the remaining data.
Once the score is shown to be concave, \Cref{thm:gen_frnd_core_priv} makes $\PFRd(\mech,\fscore,\thr)$ differentially private on every input with enough blocks, so the privacy guarantee of each application is a corollary of this theorem.
Utility, in contrast, is shown for typical data, in which every record has little influence relative to the others, so that the filter retains nearly all the records and the release is accurate.

\Cref{sec:trace} uses scalar releases as a warm-up example.
\Cref{sec:sketching} provides formal guarantees for subspace estimation, the running example of the introduction.
\Cref{sec:histogram_kl,sec:graph,sec:weighted_selection} apply the recipe to three more tasks, on user histograms, resistor networks and treatment rules.
\Cref{app:friendly-mechanisms} contains the formal details and proofs that this section only sketches, and \Cref{app:notation} gives our notation for vectors and matrices.

\subsection{Warm-Up: Scalar Releases and Trace Estimation}\label{sec:trace}

\paragraph{Setting.}
As a warm-up, we estimate the typical scale of a dataset $\ds$ of vectors in $\bbR^\dimn$, measured by its empirical trace $\smom(\ds)\coloneqq|\ds|^{-1}\sum_{j\in[|\ds|]}\lVert \ds_j\rVert^2$, the average squared norm of its records.
This scale is not known in advance, so the error of an estimate should be small relative to $\smom(\ds)$, not in absolute terms.
We therefore release the logarithm $\tracestat(\ds)\coloneqq\ln(\smom(\ds))$, whose additive error controls the multiplicative error of the trace.
For a dataset $\ds$ and a record $\cand$, denote $\psens_{\mathrm{tr}}(\ds,\cand)\coloneqq\lVert\cand\rVert^2/(|\ds|\,\smom(\ds))$.
The change $|\tracestat(\ds)-\tracestat(\ds^{i\to\nul})|$ from nulling slot $i$ of $\ds$, the per-instance sensitivity of $\tracestat$ (\Cref{def:per_instance_sens}), is at most $\psens_{\mathrm{tr}}(\ds^{i\to\nul},\ds_i)$, which depends only on the remaining data and the nulled record.
The influence of a record is thus governed by its relative scale $\lVert\ds_i\rVert^2/\smom(\ds^{i\to\nul})$, its squared norm in units of the trace of the remaining data.
A single record can be arbitrarily larger than all the others, so the global sensitivity of $\tracestat$ is unbounded.
On typical data, however, every record is on the scale of the others, and the per-instance sensitivity is $O(1/\nslots)$.
Bounded global sensitivity of $\tracestat$ can be achieved by simultaneously bounding the norm of each record from above and the trace of the remaining data from below, for example by clipping the records to a public norm and adding a public regularizer to the trace.
PFR replaces both bounds by a comparison of each record with the scale of the remaining data.

\paragraph{Score and release mechanism.}
As the release mechanism, we use the simplest one, the Gaussian mechanism $\Gauss_\sigma(\tracestat)$, which adds Gaussian noise of standard deviation $\sigma$ to the statistic.
Friendly DP requires privacy only at the pairs $\ds,\ds^{i\to\nul}$ in which $\ds_i$ is friendly with the remaining data, so for this mechanism it amounts to bounded sensitivity on these pairs.
If $\tracestat$ changes by at most a common bound on all of them, then $\sigma$ equal to $4\sqrt{\Ldef}/\eps$ times this bound makes $\Gauss_\sigma(\tracestat)$ $(\eRel,\dRel)$-friendly-DP (\Cref{fact:gauss_profile}).
Laplace noise would serve as well (\Cref{fact:laplace}), as in \Cref{thm:histogram_kl}.
We choose the score $\tracescore(\ds,\cand)$ so that $\thr$-friendliness bounds the sensitivity $\psens_{\mathrm{tr}}(\ds,\cand)$ by $1/(\thr\nslots)$, and calibrate the noise to the resulting bound.
Formally, we set
\begin{equation}\label{eq:trace_pfr}
\mech_{\mathrm{tr}}(\ds)\coloneqq\tracestat(\ds)+\Gauss(0,\sigma^2)\text{ with }\sigma\coloneqq\frac{4\sqrt{\Ldef}}{\thr\eps\nslots}
\quad\text{and}\quad
\tracescore(\ds,\cand)\coloneqq\frac{\smom(\ds)}{\lVert\cand\rVert^2}.
\end{equation}
Indeed, the score is the reciprocal of the relative scale, $1/(|\ds|\,\psens_{\mathrm{tr}}(\ds,\cand))$.
On nonempty data, the score of a zero record is $\infty$, and nulling such a record leaves the release unchanged.
On data whose records are all null or zero, such as the output of the filter when it aborts, $\mech_{\mathrm{tr}}$ releases $\ln(0)\coloneqq-\infty$.
Both neighbors have $\nslots$ slots, so a nonzero record that is $\thr$-friendly with the remaining data moves $\tracestat$ by at most $\psens_{\mathrm{tr}}(\ds^{i\to\nul},\ds_i)\le1/(\thr\nslots)$ (\Cref{lem:trace_score}), the bound to which $\sigma$ is calibrated.
A record $\cand$ is $\thr$-friendly with $\ds$ when $\lVert\cand\rVert^2\le\smom(\ds)/\thr$.
A record of a given norm is therefore acceptable when the reference data are on a comparable scale.
The score is concave, since pooling two datasets averages their traces, $\smom((\ds,\ds'))=(|\ds|\,\smom(\ds)+|\ds'|\,\smom(\ds'))/(|\ds|+|\ds'|)$ (\Cref{lem:trace_score}).
Instantiating \Cref{thm:gen_frnd_core_priv} with this score and release mechanism yields the next corollary.

\def\EmpiricalSecondMomentPFRTheorem{
    Given $\nslots,\bsz\in\bbN$; $\eps\in(0,1]$; $\delta\in(0,1)$; and a friendliness threshold $\thr>0$, if $\nslots\ge\cBlk\bsz\Lconst^2/\eps^{3/2}$, then $\PFRd(\mech_{\mathrm{tr}},\tracescore,\thr)$ is $(\eps,\delta)$-DP on datasets of $\nslots$ records, where $\Lconst\coloneqq\Ldef$, and the release mechanism $\mech_{\mathrm{tr}}$ and the score $\tracescore$ are given by \eqref{eq:trace_pfr}.
}
\begin{corollary}[Private trace estimation]\label[corollary]{thm:empirical_second_moment_PFR}
    \EmpiricalSecondMomentPFRTheorem
\end{corollary}

Nothing in this argument is specific to the trace.
For any real-valued statistic $\stat$ and any function $\psens$ with $|\stat(\ds)-\stat(\ds^{i\to\nul})|\le\psens(\ds^{i\to\nul},\ds_i)$ on every dataset $\ds$ and slot $i$, if the score $1/(|\ds|\,\psens(\ds,\cand))$ is concave, then the same noise gives the same guarantee (\Cref{thm:PFR_DP}).

\paragraph{Utility.}
If every squared norm is at most $1/(\cThr\thr)$ times the trace of the input, for $\thr\le1/\cThr$, the blocks have size of order $\ln(\nslots/\fprob)/\thr$, and $\nslots=\wtOmega(1/(\thr\eps^{3/2}))$, then with probability at least $1-\fprob$ the filter retains every record, and the exponential $e^Y$ of the output $Y$ is within a factor $e^{\pm\xi}$ of the trace of the input, where $\xi$ is of order $\sqrt{\ln(1/\delta)\ln(1/\fprob)}/(\thr\eps\nslots)$ (\Cref{cor:trace_utility}).
If the records are independent draws whose squared norms are at most $1/(\cThr\thr)$ times their unknown mean, the same holds with high probability (\Cref{cor:trace_population}).

\subsection{Gaussian Sketches and Subspace Estimation}\label{sec:sketching}

\paragraph{Setting.}
We now return to the running example of the introduction: estimating the principal subspace of a Gaussian distribution, the span of the top $\rank$ eigenvectors of its covariance $\Sigma$, from samples whose norms have no public bound.
An estimate is a projector $\widetilde P$, and its error is its distance $\lVert\widetilde P-\projk(\Sigma)\rVert$ in operator norm from the projector $\projk(\Sigma)$ onto the principal subspace.
The standard non-private estimate is the projector onto the top $\rank$ eigenvectors of the empirical second moment of the samples.
The sample size required for a given error grows with $\rank$, with the total variance in the remaining directions relative to $\lambda_\rank(\Sigma)$, and with the ratio $\lambda_{\rank+1}(\Sigma)/\lambda_\rank(\Sigma)$.

For a public ridge $\ridge\ge0$, denote the regularized second moment of a dataset $\ds$ by $\regmom_\ridge(\ds)\coloneqq\ds^T\ds/|\ds|+\ridge I$, with null records represented by zero vectors.
The leverage of a vector $\cand$ with respect to $\ds$ is $\lev_\ridge(\ds,\cand)\coloneqq\cand^T\regmom_\ridge(\ds)^{-1}\cand$ (\Cref{def:leverage}).
Nulling slot $i$ of $\ds$ removes the summand $\ds_i\ds_i^T/\nslots$ from $\regmom_\ridge(\ds)$, so $\regmom_\ridge(\ds^{i\to\nul})\preceq\regmom_\ridge(\ds)\preceq\bigl(1+\lev_\ridge(\ds^{i\to\nul},\ds_i)/\nslots\bigr)\regmom_\ridge(\ds^{i\to\nul})$.
The relative change of the second moment in every direction is thus governed by the leverage of the nulled record with respect to the remaining data, a squared Mahalanobis norm that is small along the directions in which these data vary and large across them.
A single record that sticks out across these directions has a large leverage and can change the empirical subspace drastically, regardless of the sample size.
On Gaussian data, however, a typical record is spread like the others, and its leverage does not grow with the sample size.
Its expected leverage with respect to the population is the \emph{effective dimension} $\effdim(\Sigma)\coloneqq\Tr(\Sigma(\Sigma+\ridge I)^{-1})$, roughly the number of directions whose variance exceeds the ridge, which a positive ridge can make much smaller than $\dimn$.
The analysis of the Gaussian mixing mechanism~\cite[Lemma~1]{LevSSLSW25} controls the leverage through public bounds on the norms of the rows and on the smallest eigenvalue of the data, whereas PFR tests the leverage of each record against the reference blocks and needs no such bounds.

\paragraph{Score and release mechanism.}
We release $\nsk\in\bbN$ independent \emph{Gaussian sketches}, (regularized) random Gaussian combinations of the records of $\ds$ whose randomness itself hides each record, and take the reciprocal leverage as the score:
\begin{equation}\label{eq:sketch_pfr}
\mech_{\ridge,\nsk}(\ds)\coloneqq(Z_1,\ldots,Z_\nsk)\text{ for independent }Z_j\sim\Gauss(0,\regmom_\ridge(\ds))
\quad\text{and}\quad
\fscore_\ridge(\ds,\cand)\coloneqq\frac{1}{\lev_\ridge(\ds,\cand)}.
\end{equation}
As with the samples, the estimate is the projector onto the top $\rank$ eigenvectors of the empirical second moment of the sketches.
For the sketches, friendly DP amounts to bounded leverage of the nulled record on the friendly pairs, since this leverage bounds the R\'enyi divergence between the two sketch distributions (\Cref{prop:sketch_RDP}).
If this leverage is at most $\eps\nslots/(2(\Kconst))$ on all these pairs, where $\Lconst\coloneqq\Ldef$, then $\mech_{\ridge,\nsk}$ is $(\eRel,\dRel)$-friendly-DP.
Friendliness under this score bounds the leverage: a record that is $\thr$-friendly with the remaining data has leverage at most $1/\thr$, which meets this condition once $\nslots\ge\sketchPriv$, the second condition of the next corollary.
The tightened test $\fscore_\ridge(\ds,\cand)\ge2\thr$ asks that $\cand$ lie in an ellipsoid shaped like the reference data, $\{y:y^T\regmom_\ridge(\ds)^{-1}y\le1/(2\thr)\}$, which is long along the directions in which these data vary and short across them (\Cref{fig:friendliness}b).
The ridge $\ridge$ keeps this ellipsoid from collapsing when a small block spans only a few directions, since without it the leverage is infinite outside the span of the reference data.
The score is concave, since $A\mapsto1/(\cand^TA^{-1}\cand)$ is a concave function of a positive definite matrix $A$, and pooling two datasets averages their regularized second moments, $\regmom_\ridge((\ds,\ds'))=(|\ds|\,\regmom_\ridge(\ds)+|\ds'|\,\regmom_\ridge(\ds'))/(|\ds|+|\ds'|)$ (\Cref{lem:lev_concave}).

\def\FilteredSketchTheorem{
    Given $\nslots,\bsz,\nsk\in\bbN$; $\eps\in(0,1]$; $\delta\in(0,1)$; a ridge $\ridge\ge0$ and a friendliness threshold $\thr>0$, if $\nslots\ge\max\left\{\cBlk\bsz\Lconst^2/\eps^{3/2}, \sketchPriv\right\}$, then $\PFRd(\mech_{\ridge,\nsk},\fscore_\ridge,\thr)$ is $(\eps,\delta)$-DP on datasets of $\nslots$ records, where $\Lconst\coloneqq\Ldef$, and the release mechanism $\mech_{\ridge,\nsk}$ and the score $\fscore_\ridge$ are given by \eqref{eq:sketch_pfr}.
}
\begin{corollary}[Private Gaussian sketching]\label[corollary]{thm:filt_sketch_DP}
    \FilteredSketchTheorem
\end{corollary}

The first condition pays for the filter and the second for the sketches: the more sketches are released, the more records are needed to hide each record in all of them.
Privacy holds for arbitrary unbounded vectors, and the dimension $\dimn$ does not appear in its conditions.

\phantomsection\paragraph{Utility.}\label{sec:subspace_utility}
For a rank $\rank<\dimn$, the \emph{subspace mechanism} runs $\PFRd(\mech_{\ridge,\nsk},\fscore_\ridge,\thr)$ and post-processes the sketches into the projector $\widetilde P$ onto the top $\rank$ eigenvectors of their empirical second moment $\nsk^{-1}\sum_{j\in[\nsk]}Z_jZ_j^T$.
On Gaussian data, once the blocks are somewhat larger than $\effdim(\Sigma)$, the leverage of every record against every block is at most of the order of $\effdim(\Sigma)$ plus a logarithmic term, which pays for controlling all the records at once.
With the block size and $1/\thr$ of this order and enough blocks, the filter therefore retains every record with high probability (\Cref{lem:population_retention,cor:pfr_retention}).
The sketches then approximate $\Sigma+\ridge I$, so $\widetilde P$ approximates the target as long as the top $\rank$ eigenvalues of $\Sigma$ are separated from the others.

To quantify this, we return to the geometric spectrum of the introduction, a simplified spectrum in the harder regime, where the eigenvalues on the two sides of the cutoff differ by a constant factor but their gap shrinks with $\dimn$.
For $\lceil\log_2(\dimn)\rceil<\rank\le\dimn/2$, it has $\rank-\lceil\log_2(\dimn)\rceil$ unit eigenvalues, followed by eigenvalues that halve at every step, $\lambda_{\rank-\lceil\log_2(\dimn)\rceil+j}(\Sigma)=2^{-j}$ for $j\ge1$, with arbitrary unknown eigenvectors.
Its $\rank^{th}$ eigenvalue $\lambda_\rank(\Sigma)=2^{-\lceil\log_2(\dimn)\rceil}$, between $1/(2\dimn)$ and $1/\dimn$, is twice the next one.
The remaining $\dimn-\rank$ eigenvalues sum to less than $\lambda_\rank(\Sigma)$.

\begin{theorem}[Subspace estimation on the geometric spectrum, informal]\label{thm:subspace_informal}
    Given $\dimn,\rank,\nslots\in\bbN$ with $\dimn\ge8$ and $\lceil\log_2(\dimn)\rceil<\rank\le\dimn/2$; $\acc\in(0,1/4]$; $\fprob\in(0,1/4)$; $\eps\in(0,1]$; and $\delta\in(0,\cFprobQ\fprob]$, if $\nslots=\wtOmega(\rank/\eps^{3/2}+\rank/\acc^2+\rank^{3/2}/(\acc\eps))$, then for some choice of its parameters, the subspace mechanism is $(\eps,\delta)$-DP on every input.
    With these parameters, if the records are independent draws from $\Gauss(0,\Sigma)$ for a covariance $\Sigma$ with the geometric spectrum, then with probability at least $1-\fprob$ it retains every record and returns $\widetilde P$ with $\lVert\widetilde P-\projk(\Sigma)\rVert\le\acc$, where $\projk(\Sigma)$ is the projector onto the top $\rank$ eigenvectors of $\Sigma$.
\end{theorem}

The term $\rank/\acc^2$ is the sample size that suffices for this spectrum without privacy, independently of $\dimn$, as the proof of \Cref{thm:subspace_running} shows.
The term $\rank/\eps^{3/2}$ pays for the filter and the term $\rank^{3/2}/(\acc\eps)$ for the sketches (\Cref{cor:subspace_private}).
The formal utility guarantee, \Cref{thm:subspace_running}, sizes the blocks, the leverage bound $1/\thr$ and the number of sketches by the rank rather than by the dimension, with a ridge $\ridge=\lambda_\rank(\Sigma)/\dimn$ far below the cutoff.
Its proof measures all errors relative to $|\Sigma-(3/4)\lambda_\rank(\Sigma)I|$, which weighs each direction by its distance from the cutoff (\Cref{lem:gap_weighted_projector}), so that the many tail directions together count only as a constant.
The dimension thus enters only through the ridge and the condition $\rank>\lceil\log_2(\dimn)\rceil$.
This restriction comes from the geometric spectrum, which we used for ease of presentation.
The guarantee extends, without this restriction, to every covariance whose $\rank^{th}$ eigenvalue is at least twice the next and whose effective dimension at the ridge $\lambda_\rank(\Sigma)/\dimn$ is of order $\rank$, such as the large-gap spectrum of the introduction (\Cref{cor:subspace_general_spectra}).
This ridge requires $\lambda_\rank(\Sigma)$ to be public, whereas on the geometric spectrum $\dimn$ determines it.

To our knowledge, no previously published bound for this spectrum and the projector loss is independent of $\dimn$ for a given rank, as that of \Cref{thm:subspace_informal} is, though an adaptation of the stable covariance estimator of \citet{BrownHS23} may recover similar guarantees.\footnote{We note that our dependence on $\eps$ in the filtering term $\rank/\eps^{3/2}$ is slightly worse than that of \citet{BrownHS23}, which scales linearly with $1/\eps$.}

\subsection{KL Divergence between User Histograms and a Uniform Reference}\label{sec:histogram_kl}

\paragraph{Setting.}
Consider the task of measuring, in KL divergence, how far the pooled observations of a population of users are from a reference distribution over a finite set of $\ncat$ public categories, when each user contributes $\nobs$ observations.
The privacy guarantee is at the level of users: a record is the whole histogram of a user, so zero-out adjacency protects all their observations.
For simplicity, we take the uniform distribution $\unif$ on the categories as the reference, so that the divergence measures how unevenly the users together cover the categories.
The target is the divergence $\klstat(\ds)\coloneqq\KL(\unif\Vert\sfreq(\ds))$ between the uniform reference and the smoothed pooled frequencies $\sfreq_j(\ds)\coloneqq(\pcount+\avgH_j(\ds))/(\nobs+\ncat\pcount)$, where $\avgH_j(\ds)$ is the average count of category $j$ over the records of $\ds$ and $\pcount>0$ is a public pseudocount.
The pseudocount, as if every user had contributed $\pcount$ additional observations to every category, keeps the divergence finite when a category is absent.\footnote{The pseudocount should not be much larger than $\nobs/\ncat$. Every smoothed frequency is at least $\pcount/(\nobs+\ncat\pcount)$, so the target is at most $\ln(1+\nobs/(\ncat\pcount))$ on every dataset, and the additive error guarantee below is informative only for errors smaller than this bound.}
Null records, which the filter produces, count as zero histograms, and the denominators stay fixed, so on filtered data the frequencies $\sfreq(\ds)$ may sum to less than one (\Cref{app:kl_setting}).
When the filter retains every user, $\klstat$ is the divergence above.

In the worst case, a user places all $\nobs$ observations on a category that no other user covers.
Nulling this user lowers the average count of this category from $\nobs/\nslots$ to zero, and changes the divergence by $\ncat^{-1}\ln(1+\nobs/(\nslots\pcount))$, whatever the other users contribute to the other categories (\Cref{claim:kl_global_sensitivity}).
When $\nslots\pcount\ge\nobs$, this change is of order $\nobs/(\ncat\nslots\pcount)$, which is large for a small pseudocount.
Laplace noise calibrated to the global sensitivity therefore has scale of order $\nobs/(\ncat\nslots\pcount\eps)$, so error $\acc$ requires $\nslots$ of order $\nobs/(\ncat\pcount\acc\eps)$, up to a logarithmic factor.
In general, nulling slot $i$ of $\ds$ changes $\klstat$ by at most $(\nslots\cdot\ncat)^{-1}\sum_{j\in[\ncat]}\hist_j/(\pcount+\avgH_j(\ds^{i\to\nul}))$, where $\hist\coloneqq\ds_i$ is the histogram of the user (\Cref{lem:kl_score}).
The influence of a user is thus governed by their counts, each measured in units of the regularized average count of its category among the remaining users.
The scales of the categories are unknown and may differ widely, but on typical data every count of a user is on the scale of the average count of its category, and the per-instance sensitivity is $O(1/\nslots)$.
PFR can exploit this without knowing these scales, by testing the counts of each user against the average counts of the other users.

\paragraph{Score and release mechanism.}
We release $\klstat$ with additive Laplace noise.
As for the trace, friendly DP amounts to bounded sensitivity on the friendly pairs: if $\klstat$ changes by at most a common bound on all of them, then Laplace noise of scale $2/\eps$ times this bound makes the release $(\eRel,0)$-friendly-DP (\Cref{fact:laplace}).
Gaussian noise would serve as well (\Cref{fact:gauss_profile}), as in \Cref{thm:empirical_second_moment_PFR}.
To make friendliness bound this sensitivity, we score a histogram $\hist$ against a dataset $\ds$ category by category, and add noise of the resulting scale:
\begin{equation}\label{eq:kl_pfr}
\mech_{\mathrm{KL}}(\ds)\coloneqq\klstat(\ds)+\Lap\Bigl(\frac{2}{\thr\nslots\eps}\Bigr)
\quad\text{and}\quad
\fKL(\ds,\hist)\coloneqq\Bigl(\frac{1}{\ncat}\sum_{j\in[\ncat]}\frac{\hist_j}{\pcount+\avgH_j(\ds)}\Bigr)^{-1}.
\end{equation}
The sensitivity bound above is thus $1/(\nslots\,\fKL(\ds^{i\to\nul},\ds_i))$, so a user who is $\thr$-friendly with the remaining data moves $\klstat$ by at most $1/(\thr\nslots)$, the bound to which the noise of $\mech_{\mathrm{KL}}$ is calibrated.
A user is $\thr$-friendly with $\ds$ when their relative counts $\hist_j/(\pcount+\avgH_j(\ds))$ average at most $1/\thr$.
Since the test averages over all the counts of a user, a user who concentrates on one category is not rejected for that alone.
Concavity follows since $z\mapsto(\sum_j\hist_j/z_j)^{-1}$ is a concave function of a positive vector $z$, and the average counts of a pooled dataset are $\avgH_j((\ds,\ds'))=(|\ds|\,\avgH_j(\ds)+|\ds'|\,\avgH_j(\ds'))/(|\ds|+|\ds'|)$ (\Cref{lem:kl_score}).

\begin{corollary}[Private KL divergence of user histograms]\label[corollary]{thm:histogram_kl}
    Given $\nslots,\bsz\in\bbN$; $\eps\in(0,1]$; $\delta\in(0,1)$; a public pseudocount $\pcount>0$ and a friendliness threshold $\thr>0$, if $\nslots\ge\cBlk\bsz\Lconst^2/\eps^{3/2}$, then $\PFRd(\mech_{\mathrm{KL}},\fKL,\thr)$ is $(\eps,\delta)$-DP on datasets of $\nslots$ records, where $\Lconst\coloneqq\Ldef$, and the release mechanism $\mech_{\mathrm{KL}}$ and the score $\fKL$ are given by \eqref{eq:kl_pfr}.
\end{corollary}

\paragraph{Utility.}
Utility requires only that no user be extreme relative to the others.
If each count of each user is at most $1/(\cThr\thr)$ times the regularized average count of its category in the input, for $\thr\le1/\cThr$, then with high probability the filter retains all the users, and $\nslots=\wtO(\ln(\ncat)/(\thr\eps^{3/2})+1/(\thr\acc\eps))$ users suffice for error $\acc$ with respect to $\klstat(\ds)$ (\Cref{thm:histogram_kl_utility}).
This is a deterministic condition on the input, so the users may follow different distributions, or none, and the algorithm uses no knowledge of the frequencies.
The factor $\ln(\ncat)$ enters through the block size, which the proof chooses so that every block preserves the regularized average count of every category up to a factor of two.
The condition is expected in the natural case where all the users draw their observations from the same distribution over the categories.
Each count of a user is then a binomial variable whose mean is the expected average count of its category, so it rarely exceeds twice this mean by more than a logarithmic term, which the pseudocount absorbs.
With enough users, the condition holds with high probability once $1/\thr$ is of order $1+\ln(\nslots\ncat/\fprob)/\pcount$ (\Cref{claim:kl_multinomial}).
Compared with direct noise, PFR thus replaces the factor $\nobs/(\ncat\pcount)$ by $1/\thr$, through a single concave score that adapts to the unknown scales of the categories.

\subsection{All-Pairs Effective Resistance}\label{sec:graph}

\paragraph{Setting.}
Consider the task of estimating, for every pair of vertices of a resistor network, how well the two are connected through all the paths between them.
The network is a weighted graph $G$ in which an edge of weight $w$ is an electrical conductor of conductance $w$.
The effective resistance between two vertices quantifies this connectivity: it is the voltage that sends a unit current between them.
In matrix form, the resistance between two vertices $u$ and $u'$ in the same component of $G$ is $\reff_G(u,u')\coloneqq(\chi_u-\chi_{u'})^T\lapl_G^\dagger(\chi_u-\chi_{u'})$, where $\lapl_G$ is the Laplacian of $G$ (its weighted degree matrix minus its weighted adjacency matrix) and $\chi_u$ is the standard basis vector of $u$.
The $v$ vertices are public and the records are the edges, so zero-out adjacency protects an entire edge, including its weight.
The weights have no public bound, so the resistances have no known scale, and as for the trace, the error target is relative.

The influence of an edge $e=\{u,u'\}$ of weight $w_e$ on the resistances of $G$ is measured by its \emph{leverage} $\ell_e\coloneqq w_e\reff_G(u,u')$, the share of the connectivity between its endpoints that the edge itself supplies.
The Sherman--Morrison formula makes this precise: removing an edge of leverage $\ell_e<1$ multiplies the resistance between its endpoints by $1/(1-\ell_e)$, and no resistance by more.
A bridge, whose endpoints are joined by no other path, has the maximal leverage one.
Removing it disconnects the graph, so a single edge can change some resistances arbitrarily.
In a well-connected graph, by contrast, many other paths join the endpoints of every edge, so every edge has small leverage and removing any single edge changes the resistances only by a small factor.
Two published guarantees rely on public bounds: in \citet{Graph24}, neighboring graphs differ in the weight of one edge by at most one, and \citet{Graph26} assume unweighted graphs whose maximum degree is at most a public bound.
PFR instead tests the leverage of each edge with respect to the other edges, and needs neither bound.

\paragraph{Score and release mechanism.}
The release mechanism is the graph analogue of the Gaussian sketches.
We record an edge $e=\{u,u'\}$ as the vector $\sqrt{w_e}(\chi_u-\chi_{u'})\in\bbR^v$, so that $\lapl_\ds$ is the sum of the outer products of the records.
We draw $\nsk\in\bbN$ independent potential vectors on the vertices and, as in \eqref{eq:sketch_pfr}, score a record by its reciprocal leverage, with zero ridge:
\begin{equation}\label{eq:graph_pfr}
\mech_{\mathrm{pot}}(\ds)\coloneqq(Y_1,\ldots,Y_\nsk)\text{ for independent }Y_j\sim\Gauss(0,\lapl_\ds^\dagger)
\quad\text{and}\quad
\fscore_0(\ds,\cand)=\frac{1}{\lev_0(\ds,\cand)}.
\end{equation}
On these records, the leverage is $\lev_0(\ds,\cand)=|\ds|\,\cand^T\lapl_\ds^\dagger\cand$ for $\cand$ in the range of $\lapl_\ds$, and $\infty$ otherwise (\Cref{app:graph_score}).
Since $\bbE[(Y_j(u)-Y_j(u'))^2]=\reff_\ds(u,u')$ for vertices $u$ and $u'$ connected in $\ds$, averaging these squared differences over the draws estimates all the resistances at once.
The covariance of the potentials is $\lapl_\ds^\dagger=\regmom_0(\ds)^\dagger/\nslots$, the scaled pseudoinverse of the covariance of a zero-ridge sketch of the edges.
Inverting both covariances only swaps the two directions of the R\'enyi divergence (\Cref{lem:inverse_gaussian}), so friendly DP again amounts to bounded leverage on the friendly pairs.
As for the sketches, $\mech_{\mathrm{pot}}$ is $(\eRel,\dRel)$-friendly-DP if the leverage $\lev_0(\ds^{i\to\nul},\ds_i)$ of the nulled edge is at most $\eps\nslots/(2(\Kconst))$ on all these pairs.
With this score, friendliness bounds the leverage of the nulled edge: an edge that is $\thr$-friendly with the remaining edges has $\lev_0(\ds^{i\to\nul},\ds_i)\le1/\thr$, which meets this condition for the threshold $\thr\coloneqq2(\Kconst)/(\nslots\eps)$ of the next corollary, the smallest that the condition allows.
An edge $e=\{u,u'\}$ is $\thr$-friendly with a dataset $\ds$ of edges when $u$ and $u'$ are connected in $\ds$ and $w_e\reff_\ds(u,u')\le1/(\thr|\ds|)$, that is, when the edges of $\ds$ join its endpoints well relative to its weight.
The score is concave by \Cref{lem:lev_concave}.

\begin{corollary}[Private all-pairs effective resistance]\label[corollary]{thm:graph_resistance}
    Given $\nslots,\bsz,\nsk\in\bbN$; $\eps\in(0,1]$; and $\delta\in(0,1)$, if $\nslots\ge\cBlk\bsz\Lconst^2/\eps^{3/2}$, then $\PFRd(\mech_{\mathrm{pot}},\fscore_0,\thr)$ is $(\eps,\delta)$-DP on all weighted graphs given as datasets of $\nslots$ edge records, connected or not, where $\Lconst\coloneqq\Ldef$, the release mechanism $\mech_{\mathrm{pot}}$ and the score $\fscore_0$ are given by \eqref{eq:graph_pfr}, and $\thr\coloneqq2(\Kconst)/(\nslots\eps)$.
\end{corollary}

\paragraph{Utility.}
Utility holds on well-connected graphs, in which no single edge supplies much of the connectivity.
A public target $\acc$ for the relative error and a failure probability $\fprob$ fix the number $\nsk$ of potential vectors, of order $\ln(v/\fprob)/\acc^2$, and through it the threshold $\thr$.
If the input is connected, its maximum leverage $\lmax$ is small enough for random blocks of edges to preserve its connectivity (\Cref{lem:graph_blocks}), and $\acc$ is at least of order $\lmax/\eps$, up to logarithmic factors, then with high probability the filter retains all the edges and every resistance is estimated to relative error $\acc$ (\Cref{thm:graph_resistance_utility}).
The mechanism never needs to know $\lmax$.
Being well connected in this sense does not require a large spectral gap: two cliques joined by a perfect matching have spectral gap only $2$ but maximum leverage of order $1/v$.

\subsection{Treatment-Rule Selection}\label{sec:weighted_selection}

\paragraph{Setting.}
Consider a treatment, such as a job-training program, that has so far been assigned to participants at random, with probabilities that may depend on their characteristics.
The record of each participant is $\cand=(\cov,\trt,\outc)$, with covariates $\cov$, such as education and prior earnings, an assignment (or eligibility) $\trt\in\{0,1\}$ to the treatment, and an outcome $\outc\ge0$, such as later earnings.
The assignment probabilities $\prop(\trt\mid\cov)$ are known, as in a randomized trial.
We wish to choose, from a public library $\hypclass$ of $\nhyp$ \emph{treatment rules} $h$, which map covariates to $\{0,1\}$, the rule under which future participants would have the largest average outcome.
The outcome of a past participant shows what a rule $h$ would achieve for them only if $h(\cov)=\trt$, which happens with probability $\prop(h(\cov)\mid\cov)$.
We weight each record by $\weight(\cand)\coloneqq\outc/\prop(\trt\mid\cov)$ to compensate for this probability.
Empirical welfare maximization~\cite{kitagawa2018welfare} selects the rule $h$ with the largest \emph{empirical welfare} $|\ds|^{-1}\sum_{i\in[|\ds|]}\weight(\ds_i)\ind\{h(\cov_i)=\trt_i\}$ on a dataset $\ds$ of past records $\ds_i=(\cov_i,\trt_i,\outc_i)$.
On independent records, the empirical welfare of $h$ is an unbiased estimate of the average outcome when the treatment is assigned by $h$.
We denote the total weight of $\ds$ by $\totweight(\ds)\coloneqq\sum_{i\in[|\ds|]}\weight(\ds_i)$ and the weighted risk of a rule $h$ by $\wrisk_h(\ds)\coloneqq\totweight(\ds)^{-1}\sum_{i\in[|\ds|]}\weight(\ds_i)\ind\{h(\cov_i)\ne\trt_i\}$, with $\wrisk_h(\ds)\coloneqq0$ if $\totweight(\ds)=0$, and with null records contributing zero.
The empirical welfare of $h$ on $\ds$ equals $(\totweight(\ds)/|\ds|)(1-\wrisk_h(\ds))$, so minimizing the weighted risk maximizes the welfare, which is the reduction of outcome-weighted learning to weighted classification~\cite{zhao2012outcome}.
Unlike the empirical welfare, the weighted risk lies in $[0,1]$ and is invariant to rescaling the outcomes, whose scale is not known in advance.

In the worst case, a rule $h$ agrees with the past assignment of every participant but one, whose outcome is huge.
Nulling the record $\ds_i$ of this participant lowers $\wrisk_h$ from $\weight(\ds_i)/\totweight(\ds)$ to zero, a change that approaches one, the whole range of the weighted risk, as their outcome grows.
In general, nulling slot $i$ of $\ds$ changes every weighted risk by at most $\weight(\ds_i)/\totweight(\ds^{i\to\nul})$, the weight of the record relative to the remaining data (\Cref{lem:weighted_risk_sensitivity}).
On typical data, every weight is on the scale of the average weight, and this per-instance sensitivity is $O(1/\nslots)$.
Private weighted risk minimization over a continuous class of predictors~\cite{giddens2023weighted} relies on a public bound on the weights and on clipped outcomes.
Clipping at a conservative bound makes the error depend on that bound rather than on the outcomes of typical participants.
PFR instead compares the weight of each record with the average weight of the other records, and needs no public bound.

\paragraph{Score and release mechanism.}
We select the rule with the exponential mechanism~\cite{mcsherry2007mechanism}, which samples each rule with a probability that decreases exponentially with its weighted risk.
Friendly DP again reduces to bounded sensitivity on the friendly pairs, now of every weighted risk.
If all of them change by at most some $B>0$ on these pairs, then selecting each rule $h$ with probability proportional to $\exp(-\eps\wrisk_h(\ds)/(4B))$ makes the selection $(\eRel,0)$-friendly-DP.
Instead of a public bound on the weights, we let friendliness bound the relative weight of the nulled record, and select with the resulting sensitivity:
\begin{equation}\label{eq:selection_pfr}
\Pr[\mech_{\mathrm{sel}}(\ds)=h]\propto\exp\Bigl(-\frac{\eps\thr\nslots\wrisk_h(\ds)}{4}\Bigr)\text{ for }h\in\hypclass
\quad\text{and}\quad
\sensW(\ds,\cand)\coloneqq\frac{\totweight(\ds)}{|\ds|\,\weight(\cand)}.
\end{equation}
If $\ds_i$ is $\thr$-friendly with the remaining data, then $\weight(\ds_i)\le\totweight(\ds^{i\to\nul})/(\thr\nslots)$, so $B=1/(\thr\nslots)$ bounds the change of every weighted risk, and $\mech_{\mathrm{sel}}$ is the selection above with this $B$.
A record $\cand$ is $\thr$-friendly with $\ds$ when its weight is at most $1/\thr$ times the average weight of $\ds$.
A non-null record of zero weight has score $\infty$, and nulling it leaves every weighted risk unchanged.
Total weights add up under pooling, $\totweight((\ds,\ds'))=\totweight(\ds)+\totweight(\ds')$, so the score is concave, with equality.

\begin{corollary}[Private treatment-rule selection]\label[corollary]{thm:weighted_selection}
    Given $\nslots,\bsz\in\bbN$; $\eps\in(0,1]$; $\delta\in(0,1)$; a public friendliness threshold $\thr>0$, and a public library $\hypclass$ of treatment rules, if $\nslots\ge\cBlk\bsz\Lconst^2/\eps^{3/2}$, then, for arbitrary nonnegative and unbounded weights, $\PFRd(\mech_{\mathrm{sel}},\sensW,\thr)$ is $(\eps,\delta)$-DP on datasets of $\nslots$ records, where $\Lconst\coloneqq\Ldef$, and the release mechanism $\mech_{\mathrm{sel}}$ and the score $\sensW$ are given by \eqref{eq:selection_pfr}.
\end{corollary}

\paragraph{Utility.}
If the records are independent draws whose weights are at most $1/(\cThr\thr)$ times their unknown mean $\mu$, for $\thr\le1/\cThr$, then with high probability the filter removes nothing, and $\wtO(1/(\thr\eps^{3/2})+\ln(\nhyp/\fprob)/(\thr\acc^2)+\ln(\nhyp/\fprob)/(\thr\acc\eps))$ records suffice for excess population risk $\acc$, a welfare regret of $\acc\mu$ (\Cref{cor:weighted_population}).
With a trustworthy public bound, clipping the weights and applying the exponential mechanism is simpler, whereas PFR needs no absolute bound and keeps the participants with large outcomes, as long as they are not extreme relative to the others.

%% file: sections/sec_discussion.tex
\section{Discussion}\label{sec:discussion}

This work introduces propose-filter-release, a general technique that takes a release mechanism that is private only on neighboring datasets satisfying a condition between their differing record and the remaining data, and turns it into a mechanism that is differentially private in the standard sense, using a jointly private filtering step that establishes the required condition.
The filtering step that this work develops, element--block filtering, nulls the records that fail tests against reference blocks of the data.
PFR with this filtering step is instantiated by a concave friendliness score and a friendly-DP release mechanism; in \Cref{sec:applications} we present a range of applications that vary in their data and in their release mechanisms.
In our construction, the number of blocks that suffices for privacy depends only on the privacy parameters, so the filtering overhead is linear in the block size.
The blocks are used only to test individual records, and need not depend on the target accuracy.

\subsection{Limitations and Open Directions}\label{sec:limitations}

\paragraph{Limitations.}
The filter aborts once order $\ln(1/\delta)/\eps$ records fail.
This number does not grow with $\nslots$ for fixed privacy parameters, so the filter does not tolerate a constant \emph{fraction} of contamination, the regime studied in robust statistics.
The utility results concern inputs on which the filter removes nothing, where it serves as a safeguard for privacy.
Accuracy on the filtered data carries over to the original input, or to the population, only when the filter retains the records of the input.
Retention depends on the data distribution and on the application, since it requires representative reference blocks.
The thresholds, the regularization, the block size and any spectral information used to set them are public parameters, which must be set independently of the data or privatized separately.
The constants in \Cref{thm:gen_frnd_core_priv} and in the utility results were not optimized: at $\eps=1$ and $\delta=10^{-6}$ the theorem asks for about $1.1\cdot10^{5}$ reference blocks, so the guarantees describe the order $\ln^2(1/\delta)/\eps^{3/2}$ of the block count rather than practical parameter settings.

\paragraph{Open directions.}
With its cap of order $\ln(1/\delta)/\eps$ rejections, the filter requires order $\ln^2(1/\delta)/\eps^{3/2}$ blocks.
It is possible that a cap that does not grow with $1/\eps$ lowers this requirement to order $\ln^2(1/\delta)/\eps$.
Another open question is whether testing records against blocks can be private without a lower bound on the number of blocks, as the zCDP filter of FriendlyCore is private for every number of records.
Other directions include tolerating more outliers with useful accuracy after their removal, selecting the most frequent categories from histograms of heterogeneous users, and further release mechanisms that admit useful concave scores.
Another direction is selecting the public parameters privately, for which natural starting points are a binary search, as in \citet{tsfadia2024subspace}, and, when a parameter depends on the scale of the data, an instance of PFR that estimates this scale (\Cref{thm:empirical_second_moment_PFR}).

\subsection{Related Work}\label{sec:discussion_related}
FriendlyCore~\cite{FriendlyCore22} and the independently developed framework of \citet{AL22} turn pairwise agreement among candidate estimates into private aggregation; \citet{dagan2024anisotropic} use the former, for example, for private mean estimation that exploits anisotropy, and \citet{arbas2023mixtures} extend the latter to Gaussian mixtures.
These frameworks accommodate structured candidates, such as covariance matrices or the subspace estimates of block--block FriendlyCore~\cite{tsfadia2024subspace}, but friendship remains a relation between two candidates, so structure that is visible only in a collection of records must first be summarized by a model estimated on each block.
Our larger blocks instead let a record be judged against a model supported by several records.
Smooth sensitivity~\cite{nissim2007smooth} calibrates the noise to a smooth upper bound on the local sensitivity, which keeps the mechanism private on every input but requires bounding the local sensitivity at the datasets near the input.
Generalized PTR~\cite{redberg2023gptr} privately tests a data-dependent bound on the privacy loss of a mechanism, rather than on the sensitivity of a statistic, which extends PTR beyond noise addition.
Like PTR~\cite{dwork2009differential}, both exploit the instance-dependent stability of the input as it is, and decline to answer or add large noise rather than remove individual records.
Other frameworks establish stability for specific estimators~\cite{KMV22,zampetakis2025truncation,duchi2023adaptive,BrownHHKLIPS24}.
Closest in structure to our use of a filter is the restricted sensitivity of \citet{blocki2013restricted}, who compose a query with a deterministic projection onto a family of datasets on which it is less sensitive.
Our filter plays the role of such a projection, onto the datasets in which the record of a given slot is friendly with the remaining records, but it is randomized, its stability is joint differential privacy, and it reaches this family for each slot separately, with high probability.
Friendly DP is pDP~\cite{wang2019perinstance} at the pairs that a predicate admits (\Cref{def:friendly_dp}), and down sensitivity~\cite{chen2013recursive,linder2025blackbox} measures the change of a function when records that are present are removed.
What we add to both is a filtering step that enforces the predicate before the release mechanism runs.
Finally, Gaussian sketching is private for rows of bounded norm when the smallest singular value of the data is large enough~\cite{blocki2012jl,sheffet2017ols,LevSSLSW25}, whereas \Cref{thm:filt_sketch_DP} gives privacy for arbitrary unbounded inputs, through the leverage test of the filter.

\paragraph{General frameworks for black-box statistics.}
Recent general methods estimate a real-valued statistic from black-box access, through interfaces that differ from that of PFR.
\citet{steinke2026blackbox} evaluate the statistic only on subsets of the data, and trade statistical efficiency, the number of samples, against oracle efficiency, the number of evaluations.
At one extreme, sample-and-aggregate makes few evaluations but multiplies the sample size; at the other, the sample size grows only additively, but the number of evaluations is exponential in this overhead.
Their scheme interpolates between the two, with lower bounds on the number of evaluations that show it to be nearly optimal.
\citet{brown2026monotone} use the monotonicity of a statistic, whose value on a subset of the data never exceeds its value on the whole data, to save a factor in the sample size of sample-and-aggregate at the price of a running time exponential in that factor, with a matching lower bound on the number of queries.
Both target the statistical accuracy of the estimate on i.i.d.\ data, and ask nothing of the statistic beyond its accuracy on subsamples, or its monotonicity.
PFR instead asks for a concave score and a friendly-DP release mechanism, whose output need not be a scalar, and its accuracy is analyzed in each application, in some of them under deterministic conditions on the data rather than i.i.d.\ sampling.
Their numbers of evaluations count calls to the statistic, whereas the $\nslots\cdot\nblk$ evaluations by PFR are calls to the score.

\paragraph{Stable estimators and subsample-test-aggregate.}\label{sec:bhs_comparison}
\citet{BrownHS23} stabilize their covariance estimate by averaging nested largest subsets of records with bounded leverage, and privately test a score that tracks how many records must be excluded, with an analogous rule based on pairwise distances for the mean; they also release Gaussian samples from the stabilized covariance.
Their results are concrete and strong, but they do not state a general theorem for every concave score and every friendly-DP release mechanism.
For subspace estimation, their published relative covariance guarantee depends polynomially on $\dimn$.
To our knowledge, our guarantee on the geometric spectrum (\Cref{thm:subspace_informal}), which depends on $\dimn$ only through the condition $\rank>\lceil\log_2(\dimn)\rceil$, is new among published bounds.
We expect that an adaptation of their construction to a fixed public ridge, which is not a published theorem, could release the same regularized covariance sketches on typical Gaussian data and recover the same guarantee as \Cref{thm:subspace_informal}.
Our guarantee thus need not separate PFR from all the adaptations of existing techniques, and comparing the two routes, a general theorem with a concave score and an estimator-specific stability analysis, is a natural next step.
For the histograms, a stabilization of the coordinatewise logarithms in the style of \citet{BrownHS23}, also not a published theorem, may give a guarantee comparable to that of \Cref{thm:histogram_kl_utility}.
The subsample-test-aggregate (STA) reduction of \citet{rubinstein2026down} also starts from one-sided stability under deletions: the output of a vector-valued learner on an accepted subsample, perturbed with Gaussian noise whose covariance a certificate provides, is private~\cite[Lemma~28]{rubinstein2026down}.
Such draws can be aggregated arbitrarily, with a privacy cost that grows with their number~\cite[Theorem~30]{rubinstein2026down}, after a private check of the acceptance rate~\cite[Theorem~17]{rubinstein2026down}.
PFR instead filters individual records, given a concave score, and applies one friendly-DP release mechanism to all the retained data together.
This difference in structure does not by itself lower the privacy cost, since the release may also consist of many draws, such as the $\nsk$ sketches of \Cref{thm:filt_sketch_DP}, whose number enters its privacy condition.
Like PFR, both approaches exploit typical rather than worst-case stability; a concrete open task is to compare their general hypotheses, their privacy and sample overheads, and the cost of evaluating their certificates with those of PFR.

%% file: sections/apd_preliminaries.tex
\section{Additional Preliminaries}\label{app:prelims}

This appendix gives the notational conventions omitted from \Cref{sec:prelims}, states formally the noise mechanisms and composition tools recalled there, and collects additional standard tools used in the proofs of \Cref{app:outlier-rejection,app:generalized-friendly-core,app:friendly-mechanisms}.
We use the notation of \Cref{sec:prelims}, and the notation for vectors and matrices below.

\phantomsection\paragraph{Notation and conventions.}\label{app:notation}

The set $\dom^*\coloneqq\bigcup_{\nslots\ge0}\dom^\nslots$ of all datasets includes the empty dataset $\emptyset$.
Since a null record occupies its slot, the all-null dataset $\nul^\nslots$ has size $\nslots$ and is not the empty dataset when $\nslots>0$.
We denote by $\ds^{i\to\cand}$ the dataset obtained from $\ds$ by replacing $\ds_i$ with $\cand\in\dom$.
When $\ds_i=\nul$ and $\cand\ne\nul$, this restores a record to a null slot, and $\ds^{i\to\cand}$ and $\ds$ are neighbors.
The symbol $\fail$ denotes a failure output of a mechanism, and is distinct from $\nul$.
We denote by $X\sim \dist$ a draw from a distribution $\dist$, by $\dist^\nslots$ the distribution of $\nslots$ independent draws from it, and by $\Bin(N,p)$ the binomial distribution.
The centered Laplace distribution $\Lap(\sigma)$ has density $(2\sigma)^{-1}e^{-\vert z\vert/\sigma}$, and all logarithms are natural.

For vectors and matrices, which the applications of \Cref{sec:applications} use, data vectors are column vectors in $\bbR^\dimn$, $\lVert\cdot\rVert$ is the Euclidean norm of vectors and the operator norm of matrices, and $A^T$, $A^\dagger$ and $\Tr(A)$ are the transpose, the pseudoinverse and the trace of a matrix $A$.
For symmetric matrices, $A\preceq B$ means that $B-A$ is positive semidefinite, $\lambda_i(A)$ is the $i^{th}$ largest eigenvalue of $A$, $A^{1/2}$ is the positive semidefinite square root of $A\succeq0$, and $|A|\coloneqq(A^2)^{1/2}$.
Without an index, $\lambda$ denotes the public regularizer of the application at hand, such as a ridge or a pseudocount.
For a dataset of vectors, we also use $\ds$ for the matrix whose rows are $\ds_i^T$, so $\ds^T\ds=\sum_i \ds_i \ds_i^T$.
In such sums and second moments, and in the empirical trace of \Cref{sec:trace}, a null record is represented by a zero row and is still counted in the size of the dataset.
This convention does not identify $\nul$ with an ordinary zero record, such as a zero vector: an ordinary zero is a data value that can itself be nulled, whereas $\nul$ marks a withheld or rejected record.

\phantomsection\paragraph{Zero-out and replacement adjacency.}\label{app:adjacency}

The following claim relates zero-out adjacency (\Cref{sec:prelims}) to replacement adjacency on the same domain $\dom$, which includes $\nul$.

\begin{claim}[Zero-out and replacement adjacency]\label[claim]{claim:zero_out_replacement}
    Given $\nslots\in\bbN$; $\eps_0\ge0$; $\delta_0\in[0,1]$; and a mechanism $\mech$ on $\dom^\nslots$, the following hold.
    \begin{enumerate}
        \item If $\mech(\ds)\approx_{\eps_0,\delta_0}\mech(\ds^{i\to y})$ for every $\ds\in\dom^\nslots$, $i\in[\nslots]$ and $y\in\dom$, then $\mech$ is $(\eps_0,\delta_0)$-DP.
        \item If $\mech$ is $(\eps_0,\delta_0)$-DP, then $\mech(\ds)\approx_{2\eps_0,(1+e^{\eps_0})\delta_0}\mech(\ds^{i\to y})$ for every $\ds\in\dom^\nslots$, $i\in[\nslots]$ and $y\in\dom$.
    \end{enumerate}
\end{claim}

\begin{proof}
    The first item holds since every pair of neighboring datasets has the form $(\ds,\ds^{i\to\nul})$, which is a replacement pair with $y=\nul$.
    For the second, given $\ds$, $i$ and $y$, denote $w\coloneqq\ds^{i\to\nul}=(\ds^{i\to y})^{i\to\nul}$, so each of $\ds$ and $\ds^{i\to y}$ either equals $w$ or is a neighbor of it.
    For every event $O$, the two privacy inequalities through $w$ give
    \[
        \Pr[\mech(\ds)\in O]
        \le e^{\eps_0}\Pr[\mech(w)\in O]+\delta_0
        \le e^{2\eps_0}\Pr[\mech(\ds^{i\to y})\in O]+(1+e^{\eps_0})\delta_0,
    \]
    and exchanging the roles of $\ds$ and $\ds^{i\to y}$ gives the other inequality.
\end{proof}

For $\eps\le1$, the second item with $\eps_0=\eps/2$ and $\delta_0=\delta/3$ gives replacement privacy with parameters $(\eps,\delta)$, since $1+e^{1/2}<3$.

\phantomsection\paragraph{Noise mechanisms and composition.}\label{app:noise}

We state the Gaussian and Laplace mechanisms for two fixed means, which is the form in which the proofs use them.

\begin{fact}[Gaussian mechanism, Theorem~8 in \cite{balle2018improving}]\label[fact]{fact:gauss_profile}
    Given $\sigma,\eps > 0$ and $\mu,\mu'\in\bbR$ with $\vert\mu-\mu'\vert\le1$, we have $\Gauss(\mu,\sigma^{2})\approx_{\eps,\gaussprof(\eps)}\Gauss(\mu',\sigma^{2})$, where $\gaussprof(\eps) \coloneqq \Phi(1/(2\sigma)-\eps\sigma) - e^{\eps} \Phi(-1/(2\sigma)-\eps\sigma)$ and $\Phi$ is the standard Gaussian CDF.

    In particular, if the sensitivity of a function $\stat : \dom^{*} \rightarrow \bbR$ is bounded by $1$, then the Gaussian noise addition mechanism $\Gauss_{\sigma}(\stat)$, which outputs $\stat(\ds)+\Gauss(0,\sigma^{2})$ on input $\ds$, is $(\eps,\gaussprof(\eps))$-DP.
\end{fact}

\citet{balle2018improving} state this bound for mechanisms.
The form for two fixed means follows by applying it to a function that is defined on two neighboring datasets and takes the values $\mu$ and $\mu'$ on them.

\begin{fact}[Laplace mechanism~\cite{dwork2006calibrating}]\label[fact]{fact:laplace}
    Given $\sigma > 0$ and $\mu,\mu'\in\bbR$ with $\vert\mu-\mu'\vert\le1$, we have $\mu+\Lap(\sigma)\approx_{1/\sigma,0}\mu'+\Lap(\sigma)$.
    In particular, if the sensitivity of a real-valued function $\qry$ on datasets is bounded by $1$, then $\qry(\cdot) + \Lap(\sigma)$ is $(1/\sigma, 0)$-DP.
    Moreover, for every $a \ge 0$,
    \[
        \Pr\left[\Lap(\sigma) > a\right]
        =\Pr\left[\Lap(\sigma) < -a\right]
        =\tfrac12e^{-a/\sigma}
        \qquad\text{and}\qquad
        \Pr\left[\vert \Lap(\sigma) \vert > a\right] = e^{-a/\sigma}.
    \]
\end{fact}

\begin{proof}
    The densities of the two shifted distributions have ratio $e^{(\vert z-\mu'\vert-\vert z-\mu\vert)/\sigma}\le e^{\vert\mu-\mu'\vert/\sigma}\le e^{1/\sigma}$ at every $z$, so integrating over an event gives both inequalities of \Cref{def:dp} with $\delta=0$.
    The tails follow by integrating the density $(2\sigma)^{-1}e^{-\vert z\vert/\sigma}$.
\end{proof}

\begin{fact}[Basic composition~\cite{dwork2014algorithmic}]
Basic composition adds up the privacy parameters of an adaptive interaction: if each round $j\in[N]$ is $(\eps_j,\delta_j)$-DP given the preceding outputs, then the interaction is $\bigl(\sum_{j=1}^{N}\eps_j,\sum_{j=1}^{N}\delta_j\bigr)$-DP.
\end{fact}

\begin{fact}[Advanced composition, Theorem~3.4 in \cite{kairouz2015composition}]\label[fact]{fact:advanced_composition}
    Given $N\in\bbN$; $\zeta>0$; $\hat\delta\in[0,1]$; $\delta_{0}\in(0,1)$; and an adaptive interaction of at most $N$ rounds on a private dataset, in which the mechanism of a round may depend on all preceding outputs, if this mechanism is $(\zeta,\hat\delta)$-DP after every preceding transcript, then the full transcript is $(\eps_{\mathrm{comp}},N\hat\delta+\delta_{0})$-DP, where $\eps_{\mathrm{comp}}\coloneqq\zeta\sqrt{2N\ln(1/\delta_{0})}+N\zeta(e^{\zeta}-1)/(e^{\zeta}+1)$.
    In particular, since $(e^{\zeta}-1)/(e^{\zeta}+1)\le\zeta/2$, if $\hat\delta=0$ then the full transcript is $\bigl(N\zeta^{2}/2+\zeta\sqrt{2N\ln(1/\delta_{0})},\delta_{0}\bigr)$-DP.
    Equivalently, the same conclusions hold when, in every round, the two possible output distributions are $(\zeta,\hat\delta)$-indistinguishable.
\end{fact}

\citet{kairouz2015composition} state this bound for exactly $N$ rounds, with second parameter $1-(1-\hat\delta)^{N}(1-\delta_{0})\le N\hat\delta+\delta_{0}$, and an interaction with fewer rounds is padded with rounds that output a constant.
The first advanced composition theorem, of \citet{dwork2010boosting}, has $N\zeta(e^{\zeta}-1)$ in place of $N\zeta(e^{\zeta}-1)/(e^{\zeta}+1)$.

\phantomsection\paragraph{Two-stage composition.}\label{app:two_stage}

The next proposition is the form of adaptive basic composition that our proofs use: a first stage with countably many outcomes, followed by a second stage whose distribution depends on the outcome of the first.

\begin{proposition}[Two-stage composition]\label[proposition]{prop:two_stage}
    Given $\eps_1,\eps_2\ge0$; $\delta_1,\delta_2\in[0,1]$; random variables $\mathbf{A},\mathbf{A}'$ over a countable set $U$ such that $\mathbf{A}\approx_{\eps_{2},\delta_{2}}\mathbf{A}'$, and distributions $P_{u},P'_{u}$ such that $P_{u}\approx_{\eps_{1},\delta_{1}}P'_{u}$ for each $u\in U$, if $\mathbf{B},\mathbf{B}'$ are random variables such that $\mathbf{B}$ has distribution $P_{u}$ conditionally on $\mathbf{A}=u$, and $\mathbf{B}'$ has distribution $P'_{u}$ conditionally on $\mathbf{A}'=u$, whenever these events have positive probability, then $(\mathbf{A},\mathbf{B})\approx_{\eps_{1}+\eps_{2},\,\delta_{1}+\delta_{2}}(\mathbf{A}',\mathbf{B}')$, and in particular $\mathbf{B}\approx_{\eps_{1}+\eps_{2},\,\delta_{1}+\delta_{2}}\mathbf{B}'$.
\end{proposition}

\begin{proof}
    By symmetry, it suffices to prove one of the two inequalities of \Cref{def:dp}.
    Let $O$ be an event and write $O_{u}\coloneqq\{y:(u,y)\in O\}$.
    For every $u\in U$,
    \[
        P_{u}(O_{u})\le\min\left\{1,e^{\eps_{1}}P'_{u}(O_{u})+\delta_{1}\right\}\le w(u)+\delta_{1},
        \qquad\text{where }
        w(u)\coloneqq\min\left\{1,e^{\eps_{1}}P'_{u}(O_{u})\right\}.
    \]
    Since $w$ takes values in $[0,1]$,
    \[
        \bbE[w(\mathbf{A})]=\int_{0}^{1}\Pr[w(\mathbf{A})>y]\,dy
        \le\int_{0}^{1}\left(e^{\eps_{2}}\Pr[w(\mathbf{A}')>y]+\delta_{2}\right)dy
        =e^{\eps_{2}}\bbE[w(\mathbf{A}')]+\delta_{2},
    \]
    where the inequality is the indistinguishability of $\mathbf{A}$ and $\mathbf{A}'$, applied to the event $\{u:w(u)>y\}$.
    Combining the two displays,
    \[
        \Pr[(\mathbf{A},\mathbf{B})\in O]
        =\sum_{u\in U}\Pr[\mathbf{A}=u]\,P_{u}(O_{u})
        \le\bbE[w(\mathbf{A})]+\delta_{1}
        \le e^{\eps_{2}}\bbE[w(\mathbf{A}')]+\delta_{1}+\delta_{2},
    \]
    and $\bbE[w(\mathbf{A}')]\le e^{\eps_{1}}\sum_{u\in U}\Pr[\mathbf{A}'=u]\,P'_{u}(O_{u})=e^{\eps_{1}}\Pr[(\mathbf{A}',\mathbf{B}')\in O]$.
    The claim about $\mathbf{B}$ and $\mathbf{B}'$ follows by post-processing.
\end{proof}

\phantomsection\paragraph{R\'enyi divergence.}\label{app:renyi}

The privacy of the Gaussian sketches (\Cref{thm:filt_sketch_DP}) is analyzed through R\'enyi divergence.

\begin{definition}[R\'enyi divergence]\label[definition]{def:renyi}
    Given an order $\rord > 1$ and distributions $P \ll Q$ over a common measurable space, the R\'enyi divergence of order $\rord$ is $\Renyi{\rord}(P \Vert Q)\coloneqq\frac{1}{\rord - 1}\ln\Bigl(\bbE_{z \sim Q}\bigl[\bigl(\frac{dP}{dQ}(z)\bigr)^{\rord}\bigr]\Bigr)$.
    Given also $\rho\ge0$, a mechanism $\mech$ will be called \emph{$(\rord, \rho)$-RDP} if $\Renyi{\rord}(\mech(\ds) \Vert \mech(\ds')) \le \rho$ for every pair of neighboring datasets $\ds\simeq\ds'$, and \emph{$\rho$-zCDP}~\cite{bun2016concentrated} if it is $(\rord', \rho \rord')$-RDP for every $\rord' > 1$.
\end{definition}

\begin{proposition}[RDP to DP conversion, Proposition~12 in \cite{canonne2020discrete}]\label[proposition]{prop:rdp_to_dp}
    Given $\rord > 1$; $\rho\ge0$; and distributions $P,Q$ over a common measurable space, if $\Renyi{\rord}(P \Vert Q) \le \rho$ and $\Renyi{\rord}(Q \Vert P) \le \rho$, then for every $\eps > \rho$ we have $P\approx_{\eps,\delta}Q$, where $\delta\coloneqq e^{(\rord-1)(\rho - \eps)}(1 - 1/\rord)^{\rord}/(\rord - 1)$.
    In particular, $P\approx_{\eps',\delta}Q$ for every $\delta\in(0,1)$ and $\eps'\coloneqq\rho+\ln(1/\delta)/(\rord-1)$.
\end{proposition}

\citet{canonne2020discrete} state this bound for mechanisms, and it applies to two fixed distributions through a mechanism that outputs them on two neighboring datasets.

\phantomsection\paragraph{Concentration.}\label{app:concentration}

\begin{proposition}[Multiplicative Chernoff bound, see, e.g., \cite{mitzenmacher2005probability}]\label[proposition]{prop:multiplicative_chernoff}
    Given $N\in\bbN$ and independent Bernoulli random variables $X_{1},\ldots,X_{N}$, we have
    \[
        \Pr\left[X\le(1-u)\mu\right]
        \le \exp\left(-\frac{u^{2}\mu}{2}\right)
        \quad(0\le u\le1),
        \quad
        \Pr\left[X\ge(1+u)\mu\right]
        \le \exp\left(-\frac{u^{2}\mu}{2+u}\right)
        \quad(u\ge0),
    \]
    where $X\coloneqq\sum_{i=1}^{N}X_{i}$ and $\mu\coloneqq\bbE[X]$.
    In particular, for every $p\in[0,1]$, these bounds apply to $X\sim\Bin(N,p)$ with $\mu=Np$.
\end{proposition}

The bounds are standard, and we include the short proof because the second one is usually stated in a slightly different form.

\begin{proof}
    Let $p_{i}$ be the parameter of $X_{i}$, so that $\mu=\sum_{i}p_{i}$.
    For every $\zeta\in\bbR$, independence and the inequality $1+y\le e^{y}$ give
    \[
        \bbE\left[e^{\zeta X}\right]
        =\prod_{i=1}^{N}\left(1+p_{i}(e^{\zeta}-1)\right)
        \le \exp\left(\mu(e^{\zeta}-1)\right).
    \]
    Both bounds are trivial for $u=0$.
    For the upper tail, let $u>0$ and $\zeta\coloneqq\ln(1+u)>0$.
    Markov's inequality for $e^{\zeta X}$ gives
    \[
        \Pr\left[X\ge(1+u)\mu\right]
        \le \bbE\left[e^{\zeta X}\right]e^{-\zeta(1+u)\mu}
        \le \exp\left(\mu\left(u-(1+u)\ln(1+u)\right)\right)
        \le \exp\left(-\frac{u^{2}\mu}{2+u}\right),
    \]
    where the last inequality uses $\ln(1+u)\ge2u/(2+u)$, which holds for $u\ge0$ since both sides vanish at $0$ and the derivative of the left-hand side, $1/(1+u)$, is at least that of the right-hand side, $4/(2+u)^{2}$.

    For the lower tail, let $u\in(0,1)$ and $\zeta\coloneqq\ln(1-u)<0$.
    Markov's inequality for $e^{\zeta X}$ gives
    \[
        \Pr\left[X\le(1-u)\mu\right]
        \le \bbE\left[e^{\zeta X}\right]e^{-\zeta(1-u)\mu}
        \le \exp\left(\mu\left(-u-(1-u)\ln(1-u)\right)\right)
        \le \exp\left(-\frac{u^{2}\mu}{2}\right),
    \]
    where the last inequality uses $(1-u)\ln(1-u)\ge-u+u^{2}/2$, which holds on $[0,1)$ since both sides vanish at $0$ and the derivative of the difference is $-\ln(1-u)-u\ge0$.
    Finally, for $u=1$ we have $\Pr[X\le0]=\prod_{i}(1-p_{i})\le e^{-\mu}\le e^{-\mu/2}$.
\end{proof}

Two proofs below count indicators that are not independent, but whose conditional probabilities given the past are bounded.
The following coupling reduces them to binomial counts.

\begin{lemma}[Adapted Bernoulli sequences]\label[lemma]{lem:domination}
    Given $N\in\bbN$; $\underline p\in[0,1]$; $\{0,1\}$-valued random variables $X_{1},\ldots,X_{N}$ and $\sigma$-algebras $\mathcal{G}_{0}\subseteq\cdots\subseteq\mathcal{G}_{N}$ such that $X_{j}$ is measurable with respect to $\mathcal{G}_{j}$, the following hold.
    \begin{enumerate}
        \item If $\Pr[X_{j}=1\mid\mathcal{G}_{j-1}]\le \underline p$ almost surely for every $j$, then $\Pr\left[\sum_{j}X_{j}\ge y\right]\le\Pr\left[\Bin(N,\underline p)\ge y\right]$ for every $y\in\bbR$.
        \item If $\Pr[X_{j}=1\mid\mathcal{G}_{j-1}]\ge \underline p$ almost surely for every $j$, then $\Pr\left[\sum_{j}X_{j}< y\right]\le\Pr\left[\Bin(N,\underline p)< y\right]$ for every $y\in\bbR$.
    \end{enumerate}
\end{lemma}

\begin{proof}
    We prove the first item; the second is identical with the inequalities reversed.
    For $j\in[N]$ and $w\in\{0,1\}^{j-1}$, let $p_{j}(w)\coloneqq\Pr[X_{j}=1\mid(X_{1},\ldots,X_{j-1})=w]$ if the conditioning event has positive probability, and $p_{j}(w)\coloneqq \underline p$ otherwise.
    Since $(X_{1},\ldots,X_{j-1})$ is measurable with respect to $\mathcal{G}_{j-1}$, the tower property gives $p_{j}(w)\le \underline p$.

    Let $U_{1},\ldots,U_{N}$ be independent and uniform on $[0,1]$, and define recursively $\widetilde X_{j}\coloneqq\ind\{U_{j}\le p_{j}(\widetilde X_{1},\ldots,\widetilde X_{j-1})\}$ and $Y_{j}\coloneqq\ind\{U_{j}\le \underline p\}$.
    By the chain rule, $(\widetilde X_{1},\ldots,\widetilde X_{N})$ has the distribution of $(X_{1},\ldots,X_{N})$.
    The $Y_{j}$ are independent Bernoulli variables with parameter $\underline p$, and $\widetilde X_{j}\le Y_{j}$ for every $j$.
    Hence $\Pr\bigl[\sum_{j}X_{j}\ge y\bigr]=\Pr\bigl[\sum_{j}\widetilde X_{j}\ge y\bigr]\le\Pr\bigl[\sum_{j}Y_{j}\ge y\bigr]$.
\end{proof}

%% file: sections/apd_filtering.tex
\section{\texorpdfstring{Proofs from \Cref{sec:outlier_rej}}{Proofs from Private Filtering}}\label{app:outlier-rejection}

This appendix contains the proofs that are omitted or only outlined in \Cref{sec:outlier_rej}, with the long results restated before their proofs, and the retention guarantee for i.i.d.\ data, which is stated only here.
The standard tools are those of \Cref{sec:prelims,app:prelims}.

We use two statements from the target-charging framework of \citet{cohen2023target}, specialized to pure-DP Boolean mechanisms and written in our notation.
\Cref{lem:cl_boolean_not_prior} is the decomposition that the proof of \citet[Lemma~C.3]{cohen2023target} gives for private tests with Boolean output, and \Cref{lem:cl_target_charging} is the pure Boolean specialization of their target-charging theorem~\cite[Theorem~B.4]{cohen2023target}, stated with the composition bound of \Cref{fact:advanced_composition}.
Their proofs are included for completeness.

\begin{lemma}[Boolean tests, following Lemma~C.3 in \cite{cohen2023target}]\label[lemma]{lem:cl_boolean_not_prior}
    Given $\zeta > 0$ and distributions $\mathbf{P}^{0}, \mathbf{P}^{1}$ over $\{0,1\}$ satisfying $\mathbf{P}^{0} \approx_{\zeta,0} \mathbf{P}^{1}$, there exist a weight $\pi \in [0,1]$ and distributions $\mathbf{B}^{0}, \mathbf{B}^{1}$ such that, for $z \in \{0,1\}$,
    \[
        \mathbf{P}^{z}
        = \pi \mathbf{0} + (1-\pi)\mathbf{B}^{z},
        \qquad
        \mathbf{B}^{0} \approx_{\zeta,0} \mathbf{B}^{1},
        \qquad
        \Pr\left[\mathbf{B}^{z}=1\right]
        \ge \underline p_{\zeta} \coloneqq \frac{1}{e^{\zeta}+1},
    \]
    where $\mathbf{0}$ denotes the distribution that takes $0$ with probability $1$.
\end{lemma}
\begin{proof}
	For $z \in \zo$, write $p_{z}\coloneqq\Pr\left[\mathbf{P}^{z}=1\right]$, and assume without loss of generality that $p_{0}\le p_{1}$.
	The assumption $\mathbf{P}^{0} \approx_{\zeta,0} \mathbf{P}^{1}$ implies that $p_{1}\le e^{\zeta}p_{0}$ and $1-p_{0}\le e^{\zeta}(1-p_{1})$.
	
	If $p_{0}\ge \underline p_{\zeta}$, take $\pi=0$ and $\mathbf{B}^{z}=\mathbf{P}^{z}$.
	All the conclusions are immediate.
	If $p_{0}=0$, the first inequality above forces $p_{1}=0$; in this degenerate case take $\pi=1$ and let both $\mathbf{B}^{0}$ and $\mathbf{B}^{1}$ be the constant $1$.
	
	It remains to consider $0<p_{0}<\underline p_{\zeta}$.
	In the following, define
	\[
	d\coloneqq\frac{e^{\zeta}p_{1}-p_{0}}{e^{\zeta}-1},
	\qquad
	\pi\coloneqq1-d,
	\qquad
	\Pr\left[\mathbf{B}^{z}=1\right]
	\coloneqq\frac{p_{z}}{d}
	\quad\text{for } z\in\{0,1\}.
	\]
	Since $p_0\le p_1$ and $1-p_{0}\le e^{\zeta}(1-p_{1})$, respectively, we have
	\[
	d-p_{1}=\frac{p_{1}-p_{0}}{e^{\zeta}-1}\ge0
	\qquad\text{and}\qquad
	1-d=\frac{e^{\zeta}(1-p_{1})-(1-p_{0})}{e^{\zeta}-1}\ge0.
	\]
	These two inequalities give $p_{1}\le d\le1$, meaning that $\mathbf{B}^{0}$ and $\mathbf{B}^{1}$ are well-defined, and $\pi\mathbf{0}+(1-\pi)\mathbf{B}^{z}=\mathbf{P}^{z}$ because both sides assign probability $p_z$ to the outcome $1$.
	
	The inequality $p_{1}\le e^{\zeta}p_{0}$ also gives $d\le(e^{2\zeta}p_{0}-p_{0})/(e^{\zeta}-1)=p_{0}(e^{\zeta}+1)$, and hence $\Pr\left[\mathbf{B}^{0}=1\right]=p_{0}/d\ge1/(e^{\zeta}+1)=\underline p_{\zeta}$.
	Since $p_{1}\ge p_{0}$, the same lower bound holds for $\mathbf{B}^{1}$.
	Finally, $\Pr[\mathbf{B}^{1}=1]/\Pr[\mathbf{B}^{0}=1]=p_{1}/p_{0}\le e^{\zeta}$, while the definition of $d$ gives $\Pr\left[\mathbf{B}^{0}=0\right]=e^{\zeta}\Pr\left[\mathbf{B}^{1}=0\right]$, as required.
\end{proof}

\begin{lemma}[Target charging, following Theorem~B.4 in \cite{cohen2023target}]\label[lemma]{lem:cl_target_charging}
    Given $\nslots,\svcap\in\bbN$; $\zeta>0$; $\underline p\in(0,1]$; $a>0$; $\delta_{0}\in(0,1)$; and an adaptive interaction of at most $\nslots$ Boolean rounds with one of two neighboring datasets that halts after its $\svcap^{th}$ answer equal to $1$, if, after every preceding transcript, the distributions of the next answer on the two datasets are $\pi \mathbf{0} + (1-\pi)\mathbf{B}^{0}$ and $\pi \mathbf{0} + (1-\pi)\mathbf{B}^{1}$ for a common weight $\pi\in[0,1]$, the constant-zero distribution $\mathbf{0}$, and distributions $\mathbf{B}^{0}, \mathbf{B}^{1}$ over $\zo$ that satisfy $\mathbf{B}^{0} \approx_{\zeta,0} \mathbf{B}^{1}$ and $\min_{z}\Pr[\mathbf{B}^{z}=1] \ge \underline p$, then the transcripts on the two datasets are $(\eps_{\mathrm{tc}},\delta_{0}+\delta^{*}(a))$-indistinguishable, where $W(a)\coloneqq\lceil(1+a)\svcap/\underline p\rceil$, $\delta^{*}(a)\coloneqq\exp\bigl(-a^{2}\svcap/(2(1+a))\bigr)$ and $\eps_{\mathrm{tc}}\coloneqq W(a)\zeta^{2}/2+\zeta\sqrt{2W(a)\ln(1/\delta_{0})}$.
\end{lemma}
\begin{proof}
		Let $z\in\{0,1\}$ be the private bit specifying which of the two neighboring datasets is used.
		We describe a simulator whose released transcript has exactly the law of the original interaction on dataset $z$.
		Conditional on the transcript so far, take the decomposition in the statement, output $0$ with probability $\pi$, independently of $z$, and otherwise draw from $\mathbf{B}^{z}$.
		We call the latter draw a \emph{private call}.
		The mixture identity shows, inductively over the rounds, that the simulator faithfully reproduces the original transcript.
		Let $\mathbf{V}^{z}$ denote the simulator's \emph{full view} on dataset $z$: for each round until termination, it records whether the simulator chose the constant-zero branch or made a private call, together with the resulting answer bit.
		Let $\mathbf{T}^{z}$ denote the released transcript on that dataset, consisting only of the answer bits observed in the original interaction.
		Thus $\mathbf{T}^{z}$ is obtained from $\mathbf{V}^{z}$ by deleting the branch indicators, and is therefore a post-processing of $\mathbf{V}^{z}$.
		
		For each $z$, let $H^{z}$ denote the number of private calls made before the interaction halts, and define the nonfailure event $G^{z}\coloneqq\left\{H^{z}\le W(a)\right\}$.
		Conditioned on every preceding simulated history, a private call returns $1$ with probability at least $\underline p$, whereas a constant-zero draw never returns $1$.
		For $j\in[W(a)]$, let $X_{j}$ be the answer of the $j^{th}$ private call if the interaction makes at least $j$ private calls, and an independent Bernoulli variable with parameter $\underline p$ otherwise.
		If $H^{z}>W(a)$, then the interaction did not halt during its first $W(a)$ private calls, so these calls returned fewer than $\svcap$ ones; that is, $\sum_{j}X_{j}<\svcap$.
		Each $X_{j}$ equals $1$ with probability at least $\underline p$ conditionally on the history that precedes it, so \Cref{lem:domination} gives $\Pr[H^{z}>W(a)]\le\Pr[Y<\svcap]$ for $Y\sim\Bin(W(a),\underline p)$.
		
		Let $\mu\coloneqq\bbE[Y]=W(a)\underline p\ge(1+a)\svcap$.
		Applying the multiplicative Chernoff bound (\Cref{prop:multiplicative_chernoff}) with $u=a/(1+a)$ gives, for each $z\in\{0,1\}$,
		\begin{align*}
			\Pr[\neg G^{z}]
			=\Pr[H^{z}>W(a)]
			&\le\Pr[Y<\svcap]
			\le\Pr\left[Y<\left(1-\frac{a}{1+a}\right)\mu\right] \\
			&\le\exp\left(-\frac{a^{2}\mu}{2(1+a)^{2}}\right)
			\le\exp\left(-\frac{a^{2}\svcap}{2(1+a)}\right)
			=\delta^{*}(a).
		\end{align*}

    To apply composition, modify the simulator so that it stops just before making private call $W(a)+1$ and returns a fresh failure symbol $\fail$.
    If it finishes without exceeding this limit, it returns its full view.
    Denote its output on dataset $z$ by $\widehat{\mathbf V}^{z}$.
    Between private calls, the branch choices and constant-zero answers depend only on the preceding view and fresh randomness.
    This modified simulator makes at most $\min\{W(a),\nslots\}$ pure $(\zeta,0)$-DP calls, interspersed with processing that does not access $z$.
    Grouping each private call with the steps that follow the previous private call gives an adaptive interaction of at most $W(a)$ rounds, each of which is $(\zeta,0)$-DP given the preceding view, and the steps after the last private call do not access $z$.
    \Cref{fact:advanced_composition} with $N=W(a)$ and $\hat\delta=0$, followed by post-processing, gives $\widehat{\mathbf V}^{0}\approx_{\eps_{\mathrm{tc}},\delta_0}\widehat{\mathbf V}^{1}$.
    This argument does not condition on $G^z$ and does not require $W(a)\le \nslots$.

    Couple the two simulators on each dataset using the same randomness.
    For every event $O$ of original full views, with $\fail\notin O$, we have $\Pr[\widehat{\mathbf V}^z\in O]=\Pr[\mathbf V^z\in O, G^z]$.
    Therefore, for either $z\in\{0,1\}$,
    \begin{align*}
        \Pr[\mathbf V^z\in O]
        &\le \Pr[\widehat{\mathbf V}^z\in O]+\delta^{*}(a)\\
        &\le e^{\eps_{\mathrm{tc}}}\Pr[\widehat{\mathbf V}^{1-z}\in O]+\delta_0+\delta^{*}(a)\\
        &\le e^{\eps_{\mathrm{tc}}}\Pr[\mathbf V^{1-z}\in O]+\delta_0+\delta^{*}(a).
    \end{align*}
    Post-processing gives the claimed guarantee for the released transcripts $\mathbf T^0,\mathbf T^1$.
\end{proof}

We now apply these lemmas directly to \Cref{prop:sparse_vec_priv}, restated below.

\begin{proposition}[Restatement of \Cref{prop:sparse_vec_priv}]
    \leavevmode\phantomsection\label[proposition]{prop:sparse_vec_priv_restated}
    \SVProposition
\end{proposition}
\begin{proof}
    Denote $J\coloneqq\ln(2/\delta)$, so that $\svscale=7\sqrt{\svcap J}/\eps$.
    The assumptions $\svcap\ge12J$ and $\eps\le4J$ imply $\svscale\ge7\sqrt{12}/4>6$, hence $\underline p_{1/\svscale}=1/(e^{1/\svscale}+1)\ge4/9$.

    Fix two neighboring datasets.
    Conditional on any preceding transcript, the next bit of \Cref{alg:sparse_vec} is a threshold post-processing of a sensitivity-one Laplace mechanism, so its two laws are $(1/\svscale,0)$-indistinguishable.
    \Cref{lem:cl_boolean_not_prior} supplies the decompositions required by \Cref{lem:cl_target_charging}, with $\zeta=1/\svscale$ and $\underline p=\underline p_{1/\svscale}$.
    Applying the latter lemma with the fixed choices $a=1$ and $\delta_0=\delta/2$ gives $(\eps_{\mathrm{tc}},\delta_{\mathrm{tc}})$-indistinguishability, where, since $\underline p_{1/\svscale}\ge4/9$ and $\svcap\ge12\ln(2)>2$,
    \[
        W\coloneqq\left\lceil\frac{2\svcap}{\underline p_{1/\svscale}}\right\rceil\le5\svcap,
        \qquad
        \eps_{\mathrm{tc}}=\frac{W}{2\svscale^2}
             +\frac{\sqrt{2WJ}}{\svscale},
        \qquad
        \delta_{\mathrm{tc}}\coloneqq\frac\delta2+e^{-\svcap/4}.
    \]
    Substituting the noise scale and using $\eps\le4J$ yields
    \[
        \eps_{\mathrm{tc}}
        \le\frac{5\svcap}{2\svscale^2}+\frac{\sqrt{10\svcap J}}{\svscale}
        =\frac{5\eps^2}{98J}+\frac{\sqrt{10}}7\eps
        \le\left(\frac{10}{49}+\frac{\sqrt{10}}7\right)\eps
        <\eps.
    \]
    Also, $\svcap\ge12J$ gives $\delta_{\mathrm{tc}}\le\delta/2+e^{-3J}\le\delta/2+e^{-J}=\delta$.
    Since the neighboring pair was arbitrary, this proves $(\eps,\delta)$-DP.
\end{proof}

Conditional on the partition, which is drawn independently of the data, the filter runs sparse vector on $\ds$ itself, and on a non-aborting run $\ds_i$ is retained exactly when $\cnt_{i} + \qnoise_{i} > \athr$, where $\cnt_i=\qry_i(\ds)$.
Nulling $\ds_{i}$ alters exactly one block, and since $\pred$ is $\{0,1\}$-valued, this changes the value of every query on $\ds$ by at most $1$, which is the sensitivity that \Cref{prop:sparse_vec_priv} requires.
What that guarantee does \emph{not} cover is that $\qry_{i}$ is itself parameterized by $\ds_{i}$: the $i^{th}$ query changes as a function, so its answer may flip irrespective of the others.
The random cap $\ncap$ conceals the resulting one-bit shift in the number of failures.

The proof of \Cref{thm:out_reg_jdp}, restated below, uses the following accounting rule when privacy of the distributions conditioned on success has been established separately.

\begin{fact}[Failure-event accounting, Fact~2.5 in the full version of \cite{cohen2021clustering}]\label[fact]{fact:failure_event_accounting}
    Given $\eps\ge0$; $\delta_1,\delta_2\in[0,1]$; random variables $\mathbf{X}^{0},\mathbf{X}^{1}$ and events $E^{0},E^{1}$ with $\Pr[\mathbf{X}^{z}\in E^{z}]>0$ for both $z\in\{0,1\}$, if $\left(\mathbf{X}^{0}\mid \mathbf{X}^{0}\in E^{0}\right)\approx_{\eps,\delta_{1}}\left(\mathbf{X}^{1}\mid \mathbf{X}^{1}\in E^{1}\right)$ and $\Pr[\mathbf{X}^{z}\in E^{z}]\ge1-\delta_{2}$ for both $z$, then $\mathbf{X}^{0}\approx_{\eps,\delta_{1}+\delta_{2}}\mathbf{X}^{1}$.
\end{fact}
\begin{proof}
    Fix an event $O$, and denote $p_z\coloneqq\Pr[\mathbf{X}^{z}\in E^{z}]$ and $q_z\coloneqq\Pr[\mathbf{X}^{z}\in O\mid\mathbf{X}^{z}\in E^{z}]$ for $z\in\{0,1\}$.
    Splitting on $E^{1}$ gives the first inequality below, and $p_1\ge1-\delta_2$ gives the second.
    The third is the conditional guarantee $q_1\le e^{\eps}q_0+\delta_1$, the fourth uses $p_0\ge1-\delta_2$, and the last uses $p_0q_0=\Pr[\mathbf{X}^{0}\in O\cap E^{0}]$:
    \begin{align*}
        \Pr[\mathbf{X}^{1}\in O]
        &\le p_1q_1+1-p_1
        \le(1-\delta_2)q_1+\delta_2
        \le e^{\eps}(1-\delta_2)q_0+\delta_1+\delta_2\\
        &\le e^{\eps}p_0q_0+\delta_1+\delta_2
        \le e^{\eps}\Pr[\mathbf{X}^{0}\in O]+\delta_1+\delta_2.
    \end{align*}
    Exchanging the roles of $\mathbf{X}^{0}$ and $\mathbf{X}^{1}$ gives the other direction.
\end{proof}

\begin{theorem}[Restatement of \Cref{thm:out_reg_jdp}]
    \leavevmode\phantomsection\label[theorem]{thm:out_reg_jdp_restated}
    \OutlierJDPTheorem
\end{theorem}

\begin{proof}[Proof of \Cref{thm:out_reg_jdp}]
    Fix neighboring datasets $\ds$ and $\ds'\coloneqq\ds^{i\to\nul}$, and condition on their common partition.
    For $v\in\{\ds,\ds'\}$, define the fixed queries $\qry_j^v(w)\coloneqq\sum_{h\in[\nblk]}\pred(w^{(h)},v_j)$ on datasets $w$, which have sensitivity at most $1$.
    Let $A_{w,v}$ be the output of the filter on input $w$ using the query family $\qfam^v$, with slot $i$ nulled, which we call the \emph{erased} output.
    The two query families agree except at $j=i$, and we compare $A_{\ds,\ds}$ with $A_{\ds',\ds'}$ through $A_{\ds',\ds}$ in one direction and through $A_{\ds,\ds'}$ in the other, changing first the input and then the query family.

    \emph{Changing the input, with the queries fixed.}
    Denote $\Lconst\coloneqq\Ldef$ and $\bcap\coloneqq\lceil\cBoundJDP\Lconst/\eps\rceil$, a deterministic cap that is not an input of the algorithm.
    Since $\eps\le1$ and $\Lconst\ge\ln(4/\delta)$, we have $\bcap\ge12\ln(4/\delta)$ and $\eSVJDP\le4\ln(4/\delta)$, so \Cref{prop:sparse_vec_priv}, whose noise scale for the cap $\bcap$ and the parameters $(\eSVJDP,\delta/2)$ is the stated $\qscale$, makes $\SV_{\athr,\bcap,\qscale}$ $(\eSVJDP,\delta/2)$-DP.
    The cap $\ncap$ exceeds $\bcap$ only if $\cnoise>\bcap-\svcap\ge\cCapJDP\Lconst/\eps\ge\cscale\ln(1/\delta)$, so $\Pr[\ncap>\bcap]\le\delta/2$.
    For each cap $1\le\ncap\le\bcap$, the erased output of the filter is a common post-processing of this sparse-vector transcript: return $\nul^\nslots$ if the transcript reaches its $\ncap^{th}$ failure, and otherwise retain the passing records outside slot $i$; for $\ncap\le0$ the output is constant.

    To apply \Cref{fact:failure_event_accounting}, fix $v$ and consider the augmented outputs $\mathbf X^0\coloneqq(\ncap,A_{\ds,v})$ and $\mathbf X^1\coloneqq(\ncap,A_{\ds',v})$, where the cap in each pair is the one sampled in that run.
    Take $E^0=E^1=\{(c,y):c\le\bcap\}$, so that $\mathbf X^z\in E^z$ is exactly the event $\ncap\le\bcap$, which has probability at least $1-\delta/2>0$ on both inputs.
    The sparse-vector run with deterministic cap $\bcap$ uses test noises independent of $\ncap$, so conditioning on $\ncap\le\bcap$ leaves its law, and hence its $(\eSVJDP,\delta/2)$ privacy guarantee, unchanged.
    For each $z\in\{0,1\}$, $\left(\mathbf X^z\mid\mathbf X^z\in E^z\right)$ is obtained by the following common randomized post-processing of this sparse-vector transcript: independently sample $\ncap$ from its input-independent distribution conditioned on $\ncap\le\bcap$, reconstruct the erased filter output as above, and return it together with $\ncap$.
    Post-processing therefore gives $(\mathbf X^0\mid\mathbf X^0\in E^0)\approx_{\eSVJDP,\delta/2}(\mathbf X^1\mid\mathbf X^1\in E^1)$.
    Thus \Cref{fact:failure_event_accounting}, with privacy parameter $\eSVJDP$ and $\delta_1=\delta_2=\delta/2$, gives $\mathbf X^0\approx_{\eSVJDP,\delta}\mathbf X^1$.
    Discarding the cap by post-processing yields $A_{\ds,v}\approx_{\eSVJDP,\delta}A_{\ds',v}$ for either fixed $v$.

    \emph{Changing the query family, with the input fixed.}
    Pre-sample all $\nslots$ test noises $\qnoise_j\sim\Lap(\qscale)$, and let $\nfail^v$ be the number of failed tests under $\qfam^v$.
    Since only query $i$ changes, $|\nfail^{\ds}-\nfail^{\ds'}|\le1$, and the candidate filtered datasets agree after slot $i$ is nulled.
    Under $\qfam^v$, aborting is exactly the test $\nfail^v-\cnoise\ge\svcap$, so conditional on the test noises, the erased output is a common post-processing of a sensitivity-one Laplace mechanism with scale $\cscale=\cCapJDP/\eps$, and averaging over the test noises gives $A_{w,\ds}\approx_{\eCapJDP,0}A_{w,\ds'}$ for either fixed input $w$.

    Each direction starts from the output whose probability it bounds and changes the input first, so the factor $e^{\eCapJDP}$ of the second comparison does not multiply the additive error $\delta$ of the first.
    The two comparisons therefore give $(\eSVJDP+\eCapJDP,\delta)=(\eps,\delta)$ in both directions, and averaging over the partition proves the theorem.
\end{proof}

\begin{corollary}[Restatement of \Cref{cor:jdp_comp}]
    \leavevmode\phantomsection\label[corollary]{cor:jdp_comp_restated}
    \JDPCompositionCorollary
\end{corollary}

\begin{proof}[Proof of \Cref{cor:jdp_comp}]
    Given $\ds$ and $i$ with $\cand\coloneqq\ds_i\ne\nul$, let $A\coloneqq\filt(\ds)^{i\to\nul}$, which is $(\eps_2,\delta_2)$-indistinguishable from $\filt(\ds^{i\to\nul})$ by joint differential privacy.
    Conditional on $A=y$, the release on $\ds$ is a mixture of $\mech(y)$ and $\mech(y^{i\to\cand})$, according to whether the filter retained $\cand$, while on $\ds^{i\to\nul}$ it is $\mech(y)$, since a filter only nulls records.
    The datasets $y$ and $y^{i\to\cand}$ are neighbors, so averaging the two privacy inequalities of $\mech$ shows that these second-stage releases are $(\eps_1,\delta_1)$-indistinguishable for every $y$.
    The first stage takes finitely many values, so two-stage composition (\Cref{prop:two_stage}) proves the claim.
\end{proof}

\paragraph{Retention of the filter.}

The following corollary gives a distributional condition under which the filter of \Cref{alg:priv_filt} retains most of an i.i.d.\ sample.

\def\OutlierDistributionalAccuracyCorollary{
    Given $\nslots,\bsz\in\bbN$ with $\nslots/\bsz\in\bbN$; $\ufrac, \bfail > 0$; $\delta\in(0,1)$; $\fprob \in [\fprobMin,1)$; $\athr\ge0$; $\cscale,\qscale>0$; a distribution $\dist$ over $\dom$ and a friendliness predicate $\pred$, if $\Pr_{\ds \sim \dist^{\bsz}}[\bbE_{X \sim \dist}[\pred(\ds, X)] < 1 - \ufrac] < \bfail$, $\qscale\ln(2\nslots/\fprob) \le (\nblk-\athr)/2 - 2$ and $\ufrac \le (\nblk-\athr)\cscale\Lconst/(32\nblk\nslots)$, then with probability at least $1 - \fprob - \nblk\bfail - 2e^{-\cscale\Lconst/24}$ over $\ds \sim \dist^{\nslots}$ and the randomness of the filter, $\filt_{\fpar}(\pred;\ds)$ (\Cref{alg:priv_filt}) does not abort and rejects fewer than $\cscale\Lconst/2$ records, where $\nblk\coloneqq\nslots/\bsz$, $\Lconst\coloneqq\Ldef$ and $\svcap\coloneqq2\cscale\Lconst$.
}
\begin{corollary}[Retention for i.i.d.\ data]\label[corollary]{cor:out_reg_dist_acc}
    \OutlierDistributionalAccuracyCorollary
\end{corollary}

\begin{proof}
    Denote by $\qmargin\coloneqq\qscale\ln(2\nslots/\fprob)$ and $\cmargin\coloneqq\cscale\ln(2/\fprob)$ the margins of \Cref{thm:out_reg_acc}, so that the hypothesis on $\qscale$ reads $\qmargin\le(\nblk-\athr)/2-2$, and $\fprob\ge\fprobMin$ gives $\cmargin\le\cscale\Lconst$.
    Condition on the partition, which is independent of $\ds$; the blocks $\ds^{(1)}, \ldots, \ds^{(\nblk)}$ are then i.i.d.\ draws from $\dist^{\bsz}$.
    Call block $j$ \emph{good} if $\bbE_{X \sim \dist}\left[\pred(\ds^{(j)}, X)\right] \ge 1 - \ufrac$, and let $B$ be the event that all $\nblk$ blocks are good; the hypothesis on $\dist$ and $\pred$ and a union bound give $\Pr[\neg B] < \nblk\bfail$.

    For $\cand \in \dom$ let $\bar{\cnt}(\cand) \coloneqq \vert\{j : \pred(\ds^{(j)}, \cand) = 0\}\vert$ count the blocks that $\cand$ is unfriendly with.
    Call record $i$ \emph{at risk} if $\cnt_{i} \le \athr + \qmargin$; since $\cnt_{i} = \nblk - \bar{\cnt}(\ds_{i})$, this reads $\bar{\cnt}(\ds_{i}) \ge \nblk-\athr - \qmargin \ge (\nblk-\athr)/2 + 2$.
    Split this count by whether a block precedes or follows the one containing the record: for a record in block $j$ and $\cand\in\dom$, define
    \[
        \bar{\cnt}_{<}(\cand)\coloneqq\bigl\vert\{h<j : \pred(\ds^{(h)},\cand)=0\}\bigr\vert
        \qquad\text{and}\qquad
        \bar{\cnt}_{>}(\cand)\coloneqq\bigl\vert\{h>j : \pred(\ds^{(h)},\cand)=0\}\bigr\vert,
    \]
    where the dependence on $j$ is suppressed.
    The block of the record itself contributes at most $1$ to $\bar{\cnt}(\ds_i)$, so an at-risk record satisfies $\bar{\cnt}_{<}(\ds_{i})+\bar{\cnt}_{>}(\ds_{i})\ge (\nblk-\athr)/2+1$, and hence $\bar{\cnt}_{<}(\ds_{i}) \ge (\nblk-\athr)/4$ or $\bar{\cnt}_{>}(\ds_{i}) \ge (\nblk-\athr)/4$.

    If $\cscale\Lconst>8\nslots$, the at-risk records number at most $\nslots<\cscale\Lconst/8$, which is below $\svcap-\cmargin$ since $\cmargin\le\cscale\Lconst$.
    In this case, \Cref{thm:out_reg_acc} shows, on its event of probability at least $1-\fprob$, that the run does not abort and rejects at most $\nslots<\cscale\Lconst/2$ records, as claimed.
    Assume therefore that $\cscale\Lconst\le8\nslots$, so that $\cscale\Lconst/(8\nslots)$ is a valid Bernoulli parameter below.

    Read the blocks in index order and let $\mathcal{G}_{j} \coloneqq \sigma\left(\ds^{(1)}, \ldots, \ds^{(j)}\right)$.
    For $i$ in block $j$ the indicator $\ind\{\bar{\cnt}_{<}(\ds_{i}) \ge (\nblk-\athr)/4\}$ is a function of $\ds_{i}$ and $\mathcal{G}_{j-1}$ alone, and $\ds_{i}$ is an independent draw from $\dist$; if the first $j-1$ blocks are good then $\bbE_{X}[\bar{\cnt}_{<}(X)] \le \nblk \ufrac$, so Markov's inequality bounds its conditional probability by $\nblk \ufrac/((\nblk-\athr)/4) = 4\nblk \ufrac/(\nblk-\athr) \le \cscale\Lconst/(8\nslots)$ (here $\nblk-\athr>0$, since $(\nblk-\athr)/2-2\ge0$).
    Zeroing the indicator whenever some earlier block is bad makes this bound hold unconditionally.
    Ordering the records by block and then within block, each indicator then has conditional probability at most $\cscale\Lconst/(8\nslots)$ given all its predecessors, so by \Cref{lem:domination} the resulting count $z_{<}$ over the $\nslots$ records satisfies $\Pr[z_{<}\ge y]\le\Pr[\Bin(\nslots, \cscale\Lconst/(8\nslots))\ge y]$ for every $y$, where the binomial has mean $\cscale\Lconst/8$.
    Reading the blocks in the reverse order gives the same for $z_{>}$.

    The upper-tail bound in \Cref{prop:multiplicative_chernoff}, applied with $u=1$, gives $\Pr\bigl[\Bin(\nslots, \cscale\Lconst/(8\nslots)) \ge \cscale\Lconst/4\bigr] \le e^{-\cscale\Lconst/24}$, so $z_{<}+z_{>}<\cscale\Lconst/2$ except with probability $2e^{-\cscale\Lconst/24}$.
    On $B$, the indicators were never zeroed, so their sum bounds the number of at-risk records.
    Including $\Pr[\neg B]<\nblk\bfail$, that number is below $\cscale\Lconst/2$ except with probability $\nblk\bfail+2e^{-\cscale\Lconst/24}$.
    Since $\cmargin \le \cscale\Lconst$ this is below $\svcap - \cmargin$.
    The retention guarantee of \Cref{thm:out_reg_acc}, on its event of probability at least $1-\fprob$, therefore shows that the run does not abort and retains every record that is not at risk, so fewer than $\cscale\Lconst/2$ records are rejected.
\end{proof}

%% file: sections/apd_pfr.tex
\section{\texorpdfstring{Proofs from \Cref{sec:PFR}}{Proofs from Propose-Filter-Release}}\label{app:generalized-friendly-core}

This appendix contains the proofs that are omitted or only outlined in \Cref{sec:PFR}, with the long results restated before their proofs, and states the retention guarantee of the filter with the parameters of PFR.
It then supports the comparisons of \Cref{sec:ptr_comparison}: it outlines why the zCDP filter of FriendlyCore meets the hypotheses of the PFR theorem, states propose-test-release as an instantiation of PFR, and relates friendly DP to down sensitivity.

\begin{theorem}[Restatement of \Cref{lem:friendly_comp}]
    \leavevmode\phantomsection\label[theorem]{lem:friendly_comp_restated}
    \FriendlyCompositionLemma
\end{theorem}

\begin{proof}[Proof of \Cref{lem:friendly_comp}]
    Fix $\ds$ and $i$ with $\cand\coloneqq\ds_i\ne\nul$, and write $\ds'\coloneqq\ds^{i\to\nul}$ and $\eps\coloneqq\eps_1+\eps_2$.
    Run the filter once on $\ds$, denote its output by $\fds\coloneqq\filt(\ds)$, and let $A\coloneqq\fds^{i\to\nul}$.
    Thus $\fds=A$ if the filter rejected $\cand$, and $\fds=A^{i\to\cand}$ if it retained $\cand$.
    The two releases to compare are $Y\coloneqq\mech(\fds)$ and $Y'\coloneqq\mech(\filt(\ds'))$.

    For the analysis, modify this same filter output by additionally nulling slot $i$ whenever $\cand$ is not friendly with $A$:
    \[
        \widetilde A\coloneqq
        \begin{cases}
            \fds, & \pred(A,\cand)=1,\\
            A, & \pred(A,\cand)=0,
        \end{cases}
        \qquad
        \widetilde Y\coloneqq\mech(\widetilde A).
    \]
    If the filter already rejected $\cand$, then $\fds=A$, so this modification has no effect.
    If it retained $\cand$, then $\widetilde A$ keeps $\cand$ exactly when $\pred(A,\cand)=1$.
    We first compare $\widetilde Y$ with $Y'$ using friendly DP, and then account for the probability that $\widetilde A\ne\fds$.

    Conditional on $A=y$, if $\pred(y,\cand)=0$, then $\widetilde A=y$ and $\widetilde Y$ has law $\mech(y)$.
    If $\pred(y,\cand)=1$, then $\widetilde A$ is either $y$ or $y^{i\to\cand}$, according to whether the filter rejected or retained $\cand$.
    Friendly DP gives $\mech(y^{i\to\cand})\approx_{\eps_2,\delta_2}\mech(y)$, so any mixture of these two release laws is also $(\eps_2,\delta_2)$-indistinguishable from $\mech(y)$.
    In both cases, $(\widetilde Y\mid A=y)\approx_{\eps_2,\delta_2}\mech(y)$.

    To apply \Cref{prop:two_stage}, set $A'\coloneqq\filt(\ds')$, so that $Y'=\mech(A')$.
    Its first-stage variables are $A$ and $A'$: they take values in a common finite set $U$, and joint differential privacy gives $A\approx_{\eps_1,\delta_1}A'$.
    For each $y\in U$, let $P_y$ be the law of $(\widetilde Y\mid A=y)$ and $P'_y$ the law of $\mech(y)$; when $\Pr[A=y]=0$, set $P_y=P'_y$.
    The preceding comparison gives $P_y\approx_{\eps_2,\delta_2}P'_y$ for every $y\in U$.
    These are the conditional laws of the second-stage variables $\widetilde Y$ and $Y'$, respectively.
    Thus \Cref{prop:two_stage}, with its two privacy-parameter pairs interchanged, gives $(A,\widetilde Y)\approx_{\eps_1+\eps_2,\delta_1+\delta_2}(A',Y')$.
    Discarding the first component by post-processing yields $\widetilde Y\approx_{\eps,\delta_1+\delta_2}Y'$.

    Finally, couple $Y$ and $\widetilde Y$ using the same filter output $\fds$ and the same randomness for $\mech$.
    They agree unless $\fds_i=\cand$ and $\pred(A,\cand)=0$, an event whose probability is at most $\delta_3$ by assumption.
    Hence, for every event $O$,
    \begin{align*}
        \Pr[Y\in O]
        &\le\Pr[\widetilde Y\in O]+\delta_3
        \le e^{\eps}\Pr[Y'\in O]+\delta_1+\delta_2+\delta_3,\\
        \Pr[Y'\in O]
        &\le e^{\eps}\Pr[\widetilde Y\in O]+\delta_1+\delta_2
        \le e^{\eps}\Pr[Y\in O]+\delta_1+\delta_2+e^{\eps}\delta_3.
    \end{align*}
    Since $\eps\ge0$, these two inequalities prove the claimed privacy guarantee.
\end{proof}

\begin{theorem}[Restatement of \Cref{thm:gen_frnd_core_priv}]
    \leavevmode\phantomsection\label[theorem]{thm:gen_frnd_core_priv_restated}
    \GeneralizedFriendlyCoreTheorem
\end{theorem}

\begin{proof}
    Given $\ds$ and $i$, write $\cand\coloneqq\ds_i$ and $\ds'\coloneqq\ds^{i\to\nul}$.
    If $\cand=\nul$, the datasets coincide, so assume otherwise, and abbreviate the filter by $\filt$.
    We verify the hypotheses of \Cref{lem:friendly_comp} with $\eps_1=\eFilt$, $\eps_2=\eRel$, $\delta_1=\dFilt$, $\delta_2=\dRel$ and $\delta_3=\dFriend$, for which the PFR theorem gives the additive error $(25+e^\eps)\delta/32<\delta$; the friendly DP of $\mech$ is assumed.

    \emph{Step 1: joint differential privacy of the filter.}
    Set $J\coloneqq\ln(4/\delta)$ and $\bcap\coloneqq\lceil\cBound\Lconst/\eps\rceil$.
    Since $\eps\le1$ and $\Lconst\ge\ln(\Lbase)$, we have $12J\le\bcap\le37\Lconst/\eps$.
    With $\qscale=\qscaleVal$ and $J\le\Lconst$, this gives $\eps_{\SV}\coloneqq7\sqrt{\bcap J}/\qscale\le(7\sqrt{37}/\cTest)\eps<5\eps/12\le4J$.
    Applying \Cref{prop:sparse_vec_priv} with privacy parameters $(\eps_{\SV},\delta/2)$ therefore makes the sparse vector with fixed queries and deterministic cap $\bcap$ $(\eps_{\SV},\delta/2)$-DP.
    The cap is $\svcap=2\cscale\Lconst=\capOffset$, so $\Pr[\ncap>\bcap]\le\Pr[\cnoise>\cscale\Lconst]=\frac12e^{-\Lconst}=\dCap$.
    The two comparisons in the proof of \Cref{thm:out_reg_jdp} now apply with these scales: changing the input at fixed queries costs $(\eps_{\SV},\delta/2+\dCap)$, and changing the exceptional query costs $(1/\cscale,0)$.
    Their privacy costs sum to at most $5\eps/12+\eps/\cCap=\eFilt$, and their additive errors sum to $\delta/2+\dCap=\dFilt$.
    Therefore $\filt(\ds)^{i\to\nul}\approx_{\eFilt,\dFilt}\filt(\ds')$, and since $\ds$ and $i$ are arbitrary, the filter is $(\eFilt,\dFilt)$-JDP.

    \emph{Step 2: a retained record is friendly with the remaining data.}
    Denote $\Lconst_{32}\coloneqq\ln(32/\delta)\le5\Lconst/4$, and pre-sample the test noises.
    The event $E_i\coloneqq\{\qnoise_i\le\qscale\Lconst_{32},\ \cnoise\le\cscale\Lconst_{32}\}$ has probability at least $1-\dFriend$ by the two one-sided Laplace tails.
    On $E_i$, if $\cand$ is retained, then $\cnt_i>\athrVal-\qscale\Lconst_{32}$, and since the run did not abort, fewer than $\svcap+\cscale\Lconst_{32}=\capOffset+\cscale\Lconst_{32}$ records are rejected.
    The chosen noise scales give
    \[
        \svcap+(\qscale+\cscale)\Lconst_{32}+1
        \le\frac{135\Lconst^2}{\eps^{3/2}}+\frac{39\Lconst}{\eps}+1
        \le\frac{150\Lconst^2}{\eps^{3/2}}
        \le\frac{3\nblk}{8},
    \]
    where the second inequality uses $39\Lconst/\eps+1\le40\Lconst/\eps\le15\Lconst^2/\eps^{3/2}$, since $\Lconst\ge\ln(\Lbase)>8/3$ and $\eps\le1$, and the last uses $\nblk=\nslots/\bsz\ge\mPriv$.
    Discarding the blocks of the rejected records and the block containing slot $i$ therefore leaves $\nint$ intact, $2\thr$-friendly blocks, where $\nint>\athrVal-\capOffset-(\qscale+\cscale)\Lconst_{32}-1\ge\nblk/2$.
    By permutation invariance, group the intact blocks first in $A\coloneqq\filt(\ds)^{i\to\nul}$.
    Concavity, $\nslots=\nblk\bsz$ and $\fscore\ge0$ on the remaining slots imply $\fscore(A,\cand)\ge(\nint\bsz/\nslots)\,2\thr=(\nint/\nblk)\,2\thr\ge\thr$.
    Thus the probability that $\cand$ is retained although it is not $\thr$-friendly with $A$ is at most $\dFriend$, which is the last hypothesis of \Cref{lem:friendly_comp}; only the noise of the changed slot is controlled, so no union bound over records is needed.
\end{proof}

The following corollary specializes \Cref{thm:out_reg_acc} to the parameters of \Cref{thm:gen_frnd_core_priv}, and \Cref{cor:out_reg_dist_acc} gives the corresponding condition for i.i.d.\ data.

\begin{corollary}[Retention with the parameters of PFR]\label[corollary]{cor:pfr_retention}
    Given $\nslots,\bsz\in\bbN$ with $\nslots/\bsz\in\bbN$; $\eps\in(0,1]$; $\delta\in(0,1)$; $\fprob\in[\fprobMin,1)$; a friendliness threshold $\thr>0$ and a friendliness score $\fscore$, if $\nblk\ge\mRet{\fprob}$, then the filter $\filt_{\fpar}(\pred_{2\thr}(\fscore))$ with the parameters $\athr$, $\svcap$, $\cscale$ and $\qscale$ of \Cref{thm:gen_frnd_core_priv}, which is the filter of $\PFRd(\mech,\fscore,\thr)$ for every mechanism $\mech$, satisfies the following, where $\nblk\coloneqq\nslots/\bsz$ and $\Lconst\coloneqq\Ldef$.
    For every dataset $\ds\in\dom^\nslots$, with probability at least $1-\fprob$, also conditionally on the partition of the filter, both conclusions of \Cref{thm:out_reg_acc} hold with $\retMargin$ in place of the margin $\qmargin$ and $\capSlack$ in place of the margin $\cmargin$.
    In particular, on this event, if fewer than $\capSlack$ records have $\cnt_i\le\athrRet$, where $\cnt_i$ is the number of blocks that $\ds_i$ is friendly with under $\pred_{2\thr}(\fscore)$, then the filter does not abort and retains every record with $\cnt_i>\athrRet$.
\end{corollary}

\begin{proof}
    Since $16\cdot\cTest\le\cRet$, the lower bound on $\nblk$ gives $\qscale\ln(2\nslots/\fprob)=\cTest\Lconst\ln(2\nslots/\fprob)/\eps^{3/2}\le\retMargin$, and $\fprob\ge\fprobMin$ gives $\cscale\ln(2/\fprob)=\cCap\ln(2/\fprob)/\eps\le\capSlack$.
    The margins of \Cref{thm:out_reg_acc} are therefore at most $\retMargin$ and $\capSlack$, respectively, so on its event of probability at least $1-\fprob$ its conclusions also hold with these larger margins, and substituting $\athr=\athrVal$ and $\svcap=\capOffset$ gives the stated retention condition.
\end{proof}

Half of the cap $\svcap=\capOffset$ absorbs its noise, and the other half leaves room for the fewer than $\capSlack$ failing records of the corollary.
For a smaller $\fprob$, the condition $\fprob\ge\fprobMin$ can be met by running PFR with $\delta$ replaced by $\cFprob\fprob$, which only strengthens its privacy but enlarges $\Lconst$ in the lower bounds on $\nblk$.

\phantomsection\paragraph{The zCDP filter of FriendlyCore as an instantiation of PFR.}\label{app:fc_gaussian}
FriendlyCore~\cite{FriendlyCore22} works with sequences of records without null records, under add/remove adjacency, and with a reflexive predicate $\ppred$ on pairs of records.
A sequence is \emph{$\ppred$-friendly} if every two of its elements $y,y'$ have a common friend, a data value $z$, not necessarily in the sequence, with $\ppred(y,z)=\ppred(y',z)=1$~\cite[Definition~1.1]{FriendlyCore22}.
An algorithm is \emph{$\ppred$-friendly $(\eps,\delta)$-DP} if it is $(\eps,\delta)$-indistinguishable on neighboring sequences whose union is $\ppred$-friendly~\cite[Definition~3.2]{FriendlyCore22}.
The zCDP filter of FriendlyCore~\cite[Algorithm~4.5]{FriendlyCore22} receives parameters $\rho,\delta>0$ and returns a subsequence of its input, the \emph{core}.
We outline why this filter, translated to zero-out adjacency, meets the hypotheses of \Cref{lem:friendly_comp}, citing the guarantees of \citet{FriendlyCore22} rather than repeating their analysis.
On an input $\ds\in\dom^\nslots$, the translated filter deletes the null records, runs the zCDP filter, with its parameters $\rho$ and $\delta$, on the sequence of the remaining records in slot order, and restores the records of the core to their slots, nulling the others.
On an input without non-null records, the zCDP filter runs on the empty sequence and returns the empty core, so the output is $\nul^\nslots$.
For neighboring datasets $\ds$ and $\ds^{i\to\nul}$, the two sequences of non-null records differ by the single element $\ds_i$, so they are neighbors under add/remove adjacency.
The records outside slot $i$ correspond one to one to the elements that the two sequences share, and the decisions on these elements are what joint differential privacy compares.

First, the condition of the release mechanism is the friendliness predicate $\pred_\ppred$, in the sense of \Cref{def:set_friendliness}, where, for a non-null $\cand$, $\pred_\ppred(\ds',\cand)\coloneqq1$ if every two non-null records of $(\ds',\cand)$ have a common friend, and $\pred_\ppred(\ds',\cand)\coloneqq0$ otherwise.
For a reflexive $\ppred$, a single non-null record is a friend of itself, so $\pred_\ppred(\nul^\nslots,\cand)=1$ for every $\cand\ne\nul$.
Since the non-null records of the union of $\ds$ and $\ds^{i\to\nul}$ are those of $\ds$, a $\ppred$-friendly $(\eps_2,\delta_2)$-DP algorithm, applied to the non-null records, is $(\eps_2,\delta_2)$-friendly-DP with respect to $\pred_\ppred$.
Second, \citet[Lemma~4.6]{FriendlyCore22} show that, conditionally on events of probability at least $1-\delta$ in the two executions, the decisions of the filter on the shared elements satisfy $\rho$-zCDP.
For every $\delta_0\in(0,1)$, the conversion of $\rho$-zCDP to $\bigl(\rho+2\sqrt{\rho\ln(1/\delta_0)},\delta_0\bigr)$-DP~\cite{bun2016concentrated} and the accounting of \Cref{fact:failure_event_accounting} for the events make the translated filter $\bigl(\rho+2\sqrt{\rho\ln(1/\delta_0)},\delta+\delta_0\bigr)$-JDP.
Third, on the events of the same lemma, the union of the cores of the two executions is $\ppred$-friendly.
A subset of a $\ppred$-friendly dataset is $\ppred$-friendly, so on these events every two retained records have a common friend, and a retained changed record $\cand$ satisfies $\pred_\ppred(\filt(\ds)^{i\to\nul},\cand)=1$.
The last hypothesis of \Cref{lem:friendly_comp} therefore holds with $\delta_3=\delta$, and the theorem gives an approximate-DP counterpart of the composition theorem of FriendlyCore~\cite[Theorem~4.8]{FriendlyCore22}, with different constants.

We conclude with propose-test-release, which \Cref{sec:ptr_comparison} compares with PFR.
We state it using the same notion of a friendly pair and the zero-out adjacency of this work.

\begin{definition}[Propose-test-release, PTR~\cite{dwork2009differential}]\label[definition]{def:PTR}
    Given $\nslots\in\bbN$; $\eps_0>0$; $\delta_0\in(0,1)$; a friendliness threshold $\thr > 0$, a friendliness score $\fscore$ and a mechanism $\mech$ on $\dom^\nslots$, let $\mathcal B_\thr$ consist of both members of all pairs $(v,v^{i\to\nul})$ with $v\in\dom^\nslots$, $i\in[\nslots]$, $v_i\ne\nul$ and $\pred_\thr(\fscore;v^{i\to\nul},v_i)=0$.
    Denote by $\mathrm{dist}_\thr(\ds)$ the minimum number of zero-out adjacency steps from $\ds$ to $\mathcal B_\thr$, where a step may either null a record or restore a value at a null slot, with $\mathrm{dist}_\thr(\ds)\coloneqq\infty$ if $\mathcal B_\thr$ is empty.

    The associated \emph{propose-test-release} mechanism returns $\fail$ if $\mathrm{dist}_{\thr}(\ds) + \Lap(1/\eps_{0}) \le \ln(1/\delta_{0})/\eps_{0}$, and $\mech(\ds)$ otherwise, using independent randomness for $\mech$.
\end{definition}

When $\mathcal B_\thr$ is nonempty, this distance has sensitivity at most $1$: a path from a neighboring dataset can be prefixed with one adjacency step.
If $\mech$ is $(\eps_2,\delta_2)$-friendly-DP with respect to $\pred_\thr(\fscore)$, PTR is $(\eps_0+\eps_2,\delta_0+\delta_2)$-DP.
Indeed, on a friendly neighboring pair we compose the $(\eps_0,0)$-DP test with $\mech$; on an unfriendly pair both distances are zero, so each execution returns $\fail$ with probability at least $1-\delta_0/2$.
If $\mathcal B_\thr$ is empty, every neighboring pair is friendly and the guarantee follows directly from that of $\mech$.

\phantomsection\paragraph{PTR as an instantiation of PFR.}\label{app:ptr_pfr}
The test of PTR defines the filter that returns $\ds$ if $\mathrm{dist}_\thr(\ds)+\Lap(1/\eps_0)>\ln(1/\delta_0)/\eps_0$ and $\nul^\nslots$ otherwise, which retains either all the records of $\ds$ or none.
Since the distance has sensitivity at most $1$, and the test never rejects if $\mathcal B_\thr$ is empty, this filter is $(\eps_0,0)$-JDP.
If $\pred_\thr(\fscore;\ds^{i\to\nul},\ds_i)=0$ for a non-null $\ds_i$, then $\ds\in\mathcal B_\thr$ and $\mathrm{dist}_\thr(\ds)=0$, so the filter retains $\ds_i$ with probability at most $\delta_0/2$.
\Cref{lem:friendly_comp} therefore applies with $\delta_3=\delta_0/2$ to the variant of PTR that releases $\mech(\nul^\nslots)$ instead of $\fail$ when the test rejects.
It gives $(\eps_0+\eps_2,\delta_2+e^{\eps_0+\eps_2}\delta_0/2)$-DP, which has the form of the guarantee above up to the coefficient of $\delta_0$.

The next proposition shows that this test fails on every input when every dataset with a null slot admits an unfriendly record, which is the case for the trace and leverage scores on an unbounded domain: for every dataset $w$ and every threshold, a record of sufficiently large norm is unfriendly with $w$ under both $\tracescore$ (\Cref{thm:empirical_second_moment_PFR}) and $\fscore_\ridge$ (\Cref{def:leverage}).

\begin{proposition}[PTR with unfriendly values]\label[proposition]{prop:ptr_fails}
    In the setting of \Cref{def:PTR}, if for every $w\in\dom^\nslots$ that has a null slot there exists a non-null record $y\in\dom$ with $\pred_\thr(\fscore;w,y)=0$, then $\mathrm{dist}_\thr(\ds)\le1$ for every $\ds\in\dom^\nslots$, and on every input the PTR mechanism returns $\fail$ except with probability at most $e^{\eps_0}\delta_0/2$.
\end{proposition}

\begin{proof}
    Given $\ds$ and a slot $j$, denote $w\coloneqq \ds^{j\to\nul}$, which is either $\ds$ itself or one nulling step away from it.
    Take $y$ with $\pred_\thr(\fscore;w,y)=0$.
    The neighboring pair $(w^{j\to y},w)$ is then not admitted by $\pred_\thr(\fscore)$, so $w\in\mathcal B_\thr$ and $\mathrm{dist}_\thr(\ds)\le1$.
    PTR avoids $\fail$ only if $\Lap(1/\eps_0)>\ln(1/\delta_0)/\eps_0-1$.
    For every real $a$ and $\sigma>0$ we have $\Pr[\Lap(\sigma)>a]\le\frac12e^{-a/\sigma}$: this is an equality for $a\ge0$ (\Cref{fact:laplace}), and for $a<0$, writing $c\coloneqq e^{-a/\sigma}>1$, it reads $1-1/(2c)\le c/2$, that is, $(c-1)^2\ge0$.
    With $\sigma=1/\eps_0$ and $a=\ln(1/\delta_0)/\eps_0-1$, the probability is at most $e^{\eps_0}\delta_0/2$.
\end{proof}

\phantomsection\paragraph{Down sensitivity.}\label{app:down_sensitivity}
The per-instance sensitivity bound of \Cref{thm:PFR_DP}, which suffices for friendly DP, is a bound on the down sensitivity~\cite{chen2013recursive,linder2025blackbox}, the change of a statistic when records that are present are removed, restricted to the removals that the predicate admits.
A test of such a bound in the style of PTR would use a deletion-only diagnostic, the number of nulling steps until a remaining record violates the proposed bound.
In least-squares regression, for example, records with small residuals around a hyperplane give a fit of low down sensitivity when the explanatory variables are well spread and no record dominates the fit; a single outlier with an extreme response reduces the diagnostic to zero, and removing it recovers the stable fit.
This diagnostic need not have sensitivity $1$: inserting an outlier into a null slot can change it from a large value to zero in one neighboring step, so adding $\Lap(1/\eps_0)$ to it does not in general give a private test.
A filtering step that retains all records or none cannot reject the inserted outlier alone, so its decision must be nearly the same whatever value the changed slot holds.
PTR uses the full, symmetric adjacency relation precisely to avoid this problem.
Its set $\mathcal B_\thr$ (\Cref{def:PTR}) contains both members of each violating pair, so a dataset can lie in $\mathcal B_\thr$ even if the violating value is not present in it, and restricting paths to nulling steps while keeping $\mathcal B_\thr$ would not recover the deletion-only diagnostic.
Element--block filtering takes a different route to a bound on the down sensitivity: it does not test the bound on the whole input, but filters the records individually so that the bound holds for the changed record of each neighboring pair with high probability.

\Cref{thm:gen_frnd_core_priv} takes a concave score and a friendly-DP release mechanism as its hypotheses, and it is not stated as a reduction from a bound on down sensitivity.
For release mechanisms that do not add noise to a statistic, such as the Gaussian sketches of \Cref{thm:filt_sketch_DP} and the exponential mechanism of \Cref{thm:weighted_selection}, friendly DP follows from the privacy analysis of the mechanism itself.
The leverage of a record, for example, bounds a R\'enyi divergence between the two sketch distributions, and it is neither the sensitivity of a statistic nor the privacy loss itself.

%% file: sections/apd_applications.tex
\section{\texorpdfstring{Details and Proofs from \Cref{sec:applications}}{Details and Proofs from Applications}}\label{app:friendly-mechanisms}

This appendix contains the formal definitions, proofs and utility guarantees behind \Cref{sec:applications}, with the long results restated before their proofs.

\subsection{Scalar Releases and Trace Estimation}\label{app:trace_proofs}

This subsection formalizes the scalar releases of \Cref{sec:trace} and proves the guarantees of trace estimation.
The influence of a record on a scalar release is measured by the following notion.

\begin{definition}[Per-instance sensitivity~\cite{wang2019perinstance}]\label[definition]{def:per_instance_sens}
    Given a real-valued function $\stat$ on nonempty datasets, its \emph{per-instance sensitivity} at a nonempty dataset $\ds$ and a slot $i\in[|\ds|]$ with $\ds_i\ne\nul$ is $\vert\stat(\ds)-\stat(\ds^{i\to\nul})\vert$.
    A function $\psens:\dom^*\times\dom\to[0,\infty]$ \emph{bounds the per-instance sensitivity} of $\stat$ if $\vert \stat(\ds) - \stat(\ds^{i \to \nul}) \vert \le \psens(\ds^{i \to \nul}, \ds_{i})$ for every $\nslots\ge1$, $\ds \in \dom^{\nslots}$ and $i \in [\nslots]$ with $\ds_i\ne\nul$.
\end{definition}

The bound depends only on the remaining data and the changed record, as a friendliness score does, and its scaled reciprocal is the score of the release.

\begin{theorem}[Scalar PFR]\label{thm:PFR_DP}
    Given $\nslots,\bsz\in\bbN$ with $\nslots/\bsz\in\bbN$; $\eps \in (0,1]$; $\delta \in (0,1)$; $\sigma > 0$ with $\gaussprof(\eRel) \le \dRel$; a friendliness threshold $\thr > 0$, a real-valued function $\stat$ on nonempty datasets, and a function $\psens$ that bounds its per-instance sensitivity, if the score $\fscore(\ds,\cand)\coloneqq1/(|\ds|\,\psens(\ds,\cand))$, defined for nonempty $\ds$ and $\cand\ne\nul$ with $1/0\coloneqq\infty$ and $1/\infty\coloneqq0$, is a concave friendliness score and $\nslots\ge\cBlk\bsz\ln^2(\Lbase/\delta)/\eps^{3/2}$, then $\PFRd(\Gauss_{\sigma/(\thr\nslots)}(\stat),\fscore,\thr)$ is $(\eps, \delta)$-DP on datasets of $\nslots$ records.
\end{theorem}

\begin{proof}
    On an admitted pair, $\fscore(\ds^{i\to\nul},\ds_i)\ge\thr$ and $|\ds^{i\to\nul}|=\nslots$ give $\psens(\ds^{i\to\nul},\ds_i)\le1/(\thr\nslots)$, so the two releases are Gaussians of common standard deviation $\sigma/(\thr\nslots)$ whose means differ by at most $1/(\thr\nslots)$.
    Multiplying both by $\thr\nslots$, the Gaussian mechanism (\Cref{fact:gauss_profile}) gives pDP with parameters $(\eRel,\gaussprof(\eRel))$ at every admitted pair, so the release mechanism is $(\eRel,\gaussprof(\eRel))$-friendly-DP, and hence $(\eRel,\dRel)$-friendly-DP.
    With the concavity of the score, \Cref{thm:gen_frnd_core_priv} proves the claim.
\end{proof}

For the trace, the score of \Cref{thm:empirical_second_moment_PFR} is, on nonempty datasets $\ds$, $\tracescore(\ds,\cand)=\smom(\ds)/\lVert\cand\rVert^2$ for $\cand\ne0$ and $\tracescore(\ds,0)=\infty$.
The released statistic $\tracestat=\ln(\smom)$ takes the value $\ln(0)\coloneqq-\infty$ on a dataset whose records are all null or zero, where the mechanism of \Cref{thm:empirical_second_moment_PFR} therefore releases $-\infty$, and we set $e^{-\infty}\coloneqq0$.
Since $\tracestat$ is not real-valued, \Cref{thm:PFR_DP} does not apply to it verbatim, and the proof below repeats its argument.

\begin{lemma}\label[lemma]{lem:trace_score}
    The trace score $\tracescore$ is a concave friendliness score.
    Moreover, given $\nslots\in\bbN$; $i\in[\nslots]$; $\thr>0$; and a dataset $\ds\in\dom^\nslots$ with $\ds_i\ne\nul$, if $\ds_i$ is $\thr$-friendly with $\ds^{i\to\nul}$, then either $\ds_i=0$ and $\tracestat(\ds)=\tracestat(\ds^{i\to\nul})$, or $\smom(\ds^{i\to\nul})>0$ and $|\tracestat(\ds)-\tracestat(\ds^{i\to\nul})|\le1/(\thr\nslots)$.
\end{lemma}

\begin{proof}
    The trace $\smom$ is a size-weighted average under concatenation, so the score is concave with equality for $\cand\ne0$, and on nonempty datasets it is $\infty$ for $\cand=0$.
    For the second claim, nulling slot $i$ changes nothing when $\ds_i=0$.
    For $\ds_i\ne0$, friendliness gives $\smom(\ds^{i\to\nul})\ge\thr\lVert\ds_i\rVert^2>0$, and $\smom(\ds)=\smom(\ds^{i\to\nul})+\lVert\ds_i\rVert^2/\nslots$.
    Hence $|\tracestat(\ds)-\tracestat(\ds^{i\to\nul})|=\ln\bigl(1+\lVert\ds_i\rVert^2/(\nslots\,\smom(\ds^{i\to\nul}))\bigr)\le\lVert\ds_i\rVert^2/(\nslots\,\smom(\ds^{i\to\nul}))\le1/(\thr\nslots)$, by $\ln(1+y)\le y$.
\end{proof}

\begin{corollary}[Restatement of \Cref{thm:empirical_second_moment_PFR}]
    \leavevmode\phantomsection\label[corollary]{thm:empirical_second_moment_PFR_restated}
    \EmpiricalSecondMomentPFRTheorem
\end{corollary}

\begin{proof}[Proof of \Cref{thm:empirical_second_moment_PFR}]
    We follow the proof of \Cref{thm:PFR_DP}.
    On an admitted pair, the changed record is $\thr$-friendly with the remaining data, so by \Cref{lem:trace_score}, the two releases either have the same distribution or are Gaussians of common standard deviation $\sigma$ whose finite means differ by at most $1/(\thr\nslots)$.
    Multiplying both by $\thr\nslots$ gives Gaussians of standard deviation $\sigma_0\coloneqq\thr\nslots\sigma=4\sqrt\Lconst/\eps$ whose means differ by at most $1$.
    The Gaussian mechanism (\Cref{fact:gauss_profile}) therefore gives pDP with parameters $(\eRel,\delta_{\Gauss_{\sigma_0}}(\eRel))$ at every admitted pair, and by the same fact and the standard Gaussian tail bound,
    \[
        \delta_{\Gauss_{\sigma_0}}(\eRel)
        \le\Phi\left(\frac{\eps}{8\sqrt \Lconst}-2\sqrt \Lconst\right)
        \le\Phi\left(-\sqrt{2\Lconst}\right)
        \le e^{-\Lconst}=\frac\delta{\Lbase}<\frac\delta4,
    \]
    where $\Lconst>1$ and $\eps\le1$ justify the second inequality.
    The release mechanism is therefore $(\eRel,\dRel)$-friendly-DP, and with the concavity of $\tracescore$ (\Cref{lem:trace_score}), \Cref{thm:gen_frnd_core_priv} proves the claim.
\end{proof}

The next corollary gives a condition under which the filter retains all the records, and bounds the error with respect to the input, with no bound on the norms of the records.

\begin{corollary}[Utility of private trace estimation]\label[corollary]{cor:trace_utility}
Given $\nslots,\bsz\in\bbN$ with $\nslots/\bsz\in\bbN$; $\eps\in(0,1]$; $\delta\in(0,1)$; $\fprob\in(0,1)$ with $\delta\le\cFprobQ\fprob$; a friendliness threshold $\thr\in(0,1/\cThr]$, and an input $\ds\in\dom^\nslots$ without null records such that $\smom(\ds)>0$ and $\lVert\cand\rVert^2\le\smom(\ds)/(\cThr\thr)$ for each of its records $\cand$, if $\bsz\ge(\cBsz/\thr)\ln(4\nslots/\fprob)$ and $\nslots/\bsz\ge\max\{\mPriv,\ \mRetQ\}$, then with probability at least $1-\fprob$, the filter of the mechanism of \Cref{thm:empirical_second_moment_PFR} retains every record of $\ds$, and its output $Y$ satisfies $e^{-\xi}\smom(\ds)\le e^Y\le e^\xi\smom(\ds)$, where $\Lconst\coloneqq\Ldef$ and $\xi\coloneqq4\sqrt{2\Lconst\ln(4/\fprob)}/(\thr\eps\nslots)$.
\end{corollary}

\begin{proof}[Proof of \Cref{cor:trace_utility}]
    Denote $v\coloneqq\smom(\ds)>0$ and $\fscore\coloneqq\tracescore$.

    \emph{Representative blocks.}
    The values $U_i\coloneqq2\thr\lVert\ds_i\rVert^2/v$ lie in $[0,2/\cThr]\subseteq[0,1]$ by the assumption on the norms, and their average over the $\nslots$ records is $2\thr$.
    A reference block is a uniform sample of $\bsz$ slots without replacement, and by \citet[Theorem~4]{hoeffding1963probability}, the expectation of a continuous convex function of the sum of its values $U_i$ is at most that for $\bsz$ independent uniform draws from them, with replacement.
    For such a draw $U$, the inequality $e^{-y\ln(2)}\le1-y/2$ for $y\in[0,1]$ gives $\bbE e^{-U\ln(2)}\le1-\thr\le e^{-\thr}$, so Markov's inequality gives, for every block $\ds^{(j)}$ of $\ds$,
    \[
        \Pr\Bigl[\sum_{i\in\ds^{(j)}}U_i<\thr\bsz\Bigr]
        \le\exp\bigl(\thr\bsz\ln(2)-\thr\bsz\bigr)
        \le\exp\Bigl(-\frac{\thr\bsz}{\cBsz}\Bigr).
    \]
    Since $\smom(\ds^{(j)})=(v/(2\thr\bsz))\sum_{i\in\ds^{(j)}}U_i$, a union bound over the at most $\nslots$ blocks and the lower bound on $\bsz$ show that, except with probability $\fprob/4$, every block of $\ds$ satisfies $\smom(\ds^{(j)})\ge v/2$.
    On this event, every record $\cand$ of $\ds$ satisfies $\fscore(\ds^{(j)},\cand)\ge(v/2)\cdot\cThr\thr/v=2\thr$ against every block, also when $\cand=0$, whose score is $\infty$.

    \emph{Filtering.}
    Condition on a partition in this event.
    Every record is then friendly with all $\nblk$ blocks under $\pred_{2\thr}(\fscore)$, so no record has $\cnt_i\le\athrRet$.
    \Cref{cor:pfr_retention}, with failure probability $\fprob/4\ge\fprobMin$, then shows that the filter retains every record, except with probability $\fprob/4$.

    \emph{Accuracy.}
    On full retention, the filtered dataset is $\ds$, so the output is $Y=\tracestat(\ds)+Z$, where the noise $Z$ is independent of the filter and Gaussian with standard deviation $4\sqrt \Lconst/(\thr\eps \nslots)$.
    Hence
    \[
        \Pr[|Z|>\xi]
        \le2\exp\left(-\frac{\xi^2}
                       {2\left(4\sqrt \Lconst/(\thr\eps \nslots)\right)^2}\right)
        =\frac\fprob2,
    \]
    and on the complementary event, exponentiating $|Y-\tracestat(\ds)|\le\xi$ gives the claimed inequalities.
    The failure probabilities of the three steps sum to $\fprob$.
\end{proof}

For independent records, the next corollary replaces the condition on the input by a bound on the squared norms relative to their unknown mean, as \Cref{cor:weighted_population} does for the weights of treatment-rule selection.

\begin{corollary}[Trace estimation on i.i.d.\ data]\label[corollary]{cor:trace_population}
Given $\dimn,\nslots,\bsz\in\bbN$ with $\nslots/\bsz\in\bbN$; $\eps\in(0,1]$; $\delta\in(0,1)$; $\fprob\in(0,1)$ with $\delta\le\cFprobQ\fprob$; a friendliness threshold $\thr\in(0,1/\cThr]$, and a distribution over $\bbR^\dimn$ whose draws $X$ satisfy $\lVert X\rVert^2\le\mu/(\cThr\thr)$ almost surely for $\mu\coloneqq\bbE\lVert X\rVert^2\in(0,\infty)$, if $\bsz\ge(\cBszW/\thr)\ln(4\nblk/\fprob)$ and $\nblk\ge\max\{\mPriv,\ \mRetQ\}$, then with probability at least $1-\fprob$ over a dataset $\ds$ of $\nslots$ independent draws from this distribution and the randomness of the mechanism of \Cref{thm:empirical_second_moment_PFR}, its filter retains every record of $\ds$, and its output $Y$ satisfies $e^{-\xi}\smom(\ds)\le e^Y\le e^\xi\smom(\ds)$, where $\nblk\coloneqq\nslots/\bsz$, $\Lconst\coloneqq\Ldef$ and $\xi\coloneqq4\sqrt{2\Lconst\ln(4/\fprob)}/(\thr\eps\nslots)$.
\end{corollary}

\begin{proof}
    The squared norms $\lVert\ds_i\rVert^2$ are independent and satisfy the conditions that the proof of \Cref{cor:weighted_population} places on the weights, so its representative-blocks step shows that, except with probability $\fprob/4$, every block of $\ds$ satisfies $\smom(\ds^{(j)})\ge\mu/2$.
    On this event, $\smom(\ds)>0$, and every record $\cand\ne0$ of $\ds$ satisfies $\tracescore(\ds^{(j)},\cand)\ge(\mu/2)\cdot\cThr\thr/\mu=2\thr$ against every block, while a zero record has score $\infty$.
    The filtering and accuracy steps of the proof of \Cref{cor:trace_utility} then apply without change, with failure probabilities $\fprob/4$ and $\fprob/2$.
\end{proof}

\subsection{Gaussian Sketches and Subspace Estimation}\label{app:gaussian-subspace}

This subsection proves the results of \Cref{sec:sketching}.
The following definition formalizes the score of \Cref{sec:sketching}, which \Cref{sec:graph} also uses with zero ridge.

\begin{definition}[Regularized leverage]\label[definition]{def:leverage}
    Given $\dimn\in\bbN$; $\ridge \ge 0$; and a nonempty dataset $\ds$ of vectors in $\bbR^{\dimn}$, with null records represented by zero vectors, denote its regularized second moment by $\regmom_{\ridge}(\ds) \coloneqq \ds^{T}\ds/\vert \ds \vert + \ridge I_{\dimn}$.
    The \emph{$\ridge$-regularized leverage} of $\cand \in \bbR^{\dimn}$ with respect to $\ds$ is $\lev_{\ridge}(\ds, \cand) \coloneqq \cand^{T} \regmom_{\ridge}(\ds)^{-1} \cand$ for $\ridge>0$, and its limit as $\ridge\to0^+$ for $\ridge=0$.
    The corresponding friendliness score is $\fscore_{\ridge}(\ds,\cand)\coloneqq1/\lev_{\ridge}(\ds,\cand)$.
\end{definition}

Recall that nulling a slot changes neither the denominator $\nslots$ of the regularized second moment nor its public ridge.
For this score, we use the conventions $1/0\coloneqq\infty$ and $1/\infty\coloneqq0$, and the leverage of a vector outside the span of the reference data is $\infty$ for $\ridge=0$.

\begin{lemma}\label[lemma]{lem:lev_concave}
    Given $\ridge \ge 0$, the score $\fscore_{\ridge}$ of \Cref{def:leverage} is a concave friendliness score.
\end{lemma}

\begin{proof}
    The score is invariant under permutations, and its value on the empty dataset is the one that \Cref{def:friendliness} prescribes.
    For $\cand\ne0$ and $A\succ0$, the Cauchy--Schwarz inequality in the inner product defined by $A$ gives $(\cand^TA^{-1}\cand)^{-1}=\inf_{z:\cand^Tz=1}z^TAz$, with equality at $z=A^{-1}\cand/(\cand^TA^{-1}\cand)$.
    An infimum of linear functions of $A$ is concave and nonnegative on the positive semidefinite cone.
    Since $\regmom_\ridge((\ds,\ds'))$ is the size-weighted average of $\regmom_\ridge(\ds)$ and $\regmom_\ridge(\ds')$, this proves the concavity condition of \Cref{def:friendliness} for $\ridge>0$ and $\cand\ne0$.
    For $\ridge=0$, apply this inequality with ridges $\ridge'$ that decrease to zero: for each of the three datasets, $\lev_{\ridge'}(\cdot,\cand)$ increases to $\lev_0(\cdot,\cand)\in(0,\infty]$, so the reciprocal leverages converge to finite nonnegative limits, which are zero when $\cand$ has a component outside the span of the reference data, and the inequality is preserved in the limit.
    Finally, on nonempty datasets the score is $\infty$ for an ordinary zero record, so the inequality also holds in this case.
\end{proof}

The privacy of the sketches relies on the following bound for a rank-one change of the covariance of a centered Gaussian.
The same divergences appear in the analysis of the Gaussian mixing mechanism of \citet[Lemma~1]{LevSSLSW25}; we include a short proof of the form that we use, in which the leverage is measured against the remaining data.

\def\SketchRDPProposition{
    Given $\rord>1$; a positive semidefinite matrix $A$ and a vector $y$ in its range, if $(\rord-1)\slev<1$, then both directed R\'enyi divergences of order $\rord$ between $\Gauss(0,A)$ and $\Gauss(0,A+yy^T)$ are at most $\rdpbd_\rord(\slev)$, where $\slev\coloneqq y^TA^\dagger y$ and $\rdpbd_\rord(\slev)\coloneqq\rord\slev^2/\bigl(4(1-(\rord-1)\slev)\bigr)$.
}
\begin{proposition}[R\'enyi divergence of a single sketch]\label[proposition]{prop:sketch_RDP}
    \SketchRDPProposition
\end{proposition}

\begin{proof}
    If $y=0$, the two distributions coincide, so assume $y\ne0$.
    Both distributions are supported on the range of $A$, which contains $y$, so we may restrict all matrices and vectors to this subspace, where $A$ is positive definite and $\slev=y^TA^{-1}y$.

    A common invertible change of variables preserves R\'enyi divergence.
    Applying $A^{-1/2}$ transforms the pair into $\Gauss(0,I)$ and $\Gauss(0,I+uu^T)$, where $u\coloneqq A^{-1/2}y$ satisfies $\lVert u\rVert^2=\slev$, and $I$ is the identity on the range of $A$.
    Rotate the coordinates so that $u$ lies along the first axis.
    The first coordinate then has variance $1$ in one distribution and $1+\slev$ in the other, while all the other coordinates are independent standard Gaussians in both and contribute zero divergence.
    It therefore remains to compare $\Gauss(0,1)$ and $\Gauss(0,1+\slev)$ in both directions.

    For positive variances $v,v'$, let $p$ and $p'$ be the densities of $\Gauss(0,v)$ and $\Gauss(0,v')$, and denote $\xi\coloneqq\rord/v+(1-\rord)/v'$.
    When $\xi>0$, substituting the Gaussian densities gives
    \[
        \exp\bigl((\rord-1)\Renyi{\rord}(\Gauss(0,v)\Vert\Gauss(0,v'))\bigr)
        =\int_{\bbR}p(z)^\rord p'(z)^{1-\rord}\,dz
        =v^{-\rord/2}(v')^{(\rord-1)/2}\xi^{-1/2},
    \]
    and when $\xi\le0$ the integral diverges.
    Taking logarithms, dividing by $\rord-1$, and substituting the two pairs of variances yields
    \begin{align*}
        d_1(\slev)
        &\coloneqq\Renyi{\rord}\bigl(\Gauss(0,1)\Vert \Gauss(0,1+\slev)\bigr)
        =\frac{\rord\ln(1+\slev)-\ln(1+\rord\slev)}{2(\rord-1)},\\
        d_2(\slev)
        &\coloneqq\Renyi{\rord}\bigl(\Gauss(0,1+\slev)\Vert \Gauss(0,1)\bigr)
        =-\frac{\ln(1+\slev)}2
         -\frac{\ln(1-(\rord-1)\slev)}{2(\rord-1)},
    \end{align*}
    where the second divergence is finite exactly under the hypothesis $(\rord-1)\slev<1$, since then $\xi=(1-(\rord-1)\slev)/(1+\slev)$.
    Both functions vanish at zero, and their derivatives are
    \[
        d_1'(\theta)=\frac{\rord \theta}{2(1+\theta)(1+\rord \theta)}
        \qquad\text{and}\qquad
        d_2'(\theta)=\frac{\rord \theta}{2(1+\theta)(1-(\rord-1)\theta)}.
    \]
    For $0\le \theta\le\slev$, we have $1+\theta\ge1$, $1+\rord \theta\ge1$ and $1-(\rord-1)\theta\ge1-(\rord-1)\slev>0$, so for either $j\in\{1,2\}$,
    \[
        d_j(\slev)=\int_0^\slev d_j'(\theta)\,d\theta
        \le\frac{\rord}{2(1-(\rord-1)\slev)}\int_0^\slev \theta\,d\theta
        =\rdpbd_{\rord}(\slev).
    \]
    Moreover, $\rdpbd_{\rord}$ is increasing on $[0,1/(\rord-1))$, since its numerator increases and its positive denominator decreases, so the bound also holds when $\slev$ is replaced by an upper bound in this interval.
\end{proof}

\begin{corollary}[Restatement of \Cref{thm:filt_sketch_DP}]
    \leavevmode\phantomsection\label[corollary]{thm:filt_sketch_DP_restated}
    \FilteredSketchTheorem
\end{corollary}

\begin{proof}[Proof of \Cref{thm:filt_sketch_DP}]
    The first lower bound on $\nslots$ is the one that \Cref{thm:gen_frnd_core_priv} requires, and $\fscore_\ridge$ is a concave friendliness score by \Cref{lem:lev_concave}.
    It remains to show that the second lower bound on $\nslots$ makes the release mechanism $\mech_{\ridge,\nsk}$, which draws the $\nsk$ sketches, $(\eRel,\dRel)$-DP on friendly pairs.
    Given a pair $\ds,\ds^{i\to\nul}$ admitted by $\pred_\thr(\fscore_\ridge)$, the two sketch covariances are $A\coloneqq\regmom_\ridge(\ds^{i\to\nul})$ and $A+yy^T$ for $y\coloneqq\ds_i/\sqrt\nslots$, since both datasets have $\nslots$ slots and the same ridge.
    Friendliness gives $\lev_\ridge(\ds^{i\to\nul},\ds_i)\le1/\thr<\infty$, so $y$ lies in the range of $A$, also when $\ridge=0$, and $y^TA^\dagger y\le1/(\thr\nslots)$; denote $\slev\coloneqq1/(\thr\nslots)$ and $\Lconst_{4}\coloneqq\ln(4/\delta)=\Lconst-\ln(4)$.
    \Cref{prop:sketch_RDP}, the monotonicity of $\rdpbd_\rord$, the additivity of R\'enyi divergence over the independent sketches, and \Cref{prop:rdp_to_dp} with the order $\rord=1+v$ give $(\eps_{\mathrm{sk}},\dRel)$-friendly DP, where
    \[
        \eps_{\mathrm{sk}}
        \le\frac{\nsk(1+v)\slev^2}{4(1-v\slev)}+\frac{\Lconst_{4}}{v}
        \qquad(0<v<1/\slev).
    \]
    The choice $v\coloneqq\sqrt{\Lconst_{4}}/\bigl(\slev\bigl(\sqrt{\nsk(1+\slev)}/2+\sqrt{\Lconst_{4}}\bigr)\bigr)$ lies in this interval, and substituting it yields $\eps_{\mathrm{sk}}\le \nsk\slev^2/4+\slev\sqrt{\nsk(1+\slev)\Lconst_{4}}+\slev \Lconst_{4}$, for every $\nsk\ge1$.

    To check the constant, denote $K\coloneqq\Kconst$.
    The second lower bound on $\nslots$ gives $\slev\le\eps/(2K)\le1/(2\Lconst)$, and consequently
    \[
        \sqrt{\frac{(1+\slev)\Lconst_{4}}{\Lconst}}
        \le\sqrt{1-\frac{\ln(4)-1/2}{\Lconst}}
        \le1-\frac{\ln(4)-1/2}{2\Lconst}
        \le1-\frac{1}{8\Lconst}.
    \]
    Using also $\nsk\slev/4\le\nsk\eps/(8K)\le\sqrt{\nsk\Lconst}/(8\Lconst)$, which follows from $K\ge\sqrt{\nsk\Lconst}$ and $\eps\le1$, and $\Lconst_{4}\le \Lconst$, we get
    \[
        \eps_{\mathrm{sk}}
        \le\slev\left[
            \frac{\sqrt{\nsk\Lconst}}{8\Lconst}
            +\sqrt{\nsk\Lconst}\left(1-\frac{1}{8\Lconst}\right)+\Lconst
        \right]
        =\slev K\le\frac\eps2.
    \]
    The sketches are therefore $(\eRel,\dRel)$-friendly-DP, as \Cref{thm:gen_frnd_core_priv} requires, which proves the claim.
\end{proof}

We first collect the tools of the utility analysis: concentration bounds for Gaussian vectors and a perturbation bound for projectors.
For $Y\sim\Gauss(0,B)$ and $z>0$, diagonalization and the Gaussian moment-generating function give
\begin{equation}\label{eq:subspace_quadratic}
 \Pr\left[\lVert Y\rVert^2>\Tr(B)+2\sqrt{\Tr(B^2)z}+2\lVert B\rVert z\right]\le e^{-z}.
\end{equation}
Indeed, for $0<y<(2\lVert B\rVert)^{-1}$, the logarithm of $\bbE e^{y(\lVert Y\rVert^2-\Tr(B))}$ is $\sum_j\bigl[-y\lambda_j(B)-\tfrac12\ln(1-2y\lambda_j(B))\bigr]\le y^2\Tr(B^2)/(1-2y\lVert B\rVert)$, and optimizing the Chernoff bound over $y$ gives \eqref{eq:subspace_quadratic}.
For $q$ independent copies $Y_1,\ldots,Y_q$, the effective-rank bound of \citet[Corollary~2]{koltchinskii2017concentration} gives a universal constant $C_0$ such that, for every $z\ge1$, with probability at least $1-e^{-z}$,
\begin{equation}\label{eq:subspace_effective_concentration}
 \left\lVert\frac1q\sum_{j=1}^qY_jY_j^T-B\right\rVert
 \le C_0\left[\sqrt{\frac{\lVert B\rVert\Tr(B)+\lVert B\rVert^2z}{q}}
                  +\frac{\Tr(B)+\lVert B\rVert z}{q}\right].
\end{equation}
Their effective rank $(\bbE\lVert Y\rVert)^2/\lVert B\rVert$ is at most $\Tr(B)/\lVert B\rVert$, which gives this form, and for $B=0$ the bound is trivial.

The next lemma is a perturbation bound for the projector onto the positive eigenspace of a symmetric matrix, in which the perturbation is measured relative to the magnitudes of the eigenvalues.
It is what allows the subspace application to use a small absolute gap.

\begin{lemma}[A gap-weighted projector bound]\label[lemma]{lem:gap_weighted_projector}
Given $\eta\in[0,1)$; an invertible real symmetric matrix $A$ and a real symmetric matrix $F$ with $\lVert F\rVert\le\eta$, the matrices $A$ and $B$ have the same numbers of positive and negative eigenvalues, and $\lVert P_+(B)-P_+(A)\rVert\le\eta/(2(1-\eta))$, where $H\coloneqq|A|$, $E\coloneqq H^{1/2}FH^{1/2}$, $B\coloneqq A+E$, and $P_+(\cdot)$ denotes the orthogonal projector onto the positive eigenspace.
\end{lemma}

\begin{proof}
Denote $J\coloneqq\operatorname{sign}(A)$, so that $A=H^{1/2}JH^{1/2}$ and $B=H^{1/2}(J+F)H^{1/2}$.
The eigenvalues of $J+F$ are within $\eta<1$ of the eigenvalues $\pm1$ of $J$, so they cannot cross zero, and since $H^{1/2}$ is invertible, Sylvester's law of inertia proves the first assertion.

For $\omega\in\bbR$, denote $R_A(\omega)\coloneqq(A-i\omega I)^{-1}$, and similarly $R_B(\omega)$.
Since $A$ commutes with $H$, the matrix $H^{1/2}R_A(\omega)H^{1/2}$ is normal, with eigenvalues $|a|/(a-i\omega)$ for the eigenvalues $a$ of $A$, whose absolute values are at most one.
The resolvent identity $R_B=R_A-R_AER_B$ implies
\[
 \bigl(I+H^{1/2}R_A(\omega)H^{1/2}F\bigr)H^{1/2}R_B(\omega)=H^{1/2}R_A(\omega),
\]
so a Neumann series gives $\lVert H^{1/2}R_B(\omega)x\rVert\le\lVert H^{1/2}R_A(\omega)x\rVert/(1-\eta)$ for every complex vector $x$.
Diagonalizing $A$ and integrating each coordinate yields the exact identity
\[
 \int_{\bbR}\lVert H^{1/2}R_A(\omega)x\rVert^2\,d\omega
 =\sum_j|x_j|^2\int_{\bbR}\frac{|\lambda_j(A)|}{\lambda_j(A)^2+\omega^2}\,d\omega
 =\pi\lVert x\rVert^2,
\]
where $x_j$ are the coordinates of $x$ in an eigenbasis of $A$.
For a nonzero real scalar $a$, the symmetric integral of $(a-i\omega)^{-1}$ over $\bbR$ is $\pi\operatorname{sign}(a)$, and $P_+=(I+\operatorname{sign}(\cdot))/2$.
Functional calculus, followed by the resolvent identity $R_B-R_A=-R_BER_A$, thus gives
\[
 P_+(B)-P_+(A)
 =\frac1{2\pi}\int_{\bbR}(R_B(\omega)-R_A(\omega))\,d\omega
 =-\frac1{2\pi}\int_{\bbR}R_B(\omega)ER_A(\omega)\,d\omega,
\]
where the integral converges absolutely, since its integrand is continuous and of order $|\omega|^{-2}$ at infinity.
For unit vectors $x,y$, using $R_B(\omega)^*=R_B(-\omega)$, $E=H^{1/2}FH^{1/2}$, the Cauchy--Schwarz inequality in $\omega$, and the two preceding bounds, we get
\begin{align*}
 |x^*(P_+(B)-P_+(A))y|
 &\le\frac\eta{2\pi}
 \left(\int_{\bbR}\lVert H^{1/2}R_B(-\omega)x\rVert^2\,d\omega\right)^{1/2}
 \left(\int_{\bbR}\lVert H^{1/2}R_A(\omega)y\rVert^2\,d\omega\right)^{1/2}\\
 &\le\frac\eta{2(1-\eta)}.
\end{align*}
Taking the supremum over $x$ and $y$ proves the asserted bound on the operator norm.
\end{proof}

The utility analysis starts from the observation that, against blocks of order $\effdim(\Sigma)$ records, every record passes every leverage test.

\begin{lemma}[Simultaneous leverage tests]\label[lemma]{lem:population_retention}
There exists a universal constant $C$ such that the following holds.
Given $\nslots,\bsz\in\bbN$ with $\nslots/\bsz\in\bbN$; $\ridge>0$; $\zeta\in(0,1)$; $D>0$; a friendliness threshold $\thr>0$ and a positive semidefinite matrix $\Sigma$ with $\effdim(\Sigma)\le D$, if $\bsz\ge C(D+\ln(4\nslots/\zeta))$ and $1/\thr\ge C(D+\ln(4\nslots/\zeta))$, then for independent $X_1,\ldots,X_\nslots\sim\Gauss(0,\Sigma)$ and every partition of $[\nslots]$ into blocks of $\bsz$ records that is independent of them, with probability at least $1-\zeta$ every record $X_i$ passes the tightened test $\fscore_\ridge(\ds^{(j)},X_i)\ge2\thr$ against every block $\ds^{(j)}$ of $\ds\coloneqq(X_1,\ldots,X_\nslots)$, where $\effdim(\Sigma)\coloneqq\Tr(\Sigma(\Sigma+\ridge I)^{-1})$.
\end{lemma}

\begin{proof}
Condition on the partition, and denote $B\coloneqq\Sigma+\ridge I$ and $G\coloneqq B^{-1/2}\Sigma B^{-1/2}$, so that $\lVert G\rVert\le1$ and $\Tr(G)=\effdim(\Sigma)\le D$.
Since $\ridge B^{-1}=I-G$, the regularized second moment of a block $\ds^{(j)}$ satisfies
\[
 B^{-1/2}\regmom_\ridge(\ds^{(j)})B^{-1/2}
 =I+\frac1\bsz\sum_{i\in\ds^{(j)}}B^{-1/2}X_iX_i^TB^{-1/2}-G.
\]
By \eqref{eq:subspace_effective_concentration} with $z\coloneqq\ln(4\nslots/\zeta)$, the lower bound on $\bsz$ with $C$ large compared with $C_0$, and a union bound over the at most $\nslots$ blocks, all these matrices are at least $I/2$, except with probability $\zeta/4$.
By \eqref{eq:subspace_quadratic} and another union bound, $X_i^TB^{-1}X_i\le D+2\sqrt{Dz}+2z\le3(D+z)$ for all the records simultaneously, except with probability $\zeta/4$.
On both events, $\lev_\ridge(\ds^{(j)},X_i)\le2X_i^TB^{-1}X_i\le6(D+z)\le1/(2\thr)$ for every record and every block, by the inverse-matrix order and the choice of $C$.
These events concern all the records and blocks simultaneously, so no independence between a record and its own block is needed, and averaging over the partition proves the claim.
\end{proof}

We now state and prove the formal guarantee for the geometric spectrum, which \Cref{thm:subspace_informal} summarizes.
The proof measures the sampling error and the sketching error relative to $|\Sigma-(3/4)\lambda_\rank(\Sigma)I|$, against which $\Sigma+\ridge I$ has operator norm at most $5$ and trace $O(\rank)$.
\Cref{lem:gap_weighted_projector} then turns both errors into projector error $\acc$, without requiring $\nslots\ge \dimn$ or $\nsk\ge \dimn$.
The effective dimension $\effdim(\Sigma)$ is at most $2\rank$, which is why, with blocks of $\wtO(\rank)$ records, the filter retains all the records (\Cref{lem:population_retention,cor:pfr_retention}).

\begin{theorem}[Subspace estimation on the geometric spectrum]\label{thm:subspace_running}
There exists a universal constant $C$ such that the following holds.
Given $\dimn,\rank,\nslots,\bsz\in\bbN$ with $\nslots/\bsz\in\bbN$, $\dimn\ge8$ and $\lceil\log_2(\dimn)\rceil<\rank\le\dimn/2$; $\acc\in(0,1/4]$; $\fprob\in(0,1/4)$; $\eps\in(0,1]$; $\delta\in(0,\cFprobQ\fprob]$; and a covariance $\Sigma$ with the geometric spectrum, if $\bsz\ge C(\rank+\ln(\nslots/\fprob))$, $\nslots/\bsz\ge\max\{\mPriv,\ \mRetQ\}$, $\nslots\ge C(\rank+\ln(1/\fprob))/\acc^2$, and the records are independent draws from $\Gauss(0,\Sigma)$, then with probability at least $1-\fprob$, the subspace mechanism with parameters $\ridge$, $\nsk$ and $\thr$ retains every record in its filter and returns $\widetilde P$ with $\lVert\widetilde P-\projk(\Sigma)\rVert\le\acc$, where $\Lconst\coloneqq\Ldef$, $\ridge\coloneqq2^{-\lceil\log_2(\dimn)\rceil}/\dimn$, $\nsk\coloneqq\lceil C(\rank+\ln(1/\fprob))/\acc^2\rceil$ and $\thr\coloneqq1/(C(\rank+\ln(\nslots/\fprob)))$.
\end{theorem}

\begin{proof}[Proof of \Cref{thm:subspace_running}]
Write $\ell\coloneqq\lceil\log_2(\dimn)\rceil<\rank$ and $\bdry\coloneqq2^{-\ell}=\lambda_\rank(\Sigma)$.
Since $\ridge=\bdry/\dimn$ and the eigenvalues below rank $\rank$ are $\bdry2^{-j}$, their contribution to the effective dimension gives
\[
 \effdim(\Sigma)
 \le\rank+\sum_{j\ge1}\min\{1,\dimn2^{-j}\}
 \le\rank+\ell+\dimn2^{-\ell}\le2\rank.
\]
The choices of $\bsz$, $\thr$ and $\nblk$ therefore allow \Cref{lem:population_retention,cor:pfr_retention} with failure probability $\fprob/4$ each; here $\delta\le\cFprobQ\fprob$ ensures $\fprob/4\ge\fprobMin$.
Thus the filter retains every record except with probability $\fprob/2$.

Set $B\coloneqq\Sigma+\ridge I$ and $H\coloneqq|\Sigma-3\bdry I/4|$.
Every leading eigenvalue $x\ge\bdry$ satisfies $(x+\ridge)/(x-3\bdry/4)\le5$, and every tail eigenvalue $x\le\bdry/2$ satisfies $(x+\ridge)/(3\bdry/4-x)\le3$.
The tail denominators are at least $\bdry/4$, so
\[
 B\preceq5H,
 \qquad
 \Tr(H^{-1}B)
 \le5\rank+4\sum_{j\ge1}2^{-j}+\frac{4(\dimn-\rank)}\dimn
 \le5\rank+8.
\]
These are the only spectral bounds needed below.

Let $\empmom\coloneqq\nslots^{-1}\sum_iX_iX_i^T$.
The covariance of $B^{-1/2}X_i$ has norm at most one and trace $\effdim(\Sigma)\le2\rank$.
Thus \eqref{eq:subspace_effective_concentration} and the assumed lower bound on $\nslots$ give $-(\acc/10)B\preceq\empmom-\Sigma\preceq(\acc/10)B$, except with probability $\fprob/4$.
Consequently, since $B\preceq5H$,
\[
 \left\lVert H^{-1/2}(\empmom-\Sigma)H^{-1/2}\right\rVert\le\frac\acc2,
 \qquad
 \empmom+\ridge I\preceq(1+\acc/10)B.
\]
On this event and full retention, conditional on the data and filter randomness, the vectors $H^{-1/2}Z_j$ are independent Gaussians whose covariance has norm at most $6$ and trace $O(\rank)$, by the spectral bounds above.
Writing $R\coloneqq\nsk^{-1}\sum_jZ_jZ_j^T$, another application of \eqref{eq:subspace_effective_concentration} with the stated number $\nsk$ of sketches gives $\lVert H^{-1/2}(R-\empmom-\ridge I)H^{-1/2}\rVert\le\acc/2$, except with conditional probability $\fprob/4$.
Combining the two error bounds yields $\lVert H^{-1/2}(R-\ridge I-\Sigma)H^{-1/2}\rVert\le\acc$.
By \Cref{lem:gap_weighted_projector} with $A\coloneqq\Sigma-3\bdry I/4$, so that $|A|=H$, $F\coloneqq H^{-1/2}(R-\ridge I-\Sigma)H^{-1/2}$ and $\eta\coloneqq\acc$, the matrices $A$ and $A+H^{1/2}FH^{1/2}=R-(\ridge+3\bdry/4)I$ both have exactly $\rank$ positive eigenvalues, and their positive-eigenspace projectors differ by at most $\acc/(2(1-\acc))\le\acc$.
These projectors are $\projk(\Sigma)$ and the mechanism's output $\widetilde P$, respectively.
The total failure probability is at most $\fprob/2+\fprob/4+\fprob/4=\fprob$.
\end{proof}

\paragraph{Other spectra.}
The proof uses the geometric spectrum only through two of its properties, $\lambda_{\rank+1}(\Sigma)\le\lambda_\rank(\Sigma)/2$ and $\effdim(\Sigma)\le2\rank$ at the ridge $\ridge=\lambda_\rank(\Sigma)/\dimn$, so it extends to other spectra.

\begin{corollary}[Other spectra]\label[corollary]{cor:subspace_general_spectra}
    For every $c\ge1$ there exists a constant $C_c$ such that the following holds.
    Given a covariance $\Sigma$ whose eigenvalue $\lambda_\rank(\Sigma)>0$ is public, with $\lambda_{\rank+1}(\Sigma)\le\lambda_\rank(\Sigma)/2$ and $\effdim(\Sigma)\le c\rank$, \Cref{thm:subspace_running} and \Cref{cor:subspace_private} hold for $\Sigma$ in place of a covariance with the geometric spectrum, with the condition $\lceil\log_2(\dimn)\rceil<\rank\le\dimn/2$ replaced by $\rank\le\dimn/2$, with $C_c$ in place of $C$, and with the ridge $\ridge\coloneqq\lambda_\rank(\Sigma)/\dimn$, at which $\effdim$ is also taken.
\end{corollary}

\begin{proof}
For the geometric spectrum, $\lambda_\rank(\Sigma)/\dimn$ is the ridge of \Cref{thm:subspace_running}.
Given $\Sigma$ as in the statement, denote $\bdry\coloneqq\lambda_\rank(\Sigma)$, and $B$ and $H$ as in the proof of \Cref{thm:subspace_running}.
The eigenvalue ratio and $\ridge\le\bdry/4$ give $B\preceq5H$ as there.
A tail eigenvalue $x\ge\ridge$ contributes at least $1/2$ to $\effdim(\Sigma)$, and a smaller one at least $x/(2\ridge)$, so the tail eigenvalues sum to at most $(1+2/\dimn)\bdry\,\effdim(\Sigma)$, and $\Tr(H^{-1}B)\le5\rank+4(1+2/\dimn)c\rank+4$.
The proof of \Cref{thm:subspace_running} uses the condition $\lceil\log_2(\dimn)\rceil<\rank$ only to bound $\effdim(\Sigma)$, which is now a hypothesis.
With $\effdim(\Sigma)\le c\rank$ and this bound in place of $\effdim(\Sigma)\le2\rank$ and $\Tr(H^{-1}B)\le5\rank+8$, the proof of \Cref{thm:subspace_running} goes through with constants that depend only on $c$, and so does that of \Cref{cor:subspace_private}, which adds only the sketch condition.
\end{proof}

On the geometric spectrum, $\dimn$ determines $\lambda_\rank(\Sigma)=2^{-\lceil\log_2(\dimn)\rceil}$, so only other spectra need this value as a public parameter.
Selecting it privately is one of the open directions of \Cref{sec:limitations}.

\Cref{thm:subspace_running} concerns utility only: its hypotheses include the block condition of \Cref{thm:filt_sketch_DP}, but not the sketch condition $\nslots\ge\sketchPriv$.
The next corollary gives a sample size at which the conditions of both results hold, so that the subspace mechanism is private on every input and accurate on Gaussian data.

\begin{corollary}[Sample size for privacy and utility]\label[corollary]{cor:subspace_private}
    Given $\dimn,\rank\in\bbN$ with $\dimn\ge8$ and $\lceil\log_2(\dimn)\rceil<\rank\le\dimn/2$; $\acc\in(0,1/4]$; $\fprob\in(0,1/4)$; $\eps\in(0,1]$; and $\delta\in(0,\cFprobQ\fprob]$, with the constant $C$ and the choices of $\ridge$, $\nsk$ and $\thr$ of \Cref{thm:subspace_running}, the conditions of \Cref{thm:subspace_running} and the condition $\nslots\ge\sketchPriv$ of \Cref{thm:filt_sketch_DP} hold for a block size $\bsz$ of order $\rank+\ln(\nslots/\fprob)$ and a sample size $\nslots=\wtO(\rank/\eps^{3/2}+\rank/\acc^2+\rank^{3/2}/(\acc\eps))$ that is a multiple of $\bsz$.
    With these parameters, the subspace mechanism is $(\eps,\delta)$-DP on every input and satisfies the utility guarantee of \Cref{thm:subspace_running}.
\end{corollary}

None of the terms of this sample size depends explicitly on $\dimn$, and for a fixed number $\rank-\lceil\log_2(\dimn)\rceil$ of unit eigenvalues, it grows polylogarithmically with $\dimn$.

\begin{proof}
With $\bsz$ of order $\rank+\ln(\nslots/\fprob)$, the filtering condition costs $\wtO(\rank/\eps^{3/2})$ records, the sampling condition $\wtO(\rank/\acc^2)$, and the sketch-privacy condition $\wtO(\rank^{3/2}/(\acc\eps)+\rank/\eps)$, whose last term is absorbed by the filtering term since $\eps\le1$.
Privacy then follows from \Cref{thm:filt_sketch_DP}, and utility from \Cref{thm:subspace_running}.
\end{proof}

\subsection{KL Divergence of User Histograms}\label{app:histogram_kl}

This subsection proves the guarantees of \Cref{sec:histogram_kl} and gives the calculation behind its comparison with direct noise.

\phantomsection\paragraph{Setting and target.}\label{app:kl_setting}
A record is either $\nul$ or a whole histogram $\hist\in\mathbb Z_{\ge0}^\ncat$ with $\sum_{j\in[\ncat]}\hist_j=\nobs$, for public integers $\nobs\ge1$ and $\ncat\ge2$.
Zero-out adjacency thus protects all the observations of a user, and \Cref{claim:zero_out_replacement} converts this into privacy with respect to replacing the histogram of one user.
Denote by $\avgH_j(\ds)$ the average count of category $j$ over the records of $\ds$, where null records count as zero histograms and are included in $|\ds|$.
Given a public pseudocount $\pcount>0$, which is added to every average count, the smoothed pooled frequencies and the released statistic are
\[
 \sfreq_j(\ds)\coloneqq\frac{\pcount+\avgH_j(\ds)}{\nobs+\ncat\pcount},
 \qquad
 \klstat(\ds)\coloneqq\KL(\unif\Vert\sfreq(\ds))
 =\ln\Bigl(\frac{\nobs+\ncat\pcount}{\ncat}\Bigr)-\frac1\ncat\sum_{j\in[\ncat]}\ln\bigl(\pcount+\avgH_j(\ds)\bigr),
\]
for a nonempty dataset $\ds$, where $\unif = (1/\ncat,\ldots,1/\ncat)$ is the uniform distribution on the categories and $\KL(p\Vert q)\coloneqq\sum_jp_j\ln(p_j/q_j)$.
Of the two orientations of the divergence between the smoothed pooled frequencies and the uniform reference, the target is the one with the uniform distribution as the first argument.
It is therefore large when some categories are covered much less than the average, and it is finite even when a category is absent, since every smoothed frequency is at least $\pcount/(\nobs+\ncat\pcount)$.
On a dataset without null records, the smoothed average counts $\pcount+\avgH_j(\ds)$ sum to $\nobs+\ncat\pcount$, so $\klstat(\ds)$ is the logarithm of the ratio between their arithmetic and geometric means, which vanishes when they are all equal.
For example, three users with $\nobs=6$ observations over $\ncat=4$ categories and the histograms $(3,2,1,0)$, $(4,1,1,0)$ and $(2,3,1,0)$ have the average counts $(3,2,1,0)$.
With $\pcount=1$, the smoothed frequencies are $(0.4,0.3,0.2,0.1)$ and $\klstat\approx0.12$, while without the pseudocount the absent fourth category would make the divergence infinite.
On datasets with null records, which the filter can produce, we use the same formula, with the denominators $\nslots$ and $\nobs+\ncat\pcount$ unchanged, so that $\sfreq(\ds)$ sums to less than one.
This fixed-denominator extension of the target is not a divergence between two distributions, and the interpretations above hold only without null records.

\paragraph{Score and release mechanism.}
The score $\fKL(\ds,\hist)=\bigl(\ncat^{-1}\sum_{j\in[\ncat]}\hist_j/(\pcount+\avgH_j(\ds))\bigr)^{-1}$ of \Cref{thm:histogram_kl} is defined for a nonempty dataset $\ds$ and a non-null histogram $\hist$, and the release mechanism $\mech_{\mathrm{KL}}$ adds Laplace noise of scale $2/(\thr\nslots\eps)$ to the target $\klstat(\ds)$ defined above.
The privacy of this release mechanism rests on two properties of its score.

\begin{lemma}[Concavity and sensitivity]\label[lemma]{lem:kl_score}
Given $\pcount>0$, the score $\fKL$ is a concave friendliness score.
Moreover, given $\nslots\in\bbN$; $i\in[\nslots]$; and a dataset $\ds\in\dom^\nslots$ with $\ds_i\ne\nul$, the per-instance sensitivity of $\klstat$ at $\ds$ and $i$ is at most $1/(\nslots\fKL(\ds^{i\to\nul},\ds_i))$.
\end{lemma}

\begin{proof}
The score is invariant under permutations, and its value on the empty dataset is $0$ by definition.
Given a non-null histogram $\hist$, denote $a_j\coloneqq\hist_j/\ncat$ and $S\coloneqq\{j:a_j>0\}$, which is nonempty since $\sum_j\hist_j=\nobs\ge1$, and for a nonempty dataset $\ds$ denote $z_j\coloneqq\pcount+\avgH_j(\ds)>0$.
By the Cauchy--Schwarz inequality, $\bigl(\sum_{j\in S}y_j\bigr)^2\le\bigl(\sum_{j\in S}a_j/z_j\bigr)\bigl(\sum_{j\in S}z_jy_j^2/a_j\bigr)$ for every vector $y$, with equality when $y_j$ is proportional to $a_j/z_j$, so
\[
 \fKL(\ds,\hist)=\Bigl(\sum_{j\in S}\frac{a_j}{z_j}\Bigr)^{-1}
 =\inf\Bigl\{\sum_{j\in S}\frac{z_jy_j^2}{a_j}:\ \sum_{j\in S}y_j=1\Bigr\}
\]
is an infimum of linear functions of $z$, and hence a concave function of the average histogram of $\ds$.
Under concatenation, the average histogram is the size-weighted average of the two average histograms, with the same public $\pcount$, which proves the concavity condition of \Cref{def:friendliness}.

For the sensitivity bound, given $\ds$ and $i$ with $\hist\coloneqq\ds_i\ne\nul$, denote $\ds'\coloneqq\ds^{i\to\nul}$.
The first term of $\klstat$ does not depend on the dataset, and since both datasets have $\nslots$ slots, $\avgH_j(\ds)=\avgH_j(\ds')+\hist_j/\nslots$, so
\[
 0\le\klstat(\ds')-\klstat(\ds)
 =\frac1\ncat\sum_{j\in[\ncat]}\ln\Bigl(1+\frac{\hist_j}{\nslots(\pcount+\avgH_j(\ds'))}\Bigr)
 \le\frac1{\nslots\ncat}\sum_{j\in[\ncat]}\frac{\hist_j}{\pcount+\avgH_j(\ds')}
 =\frac{1}{\nslots\,\fKL(\ds',\hist)},
\]
by $\ln(1+y)\le y$.
\end{proof}

\paragraph{Utility.}
For utility, we assume only a deterministic condition, under which no user contributes to a category much more than its regularized average count: given $\thr\le1/\cThr$, the input has no null records and its histograms $\hist_i\coloneqq\ds_i$ satisfy
\begin{equation}\label{eq:relative_histogram}
 \hist_{ij}\le\frac{\pcount+\avgH_j(\ds)}{\cThr\thr}
 \qquad\text{for all }i\in[\nslots]\text{ and }j\in[\ncat].
\end{equation}
The condition gives no lower bound on the frequencies other than the pseudocount.

\begin{theorem}[Utility of the private KL divergence]\label{thm:histogram_kl_utility}
Given $\nslots,\bsz\in\bbN$ with $\nslots/\bsz\in\bbN$; $\eps\in(0,1]$; $\delta\in(0,1)$; $\fprob\in(0,1)$ with $\delta\le\cFprobQ\fprob$; a public pseudocount $\pcount>0$, a friendliness threshold $\thr\in(0,1/\cThr]$, and an input $\ds\in\dom^\nslots$ without null records that satisfies \eqref{eq:relative_histogram}, if $\bsz\ge(\cBsz/\thr)\ln(4\nslots\ncat/\fprob)$ and $\nslots/\bsz\ge\max\{\mPriv,\ \mRetQ\}$, then with probability at least $1-\fprob$, the filter of $\PFRd(\mech_{\mathrm{KL}},\fKL,\thr)$ retains every record, and its output $Y$ satisfies $|Y-\klstat(\ds)|\le2\ln(2/\fprob)/(\thr\nslots\eps)$, where $\ncat$ is the number of categories and $\Lconst\coloneqq\Ldef$.
\end{theorem}

The proof shows that in random blocks of $\wtO(1/\thr)$ users, every regularized average count is at least half of that of the input, so every user passes every test; consequently, $\nslots=\wtO(\ln(\ncat)/(\thr\eps^{3/2})+1/(\thr\acc\eps))$ users suffice for error $\acc$.

\begin{proof}[Proof of \Cref{thm:histogram_kl} and \Cref{thm:histogram_kl_utility}]
\emph{Privacy.}
By \Cref{lem:kl_score}, a pair admitted by $\pred_\thr(\fKL)$ changes $\klstat$ by at most $1/(\thr\nslots)$, so the Laplace mechanism (\Cref{fact:laplace}) with scale $2/(\thr\nslots\eps)$ makes $\mech_{\mathrm{KL}}$ $(\eRel,0)$-friendly-DP with respect to $\pred_\thr(\fKL)$, and since the score is concave, \Cref{thm:gen_frnd_core_priv} proves privacy on every input.

\emph{Representative blocks.}
Given the input $\ds$, which has no null records and satisfies \eqref{eq:relative_histogram}, fix a category $j$ and denote $\mu\coloneqq\avgH_j(\ds)$.
If $\mu\le\pcount$, every block satisfies $\pcount+\avgH_j(\ds^{(\ell)})\ge\pcount\ge(\pcount+\mu)/2$.
Otherwise, every count satisfies $\hist_{ij}\le(\pcount+\mu)/(\cThr\thr)<\mu/(2\thr)$, so the normalized counts $U_i\coloneqq2\thr\hist_{ij}/\mu$ lie in $[0,1]$ and have average $2\thr$ over the $\nslots$ records.
The representative-blocks step of the proof of \Cref{cor:trace_utility} uses only that its values lie in $[0,1]$ and average $2\thr$, so the same comparison of sampling without and with replacement and the same Markov bound give $\Pr[\avgH_j(\ds^{(\ell)})<\mu/2]=\Pr[\sum_{i\in\ds^{(\ell)}}U_i<\thr\bsz]\le\exp(-\thr\bsz/\cBsz)$ for every block $\ell$.
A union bound over the at most $\nslots\ncat$ pairs of a block and a category, together with the lower bound on $\bsz$, shows that, except with probability $\fprob/4$,
\[
 \pcount+\avgH_j(\ds^{(\ell)})\ge\tfrac12\bigl(\pcount+\avgH_j(\ds)\bigr)
 \qquad\text{for every block $\ell$ and category $j$}.
\]
On this event, \eqref{eq:relative_histogram} gives, for every record and every block,
\[
 \frac{1}{\fKL(\ds^{(\ell)},\hist_i)}
 =\frac1\ncat\sum_{j\in[\ncat]}\frac{\hist_{ij}}{\pcount+\avgH_j(\ds^{(\ell)})}
 \le\frac2\ncat\sum_{j\in[\ncat]}\frac{\hist_{ij}}{\pcount+\avgH_j(\ds)}
 \le\frac{2}{\cThr\thr}
 =\frac1{2\thr},
\]
so every record passes every test, since none is null.
This event concerns all the records and blocks simultaneously, so no independence between a record and its own block is needed.

\emph{Retention and release.}
Conditional on such a partition, \Cref{cor:pfr_retention}, with failure probability $\fprob/4\ge\fprobMin$ and the assumed number of blocks, shows that the filter retains every record, except with probability $\fprob/4$.
On full retention, the output is $\klstat(\ds)$ plus independent Laplace noise of scale $2/(\thr\nslots\eps)$, whose magnitude exceeds $2\ln(2/\fprob)/(\thr\nslots\eps)$ with probability $\fprob/2$.
The three failure probabilities sum to $\fprob$.
\end{proof}

\paragraph{Multinomial users.}
The condition \eqref{eq:relative_histogram} is expected when all the users draw their observations from the same distribution.

\begin{claim}[Typicality of multinomial users]\label[claim]{claim:kl_multinomial}
    Given $\nslots,\nobs,\ncat\in\bbN$; $\fprob\in(0,1)$; a pseudocount $\pcount>0$, a friendliness threshold $\thr>0$, and a distribution $p$ on $[\ncat]$, if $\nslots\pcount\ge8\ln(2\ncat/\fprob)$ and $1/\thr\ge2\cThr(2+3\ln(2\nslots\ncat/\fprob)/\pcount)$, then $\thr\le1/\cThr$, and for independent histograms $\hist_1,\ldots,\hist_\nslots$, each of $\nobs$ independent observations drawn from $p$, the dataset $\ds\coloneqq(\hist_1,\ldots,\hist_\nslots)$ satisfies \eqref{eq:relative_histogram} with probability at least $1-\fprob$.
\end{claim}

\begin{proof}
The hypothesis on $\thr$ gives $1/\thr\ge4\cThr\ge\cThr$.
Fix a category $j$ and denote $\mu_j\coloneqq\nobs p_j$, so that each count $\hist_{ij}$ is binomial with mean $\mu_j$, and $\nslots\avgH_j(\ds)$ is binomial with mean $\nslots\mu_j$.
For a binomial variable $X$ with mean $\mu$ and $y\ge0$, the Bernstein inequality gives $\Pr[X\ge2\mu+y]\le\exp\bigl(-(\mu+y)^2/(2\mu+2(\mu+y)/3)\bigr)\le e^{-3y/8}$, and the Chernoff bound gives $\Pr[X\le\mu/2]\le e^{-\mu/8}$.
With $J\coloneqq3\ln(2\nslots\ncat/\fprob)$ and a union bound over the $\nslots\ncat$ counts, every count satisfies $\hist_{ij}\le2\mu_j+J$, except with probability $\fprob/2$.
If $\mu_j\le\pcount$, then $\pcount+\avgH_j(\ds)\ge\pcount\ge(\pcount+\mu_j)/2$.
Otherwise, $\nslots\mu_j>\nslots\pcount\ge8\ln(2\ncat/\fprob)$, so $\avgH_j(\ds)\ge\mu_j/2$ and again $\pcount+\avgH_j(\ds)\ge(\pcount+\mu_j)/2$, except with probability $\fprob/(2\ncat)$.
A union bound over the categories shows that both events hold together, except with probability $\fprob$, and on them
\[
 \hist_{ij}
 \le2\mu_j+J
 \le(\pcount+\mu_j)\Bigl(2+\frac{J}{\pcount}\Bigr)
 \le2\bigl(\pcount+\avgH_j(\ds)\bigr)\Bigl(2+\frac{J}{\pcount}\Bigr)
 \le\frac{\pcount+\avgH_j(\ds)}{\cThr\thr}
\]
for every user $i$ and category $j$, by the hypothesis on $\thr$, which is \eqref{eq:relative_histogram}.
\end{proof}

With \Cref{thm:histogram_kl_utility}, a pseudocount of order $\ln(\nslots\ncat/\fprob)$ thus allows a constant threshold on such data.

\phantomsection\paragraph{Direct noise.}\label{app:kl_direct_noise}
The next claim bounds the global sensitivity of $\klstat$ and gives a user who nearly attains it.

\begin{claim}[Global sensitivity of the divergence]\label[claim]{claim:kl_global_sensitivity}
    Given $\nslots\in\bbN$ and a pseudocount $\pcount>0$, nulling a record of a dataset in $\dom^\nslots$ increases $\klstat$ by at most $\nobs/(\nslots\ncat\pcount)$, where $\ncat$ is the number of categories and $\nobs$ the number of observations of each user.
    If the record places all its $\nobs$ observations on a category that no other record covers, the increase is $\ncat^{-1}\ln(1+\nobs/(\nslots\pcount))$, which is at least $\nobs/(2\nslots\ncat\pcount)$ when $\nslots\pcount\ge\nobs$.
\end{claim}

\begin{proof}
Nulling a null record changes nothing.
Given $\ds\in\dom^\nslots$ and a slot $i$ with $\hist\coloneqq\ds_i\ne\nul$, the identity in the proof of \Cref{lem:kl_score}, together with $\avgH_j(\ds^{i\to\nul})\ge0$ and $\ln(1+y)\le y$, gives
\[
 \klstat(\ds^{i\to\nul})-\klstat(\ds)
 =\frac1\ncat\sum_{j\in[\ncat]}\ln\Bigl(1+\frac{\hist_j}{\nslots(\pcount+\avgH_j(\ds^{i\to\nul}))}\Bigr)
 \le\frac{1}{\nslots\ncat\pcount}\sum_{j\in[\ncat]}\hist_j
 =\frac{\nobs}{\nslots\ncat\pcount}.
\]
If $\hist$ places all its observations on a category $j$ with $\avgH_j(\ds^{i\to\nul})=0$, only the term of $j$ is nonzero, and it equals $\ncat^{-1}\ln(1+\nobs/(\nslots\pcount))$.
For $y\coloneqq\nobs/(\nslots\pcount)\in(0,1]$, the concavity of the logarithm gives $\ln(1+y)\ge y\ln(2)\ge y/2$.
\end{proof}

Laplace noise calibrated to this global sensitivity therefore needs order $\nobs\ln(1/\fprob)/(\ncat\pcount\acc\eps)$ users for error $\acc$ with probability $1-\fprob$ when $\nslots\pcount\ge\nobs$, a statement about this calibration and not a lower bound for other mechanisms.
A public lower bound on the smoothed frequencies would remove the factor $\nobs/(\ncat\pcount)$, but \eqref{eq:relative_histogram} supplies no such bound, and the uniformity of the reference does not either.

\subsection{All-Pairs Effective Resistance}\label{app:graph_resistance}

This subsection proves the guarantees of \Cref{sec:graph}.

\paragraph{Setting.}
Given a public set $\verts$ of $v\coloneqq|\verts|\ge2$ vertices, there is one record slot for every unordered pair of vertices.
An edge $e=\{u,u'\}$ of weight $w_e\ge0$ has the record $\cand_e\coloneqq\sqrt{w_e}(\chi_u-\chi_{u'})$, where $\chi_u$ is the standard basis vector of $u$.
An unweighted graph has weights in $\{0,1\}$, and an absent edge has the ordinary zero record, which is not $\nul$.
The dataset may be padded with a public number of ordinary zero records, and $\nslots$ denotes the number of slots after padding.
The Laplacian of a graph $G$ is $\lapl_G\coloneqq\sum_e\cand_e\cand_e^T$, and the effective resistance between two vertices in the same component is $\reff_G(u,u')\coloneqq(\chi_u-\chi_{u'})^T\lapl_G^\dagger(\chi_u-\chi_{u'})$, where $\lapl_G^\dagger$ is the pseudoinverse of the singular matrix $\lapl_G$.
The leverage of an edge $e=\{u,u'\}$ is $\ell_e\coloneqq w_e\reff_G(u,u')$, which is unchanged by scaling all the weights, and we denote $\lmax\coloneqq\max_{e:w_e>0}\ell_e$.

\phantomsection\paragraph{Score and release mechanism.}\label{app:graph_score}
We use the score $\fscore_0$ of \Cref{def:leverage}, with zero ridge, on edge records, which is concave by \Cref{lem:lev_concave}.
For a nonempty reference dataset $\ds$ of edge records, $\lapl_{\ds}=\sum_{e\in\ds}\cand_e\cand_e^T$, where null records contribute zero and are still counted in $|\ds|$.
For an ordinary record $\cand\ne0$, the leverage of \Cref{def:leverage} with zero ridge is
\[
 \lev_0(\ds,\cand)=\lim_{\ridge\downarrow0}\cand^T\left(\frac{\lapl_{\ds}}{|\ds|}+\ridge I\right)^{-1}\cand
 =\begin{cases}|\ds|\,\cand^T \lapl_{\ds}^\dagger \cand,&\cand\in\operatorname{range}(\lapl_{\ds}),\\
 \infty,&\cand\notin\operatorname{range}(\lapl_{\ds}),
 \end{cases}
\]
and $\fscore_0(\ds,\cand)=1/\lev_0(\ds,\cand)$.
For an edge record, being outside the range means that the endpoints of the edge lie in different components of the reference graph, so a finite pseudoinverse quadratic form across components is never used as a certificate.
An ordinary zero record has score $\infty$ and is friendly with every nonempty block.
After filtering, the release mechanism $\mech_{\mathrm{pot}}$ draws independent potential vectors $Y_1,\ldots,Y_\nsk\sim\Gauss(0,\lapl_{\fds}^\dagger)$, which form an embedding of the vertices rather than a synthetic graph.
They answer every pairwise query through $\widehat\reff(u,u')\coloneqq\nsk^{-1}\sum_{j\in[\nsk]}(Y_j(u)-Y_j(u'))^2$, since $Y_j(u)-Y_j(u')$ has variance $(\chi_u-\chi_{u'})^T\lapl_{\fds}^\dagger(\chi_u-\chi_{u'})$, which is the effective resistance $\reff_{\fds}(u,u')$ when $u$ and $u'$ are connected in $\fds$.

The potential vectors have the pseudoinverse of a Laplacian as their covariance, whereas a sketch has the second moment itself, and the next lemma reduces the privacy of the potentials to that of the sketches.

\begin{lemma}[Inversion and common support]\label[lemma]{lem:inverse_gaussian}
Given an order $\rord>1$ and positive definite matrices $C,C'$ on a common finite-dimensional space, we have $\Renyi{\rord}\bigl(\Gauss(0,C^{-1})\Vert\Gauss(0,C'^{-1})\bigr)=\Renyi{\rord}\bigl(\Gauss(0,C')\Vert\Gauss(0,C)\bigr)$, including the value $\infty$.
Consequently, the maximum of the two directed divergences is unchanged by inversion.
Moreover, given $\thr>0$ and a pair $\ds,\ds^{i\to\nul}$ of edge datasets admitted by $\pred_\thr(\fscore_0)$, the Laplacians $\lapl_{\ds^{i\to\nul}}$ and $\lapl_{\ds}$ have the same range, so the identity applies on that range.
\end{lemma}

\begin{proof}
The Gaussian integral that defines the left side is finite exactly when $B\coloneqq \rord C+(1-\rord)C'$ is positive definite, and then the R\'enyi divergence on the left side equals $(\rord\ln(\det C)+(1-\rord)\ln(\det C')-\ln(\det B))/(2(\rord-1))$.
For the right side, the precision matrix in the Gaussian integral is $\rord C'^{-1}+(1-\rord)C^{-1}=C^{-1}BC'^{-1}$, whose determinant is $\det B/(\det C\det C')$, which gives the same value.
The two finiteness conditions agree: diagonalizing the positive definite matrix $C^{-1/2}C'C^{-1/2}$ reduces both to the positivity of $\rord+(1-\rord)\lambda_j$ for each of its eigenvalues $\lambda_j$.

For the support claim, write the two Laplacians as $\lapl$ and $\lapl+\cand\cand^T$.
Friendliness at a positive threshold implies that the leverage of $\cand$ with respect to $\lapl$ is finite, hence $\cand\in\operatorname{range}(\lapl)$, and since the range of a sum of positive semidefinite matrices is the span of their ranges, the range of $\lapl+\cand\cand^T$ equals that of $\lapl$.
Restricting to this range, where $\lapl$ is positive definite, proves the claim, and if the range is zero-dimensional, both laws are the same point mass.
\end{proof}

\paragraph{Privacy.}
The release mechanism $\mech_{\mathrm{pot}}$ is defined even for a disconnected or empty graph, and makes no connectivity test.
On a friendly pair, \Cref{lem:inverse_gaussian} reduces the two directed divergences between the laws of one potential vector to those between $\Gauss(0,\lapl)$ and $\Gauss(0,\lapl+\cand\cand^T)$ on their common range, and scaling both covariances by $1/\nslots$ does not change them, so \Cref{prop:sketch_RDP} bounds them exactly as for the zero-ridge Gaussian sketch, since friendliness gives $\cand^T\lapl^\dagger\cand\le1/(\thr\nslots)$.
The divergences add over the independent potential vectors, and the threshold $\thr=2(\Kconst)/(\nslots\eps)$ of \Cref{thm:graph_resistance} gives $1/(\thr\nslots)=\eps/(2(\Kconst))$, so the calibration in the proof of \Cref{thm:filt_sketch_DP}, with $\slev\coloneqq1/(\thr\nslots)$, proves the privacy guarantee of \Cref{thm:graph_resistance}, also on inputs on which the utility conditions of \Cref{thm:graph_resistance_utility} fail.
For a desired budget $(\eps,\delta)$ with respect to replacing one edge, run the mechanism with $\eps_0=\eps/2$ and $\delta_0=\delta/3$, as in the remark after \Cref{claim:zero_out_replacement}, and compute all the public parameters, including $\Lconst$, from $(\eps_0,\delta_0)$.

The utility analysis requires that, for a fixed graph, all the random blocks of the filter represent its connectivity.

\begin{lemma}[Random blocks of a fixed graph]\label[lemma]{lem:graph_blocks}
Given $\bsz,\nblk,v\in\bbN$ and a connected graph $G$ on $v\ge2$ vertices, represented by $\nslots\coloneqq\nblk\bsz$ ordinary records, if these records are partitioned uniformly at random into $\nblk$ blocks $\ds^{(1)},\ldots,\ds^{(\nblk)}$ of $\bsz$ records, then
\[
 \Pr\left[\exists j\in[\nblk]:\ \lapl_{\ds^{(j)}}\not\succeq \frac{\lapl_G}{2\nblk}\right]
 \le\nblk(v-1)\exp\left(-\frac1{8\nblk\lmax}\right),
\]
and on the complementary event, every record has score at least $1/(2\nslots\lmax)$ against every block, where $\lmax$ is the maximum leverage of an edge of $G$.
\end{lemma}

\begin{proof}
On the subspace $\mathbf1^\perp$, with identity $I_\perp$, denote $A_e\coloneqq\lapl_G^{\dagger/2}\cand_e\cand_e^T\lapl_G^{\dagger/2}$.
These matrices satisfy $\sum_eA_e=I_\perp$ and $0\preceq A_e\preceq\ell_eI_\perp\preceq\lmax I_\perp$, and zero records contribute zero.
A fixed block is a uniform sample of $\bsz$ slots without replacement, so $\bbE\sum_{e\in\ds^{(j)}}A_e=I_\perp/\nblk$.
The comparison of trace exponentials in \citet[Section~D]{GrossNesme} shows that the moment-generating bounds behind the matrix Chernoff inequality for sampling with replacement also hold for sampling without replacement.
The lower tail of the matrix Chernoff inequality~\cite[Corollary~5.2]{tropp2012user}, in dimension $v-1$, with $\mu_{\min}=1/\nblk$ and $R=\lmax$, and the simplification of \citet[Remark~5.3]{tropp2012user} then give, for $0<\eta<1$,
\[
 \Pr\left[\lambda_{\min}\left(\sum_{e\in\ds^{(j)}}A_e\right)
 \le\frac{1-\eta}{\nblk}\right]
 \le(v-1)\left(\frac{e^{-\eta}}{(1-\eta)^{1-\eta}}\right)^{1/(\nblk\lmax)}
 \le(v-1)e^{-\eta^2/(2\nblk\lmax)}.
\]
Take $\eta=1/2$ and apply a union bound over the $\nblk$ blocks, which need not be independent.
Conjugating by $\lapl_G^{1/2}$ gives the asserted Loewner inequality on $\mathbf1^\perp$, and both sides vanish on $\mathbf1$.
On this event every block has range $\mathbf1^\perp$, and inverting the inequality on $\mathbf1^\perp$ gives, for every edge,
\[
 |\ds^{(j)}|\,\cand_e^T \lapl_{\ds^{(j)}}^\dagger \cand_e
 \le\frac{\nslots}{\nblk}\,2\nblk\, \cand_e^T \lapl_G^\dagger \cand_e
 =2\nslots\ell_e\le2\nslots\lmax,
\]
so every edge has score at least $1/(2\nslots\lmax)$ against every block, and absent edges and padding have score $\infty$.
\end{proof}

The following theorem states the utility of the mechanism of \Cref{thm:graph_resistance}; privacy needs only the public lower bound $\mPriv$ on $\nblk$, and the second lower bound on $\nblk$ serves retention.

\begin{theorem}[Utility of private all-pairs effective resistance]\label{thm:graph_resistance_utility}
Given $\nslots,\bsz\in\bbN$ with $\nslots/\bsz\in\bbN$; $\acc\in(0,1)$; $\eps\in(0,1]$; $\delta\in(0,1)$; and $\fprob_B,\fprob_F,\fprob_Q>0$ with $\fprob\coloneqq\fprob_B+\fprob_F+\fprob_Q<1$ and $\delta\le\cFprob\fprob_F$, if $\nblk\ge\max\{\mPriv,\ \mRet{\fprob_F}\}$ and the input is a connected graph $G$ on the public vertices, given as $\nslots$ records without null records, whose maximum leverage $\lmax$ satisfies
\begin{equation}\label{eq:graph_utility_conditions}
 \lmax\le\frac{\eps}{8(\Kconst)}
 \qquad\text{and}\qquad
 \nblk(v-1)\exp\left(-\frac1{8\nblk\lmax}\right)\le\fprob_B,
\end{equation}
then with probability at least $1-\fprob$, the filter of $\PFRd(\mech_{\mathrm{pot}},\fscore_0,\thr)$ with $\nsk$ potential vectors retains every record, and every pair of distinct vertices $u,u'$ satisfies $|\widehat\reff(u,u')-\reff_G(u,u')|\le\acc\,\reff_G(u,u')$, where $v$ is the number of vertices, $\nblk\coloneqq\nslots/\bsz$, $\Lconst\coloneqq\Ldef$, $\tQ\coloneqq\ln(v^2/\fprob_Q)$, $\nsk\coloneqq\lceil8\tQ/\acc^2\rceil$ and $\thr\coloneqq2(\Kconst)/(\nslots\eps)$.
\end{theorem}

The mechanism never receives or computes $\lmax$, which enters only the two utility conditions of \eqref{eq:graph_utility_conditions}.

\begin{proof}[Proof of \Cref{thm:graph_resistance_utility}]
By \Cref{lem:graph_blocks} and the second condition of \eqref{eq:graph_utility_conditions}, with probability at least $1-\fprob_B$ every record has score at least $1/(2\nslots\lmax)$ against every block, and the first condition gives $1/(2\nslots\lmax)\ge2\thr$.
All the friend counts are then $\nblk$, so no record has $\cnt_i\le\athrRet$.
Conditional on any such partition, \Cref{cor:pfr_retention}, with failure probability $\fprob_F\ge\fprobMin$, and the assumed lower bound on $\nblk$ ensure that the filter retains every record, except with probability $\fprob_F$.

On full retention, the release mechanism uses exactly $\lapl_G^\dagger$.
For any distinct $u,u'$, the standardized differences $(Y_j(u)-Y_j(u'))/\sqrt{\reff_G(u,u')}$ are independent standard Gaussians, so $\nsk\widehat{\reff}(u,u')/\reff_G(u,u')$ has the chi-square distribution with $\nsk$ degrees of freedom.
The chi-square tail bounds of \citet[Lemma~1]{laurent2000adaptive} give
\[
 \Pr\left[\left|\frac{\widehat{\reff}(u,u')}{\reff_G(u,u')}-1\right|
 >2\sqrt{\frac{\tQ}{\nsk}}+\frac{2\tQ}{\nsk}\right]\le2e^{-\tQ},
\]
and a union bound over the $\binom v2$ pairs costs at most $v^2e^{-\tQ}=\fprob_Q$.
Since $\nsk\ge8\tQ/\acc^2$ and $0<\acc<1$, we have $2\sqrt{\tQ/\nsk}+2\tQ/\nsk\le\acc/\sqrt2+\acc^2/4<\acc$.
The randomness of the potentials is independent of that of the filter, so combining the three failure events proves the theorem.
\end{proof}

The second utility condition can be written as $\lmax\le1/(8\nblk\ln(\nblk(v-1)/\fprob_B))$, which, when $\nblk$ is of the order of its lower bounds in \Cref{thm:graph_resistance_utility}, as padding with zero records allows, is a restriction of the form $\lmax=\wtO(\eps^{3/2})$, where the logarithmic factors involve $1/\delta$, $1/\fprob_B$, $1/\fprob_F$, $\nslots$ and $v$.
The first condition holds when $\acc\ge64\lmax\sqrt{\tQ\Lconst}/\eps$ and $\lmax\le\eps/(16\Lconst)$, since $\nsk\le16\tQ/\acc^2$, and it implies $\acc\ge8\lmax\sqrt{8\tQ\Lconst}/\eps$, so the smallest relative error that it certifies is $\wtO(\lmax/\eps)$.
These are guarantees for a family of mechanisms indexed by $\acc$, and not a private adaptation of $\acc$ to the observed graph.
Since the leverages of the edges of a connected graph sum to $v-1$, a small $\lmax$ requires at least $(v-1)/\lmax$ edges, so trees and other graphs with a bridge fall outside this regime.
At most $v(v-1)/2$ edges have positive weight, so $\lmax\ge2/v$, and the second condition then requires $v\ge16\nblk\ln(\nblk(v-1)/\fprob_B)$.
The guarantee therefore concerns graphs with many more vertices than reference blocks: with $\eps=1$ and $\delta=10^{-6}$, the privacy condition $\nblk\ge\mPriv$ alone requires more than $5\cdot10^7$ vertices.

\subsection{Treatment-Rule Selection}\label{app:weighted_selection}

This subsection proves the guarantees of \Cref{sec:weighted_selection}.

\paragraph{Weighted classification.}
A record is $\cand=(\cov,\trt,\outc)$, with covariates $\cov$, an assignment $\trt\in\{0,1\}$ and an outcome $\outc\ge0$; in the JTPA example of \citet{kitagawa2018welfare}, $\cov$ consists of the years of education and the prior earnings of a participant, $\trt$ indicates assignment to job training, and $\outc$ is their 30-month earnings.
The propensity $\prop(\trt\mid\cov)>0$ is a known public function, as in a randomized trial,\footnote{Propensities fitted to the protected sample would make every weight depend on all the records.} and in the example $\prop(1\mid\cov)=2/3$ and $\prop(0\mid\cov)=1/3$ for every $\cov$.
The library $\hypclass$ is a public set of $\nhyp$ treatment rules, each mapping covariates to $\{0,1\}$, such as the threshold rules on the two covariates with thresholds on a public grid.
The weight of a record is $\weight(\cand)\coloneqq\outc/\prop(\trt\mid\cov)$, with $\weight(\nul)\coloneqq0$, so in the example a participant assigned to training has weight $3\outc/2$ and any other participant $3\outc$.
The loss of a rule $h\in\hypclass$ is $\loss_h(\cand)\coloneqq\ind\{h(\cov)\ne\trt\}$, so that the outcome of a participant counts for $h$ exactly when $h$ agrees with their actual assignment.
For a dataset $\ds\in\dom^\nslots$, we denote its total weight and its weighted risks by
\[
 \totweight(\ds)\coloneqq\sum_{i\in[\nslots]}\weight(\ds_i),
 \qquad
 \wrisk_h(\ds)\coloneqq\frac{\sum_{i\in[\nslots]}\weight(\ds_i)\loss_h(\ds_i)}{\totweight(\ds)},
\]
with $\wrisk_h(\ds)\coloneqq0$ if $\totweight(\ds)=0$, and with null records contributing zero.
The raw earnings of the example are nonnegative and fit this setting without clipping, whereas the cost-adjusted earnings that \citet{kitagawa2018welfare} also use can be negative and do not.

\begin{lemma}[Per-instance sensitivity of the weighted risks]\label[lemma]{lem:weighted_risk_sensitivity}
Given $\nslots\in\bbN$; $i\in[\nslots]$; a dataset $\ds\in\dom^\nslots$ and a rule $h\in\hypclass$, if $\totweight(\ds^{i\to\nul})>0$, then
\[
 \wrisk_h(\ds)-\wrisk_h(\ds^{i\to\nul})
 =\frac{\weight(\ds_i)}{\totweight(\ds)}\bigl(\loss_h(\ds_i)-\wrisk_h(\ds^{i\to\nul})\bigr)
 \qquad\text{and}\qquad
 \bigl|\wrisk_h(\ds)-\wrisk_h(\ds^{i\to\nul})\bigr|\le\frac{\weight(\ds_i)}{\totweight(\ds^{i\to\nul})}.
\]
\end{lemma}

\begin{proof}
Denote $\ds'\coloneqq\ds^{i\to\nul}$ and $w\coloneqq\weight(\ds_i)$.
Nulling slot $i$ removes $w$ from the total weight and $w\loss_h(\ds_i)$ from the weighted sum of the losses, so $\totweight(\ds)=\totweight(\ds')+w>0$ and $\totweight(\ds)\wrisk_h(\ds)=\totweight(\ds')\wrisk_h(\ds')+w\loss_h(\ds_i)$.
Subtracting $\totweight(\ds)\wrisk_h(\ds')=\totweight(\ds')\wrisk_h(\ds')+w\wrisk_h(\ds')$ and dividing by $\totweight(\ds)$ gives the identity.
Both $\loss_h(\ds_i)$ and $\wrisk_h(\ds')$ lie in $[0,1]$, so their difference is at most one in absolute value, and $\totweight(\ds)\ge\totweight(\ds')$ gives the bound.
\end{proof}

\paragraph{Relative-weight score and release mechanism.}
The score $\sensW(\ds,\cand)=\totweight(\ds)/(|\ds|\,\weight(\cand))$ of \Cref{thm:weighted_selection} is defined for a nonempty reference dataset $\ds$ and a non-null record $\cand$, and it is $\infty$ when $\weight(\cand)=0$.
The release mechanism $\mech_{\mathrm{sel}}$ selects $h\in\hypclass$ with probability proportional to $\exp(-\eps\thr\nslots\wrisk_h(\ds)/4)$, with the weighted risks defined above, so it is uniform on the all-null dataset.

\begin{proof}[Proof of \Cref{thm:weighted_selection}]
\emph{Score and per-instance sensitivity.}
The score is invariant under permutations and has the prescribed value on empty reference datasets.
For nonempty datasets $\ds,\ds'$ and a record $\cand$ of positive weight, additivity of the total weight gives $\sensW((\ds,\ds'),\cand)=(|\ds|\,\sensW(\ds,\cand)+|\ds'|\,\sensW(\ds',\cand))/(|\ds|+|\ds'|)$, and for a non-null record of zero weight all three scores are $\infty$, so $\sensW$ is a concave friendliness score.

Given a friendly pair $\ds,\ds'\coloneqq\ds^{i\to\nul}$, denote $w\coloneqq\weight(\ds_i)$.
If $w=0$, every weighted sum and the total weight are unchanged, so every risk and the distribution of the release are identical, also under the convention for zero total weight.
If $w>0$, friendliness gives $w\le\totweight(\ds')/(\thr\nslots)$, which implies $\totweight(\ds')>0$, and \Cref{lem:weighted_risk_sensitivity} yields $|\wrisk_h(\ds)-\wrisk_h(\ds')|\le w/\totweight(\ds')\le1/(\thr\nslots)$ for every $h\in\hypclass$.
This bound holds without a positive lower bound on the weights.

\emph{Privacy.}
Denote $c\coloneqq\eps\thr\nslots/4$ and $Z(\ds)\coloneqq\sum_{h\in\hypclass}e^{-c\wrisk_h(\ds)}$.
By the sensitivity bound, every unnormalized probability changes by a factor between $e^{-\eps/4}$ and $e^{\eps/4}$, and so does $Z$, so for every $h\in\hypclass$,
\[
 \frac{\Pr[\mech_{\mathrm{sel}}(\ds)=h]}{\Pr[\mech_{\mathrm{sel}}(\ds')=h]}
 =e^{-c(\wrisk_h(\ds)-\wrisk_h(\ds'))}\frac{Z(\ds')}{Z(\ds)}
 \in[e^{-\eps/2},e^{\eps/2}].
\]
Summing over events proves $(\eRel,0)$-friendly DP, and \Cref{thm:gen_frnd_core_priv} gives $(\eps,\delta)$-DP for PFR.
\end{proof}

\paragraph{Utility.}
The next corollary shows that the filter retains all the records when no weight is much larger than the mean weight, and bounds the excess risk with respect to the population.

\begin{corollary}[Population welfare]\label[corollary]{cor:weighted_population}
There exists a universal constant $C$ such that the following holds.
Given $\nslots,\bsz\in\bbN$ with $\nslots/\bsz\in\bbN$; $\eps\in(0,1]$; $\delta\in(0,1)$; $\fprob\in(0,1/4)$ with $\delta\le\cFprobQ\fprob$; a friendliness threshold $\thr\in(0,1/\cThr]$, a public library $\hypclass$ of $\nhyp$ treatment rules, and a distribution over non-null records whose weight satisfies $0\le\weight\le\mu/(\cThr\thr)$ almost surely for $\mu\coloneqq\bbE[\weight]\in(0,\infty)$, if the $\nslots$ records are independent draws from this distribution, $\bsz\ge(\cBszW/\thr)\ln(4\nblk/\fprob)$ and $\nblk\ge\max\{\mPriv,\ \mRetQ\}$, then with probability at least $1-\fprob$, $\PFRd(\mech_{\mathrm{sel}},\sensW,\thr)$ retains every record and selects a rule $\widehat h$ with
\[
 \wrisk(\widehat h)-\min_{h\in\hypclass}\wrisk(h)
 \le C\left[\sqrt{\frac{\ln(8(\nhyp+1)/\fprob)}{\thr\nslots}}+\frac{\ln(8(\nhyp+1)/\fprob)}{\thr\eps\nslots}\right],
\]
where $\nblk\coloneqq\nslots/\bsz$, $\Lconst\coloneqq\Ldef$, and $\wrisk(h)\coloneqq\bbE[\weight\loss_h]/\bbE[\weight]$ is the population risk of a rule $h$.
\end{corollary}

Consequently, $\nslots=\wtO(1/(\thr\eps^{3/2})+\ln(\nhyp/\fprob)/(\thr\acc^2)+\ln(\nhyp/\fprob)/(\thr\acc\eps))$ records suffice for excess risk $\acc\in(0,1)$.
Since $\wrisk(h)=1-V(h)/\mu$ for the expected welfare $V(h)\coloneqq\bbE[\outc\ind\{\trt=h(\cov)\}/\prop(\trt\mid\cov)]$ of a rule under the randomized design, this is a welfare regret of $\mu\acc$, although the algorithm uses no value of $\mu$.
The blocks matter for retention: when only a small fraction of the records have positive outcomes, every record of positive weight is unfriendly with a singleton reference that has a zero outcome, whereas with blocks of $\wtO(1/\thr)$ records the filter retains all such records.
The condition on the weights bounds them relative to their unknown mean, so it describes a bounded relative support with an unknown scale rather than a heavy-tailed distribution; privacy, in contrast, holds for arbitrary weights.
Scale-adaptive alternatives, such as clipping at a privately estimated scale of the weights, may also avoid an absolute bound, and we do not compare with them.

\begin{proof}[Proof of \Cref{cor:weighted_population}]
\emph{Representative blocks.}
For the analysis only, normalize the weights as $U_i\coloneqq\cThr\thr\weight(\ds_i)/\mu\in[0,1]$, so that $\bbE U_i=\cThr\thr$.
For every $\xi\ge0$, the inequality $e^{-\xi y}\le1+(e^{-\xi}-1)y$ for $y\in[0,1]$ bounds the moment-generating function of $-U_i$ by that of a Bernoulli variable with mean $\cThr\thr$, and Markov's inequality at $\xi=\ln(2)$ gives, for any fixed block of $\bsz$ records,
\[
 \Pr\left[\frac1\bsz\sum_{i\in\ds^{(j)}}\weight(\ds_i)<\frac\mu2\right]
 \le\exp\bigl(-2(1-\ln(2))\thr\bsz\bigr)\le\exp\left(-\frac{\thr\bsz}{\cBszW}\right).
\]
The partition is independent of the records, so conditional on it the records of each block are independent draws, and a union bound over the blocks and the lower bound on $\bsz$ show that all the block averages are at least $\mu/2$, except with probability $\fprob/4$.
On this event, every record of positive weight has $\sensW(\ds^{(j)},\ds_i)\ge(\mu/2)\cdot\cThr\thr/\mu=2\thr$ against every block, and every record of zero weight, which is not null by assumption, has score $\infty$.
\Cref{cor:pfr_retention}, with failure probability $\fprob/4\ge\fprobMin$, since $\delta\le\cFprobQ\fprob$, and the assumed number of blocks, then shows that the filter retains all the records, except with probability $\fprob/4$.

\emph{Population concentration.}
Denote $v_i\coloneqq\weight(\ds_i)/\mu$, $\bar v\coloneqq\nslots^{-1}\sum_iv_i$ and $\bar v_h\coloneqq\nslots^{-1}\sum_iv_i\loss_h(\ds_i)$, so that $0\le v_i\le1/(\cThr\thr)$, $\bbE v_i=1$ and $\bbE[v_i\loss_h(\ds_i)]=\wrisk(h)$.
The variance of $v_i$ is at most $\bbE v_i^2\le1/(\cThr\thr)$, and since $(v_i\loss_h)^2\le v_i\loss_h/(\cThr\thr)$, the variance of every summand of $\bar v_h$ is at most $1/(\cThr\thr)$ as well.
With $J\coloneqq\ln(8(\nhyp+1)/\fprob)$, Bernstein's inequality and a union bound over these $\nhyp+1$ averages give, except with probability $\fprob/4$,
\[
 |\bar v-1|\le\zeta
 \qquad\text{and}\qquad
 |\bar v_h-\wrisk(h)|\le\zeta\ \ \text{for all }h\in\hypclass,
 \qquad\text{where }\zeta\coloneqq\sqrt{\frac{J}{2\thr\nslots}}+\frac{J}{6\thr\nslots}.
\]
If $\zeta\le1/2$, then $\bar v\ge1/2$, the empirical total weight is positive, and for every rule
\[
 |\wrisk_h(\ds)-\wrisk(h)|
 =\left|\frac{\bar v_h}{\bar v}-\wrisk(h)\right|
 \le2\bigl(|\bar v_h-\wrisk(h)|+\wrisk(h)|1-\bar v|\bigr)
 \le4\zeta.
\]
This concentration holds simultaneously for the whole public library, so it also applies to the selected rule.

\emph{Selection.}
Denote $c\coloneqq\eps\thr\nslots/4$, and condition on any realized filtered dataset $\fds$, with $\wrisk_*\coloneqq\min_{h\in\hypclass}\wrisk_h(\fds)$.
The normalizing sum $\sum_{h\in\hypclass}e^{-c\wrisk_h(\fds)}$ of the exponential mechanism is at least $e^{-c\wrisk_*}$, so for every $\xi>0$, a union bound over the rules gives $\Pr[\wrisk_{\widehat h}(\fds)>\wrisk_*+\xi\mid\fds]\le\nhyp e^{-c\xi}$.
With $\xi\coloneqq c^{-1}\ln(4\nhyp/\fprob)$, this gives $\wrisk_{\widehat h}(\fds)-\wrisk_*\le4\ln(4\nhyp/\fprob)/(\thr\eps\nslots)$, except with probability $\fprob/4$, and on full retention $\fds=\ds$.
If $\zeta\le1/2$, adding the population-to-empirical errors of the selected rule and of a population minimizer bounds the excess population risk by $8\zeta+4\ln(4\nhyp/\fprob)/(\thr\eps\nslots)$, which, since $\eps\le1$, is at most the bound of \Cref{cor:weighted_population} for a suitable universal constant $C$.
If $\zeta>1/2$, the same conclusion follows from the trivial bound $1$ on the excess risk, after enlarging $C$.
The probabilities of the four failure events, for the blocks, the filter, the concentration and the selection, sum to at most $\fprob$.
Finally, $\bbE[\weight\loss_h]=\mu-V(h)$ gives the identity $\wrisk(h)=1-V(h)/\mu$ for the welfare.
\end{proof}